\documentclass[sageh,Afour]{sagej}

\usepackage{bm}

\usepackage{enumitem}
\setlistdepth{9}
\setlist[itemize,1]{label=$\bullet$}
\setlist[itemize,2]{label=$\circ$}
\setlist[itemize,3]{label=$\ast$}
\setlist[itemize,4]{label=$\diamond$}
\setlist[itemize,5]{label=$-$}
\renewlist{itemize}{itemize}{5}

\begin{document}

\runninghead{Redfield}

\title{A review of robotics taxonomies in terms of form and structure}

\author{Signe A. Redfield\affilnum{1}}

\affiliation{\affilnum{1}Naval Research Laboratory}

\corrauth{Signe Redfield\\
Naval Center for Space Technologies\\
Naval Research Laboratory\\
4555 Overlook Ave, SW\\
Washington, DC, 20375}

\email{signe.redfield@nrl.navy.mil}

\begin{abstract}
Identifying and categorizing specific robot tasks, behaviors, and resources is an essential precursor to reproducing and evaluating robotics experiments across laboratories and platforms.  Without some means of capturing how one environment, platform, or behavior differs from another, we cannot begin to establish the performance impact of these changes or predict a robot's performance in a novel environment.  As a first step towards experimental reproducibility, existing taxonomies in the field of robotics are reviewed and common patterns of structure and form extracted, identifying both the properties they share with traditional taxonomies and the necessary structural elements that draw from other classification and categorization systems.  The diversity of taxonomy subjects and subsequent difficulty in harmonization of conceptual underpinnings is noted.  Robotics taxonomies are shown to be deeply fragmented in structure and form and to require notation that can support complex relationships. 
\end{abstract}

\maketitle

\section{Introduction}
\label{sec:introduction}

Experimental reproducibility is a significant problem in robotics.  Without the ability to reliably reproduce experiments, we cannot use the scientific method to improve robot behavior design.  Beyond comparing the performance of two algorithms, without experimental reproducibility we cannot even determine whether we have optimally implemented someone else's algorithm, or ensure that the bug fix we developed actually prevents the bug from happening.  

Significant advances in robotics over the last thirty years have been primarily driven by improved computing power, improved sensors and perceptual processing, and improved standardization of systems across labs.  Anyone who engages in a craft knows that having professional tools will improve even an adequate craftsman's work.  The field is experiencing a Cambrian explosion \cite[The Next Big Thing(s) In Robotics]{Winfield14} in robotic capability not because improvements in fundamental understanding of our systems have improved our behavior designs, but because our tools have improved.  Our tools have gotten better and better, but outside of the controls analysis community, our understanding of what makes an autonomous behavior successful has barely altered.  One of the major reasons for this is that, in the general sense, we cannot reproduce an experiment.  We are unable to use the scientific method to improve our understanding of the relationships between behavior design, task context, and system performance because we are unable to specify what has changed between experiment runs.

There are several ways we could improve our ability to reproduce robotics experiments.  Publication drives all aspects of academic participation, so we as a community have established a publication venue that encourages researchers to perform reproducible experiments and to reproduce experiments \citep{Bonsignorio17}, reducing the penalty for spending time on something that does not create new knowledge (beyond knowledge related to the robustness of the results in the first paper).  We have defined the GEM guidelines \citep{GEMGuidelines} that define what a reproducible experiment looks like --- what information needs to captured and published, and what kinds of controls need to be applied --- a first step towards a code of professional conduct for robotics research.  

But we also need a mechanism to ensure that when two researchers are using the same behavior, we understand both the differences and the similarities in their implementations.  We need a way to capture the details of our experiments in the same way that biologists need the Taxonomy of Species to label organisms.  Without the Taxonomy of Species, biologists can't specify what organism they work with.  Similarly, without taxonomies of robots and tasks, we cannot describe precisely what system we used and its performance goals.  Without a way to describe the differences between two experiments, we cannot identify the contextual information that may predict why sometimes we can reproduce an experiment and sometimes we can't.  We need something that can capture that information.  The set of existing robots is not larger or more complex than the set of existing and prior organisms; presumably we could define both a taxonomy of robots and a taxonomy-like structure that classifies tasks, including their contextual information.

As a precursor to the development of those taxonomies, this paper aggregates the taxonomies that have been proposed in the literature for many different aspects of robot systems and tasks, and identifies patterns and properties of structure and form in support of future development of these larger taxonomies.  The effort to develop a notation that can capture these patterns and structures and the effort to integrate the presented taxonomies into a larger structure are left for future work.

\section{Background}
\label{sec:background}
\begin{figure}[t]
\begin{center}
\includegraphics[width=3.25in, viewport=90in 100.5in 102in 107in, clip=true]{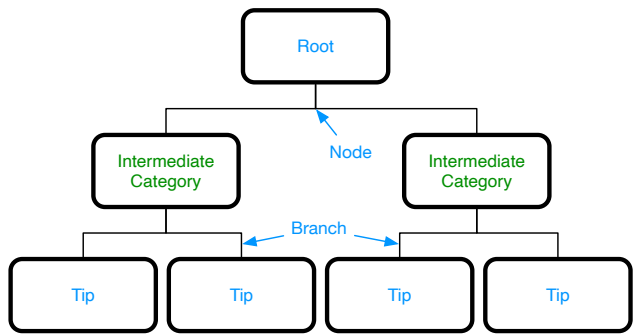}
\caption{Nomenclature for the components of a phylogenetic taxonomy.}
\label{fig:Nomenclature}
\end{center}
\end{figure}

The terminology describing the structure of the taxonomies as presented here is taken from the evolutionary genetics community.  Phylogenetic taxonomies have a purely hierarchical structure, where the lines connecting levels of the hierarchy are called {\em branches}, the decision points are called {\em nodes}, the top-most level of the hierarchy is called the {\em root}, and the termination points are called {\em tips} \citep{Baum08}.  In phylogenetic taxonomies, tips can represent individual genes, individual organisms, species, or sets of species.  Here, they represent the last defined element (category or value assignment) in the structure.  Because we need to assign categorical names to the elements that define the branches at each level of the hierarchy, the term {\em intermediate category} is used for the branch descriptors defined between the root and the tips, as illustrated in Fig.~\ref{fig:Nomenclature}.
% Taxonomy component nomenclature figure

In this paper, an {\em instance} of a taxonomy is the classification that results from using the taxonomy to categorize a specific item.  An instance of the Taxonomy of Species is the classification of a specific animal as {\em Homo Sapiens}.  

{\em Branch selection} mechanisms define how decisions are made when a node is reached in the taxonomy.  

Because the taxonomies reviewed in this paper are not consistent in their representation or assignment of the root, the intermediate categories, the nodes and branch selection mechanisms, and the tips, some degree of judgement was necessary to determine what types of tips a given taxonomy uses, and to determine the branch selection mechanism implied at any given node.  In these cases, the illustrative figures capture the specific taxonomic structure and tips extracted from each paper in a consistent format. Where a taxonomy provided a diagram or explicitly stated relationships between categories, that structure and those relationships were maintained in these figures and used to inform the data analysis section at the end.

\subsection{Branch Selection Categories}
Traditional taxonomies have a strictly hierarchical structure, with the user expected to follow a single path along the branches of  the taxonomy, making a classification decision at each node.  When the end of the path is reached, the path through the taxonomy defines the classification of the subject.

The Library of Congress Cataloging System (LCC), first developed by Herbert Putnam in 1876~\citep{LCC76}, is an example of this type of taxonomy.  Each book is categorized according to a single path through the LCC that results in an instance of the taxonomy:  a library call number referencing a single book.  We call this branch selection mechanism {\em branch select}.  Because there will be many copies of the same book, an instance of this taxonomy terminates in a {\em category} tip that refers to the set of all copies of the book, not the particular book you are holding in your hand.   

There are alternatives to this type of traditional taxonomic structure.  In the first edition of {\em Colon Classification}, \citet{Ranganathan33} proposed a classification system that enables simultaneous classification along multiple {\em facets} --- independent axes along which classification may occur.  The instance in this case consists of a set of categories, one for each facet.  For example, a given system may classify along one facet representing the time of year, another facet representing the organism, and yet another facet representing systems of the body.  For a given instance, individual facets are separated with colons.  In this classification system, the combination [winter : polar bear : respiratory system] would lead to information about how hibernation affects the polar bear respiratory system.  By relaxing the organism constraint, information about winter hibernation respiration of other animals could be found.  By relaxing the seasonal constraint, information about the polar bear respiratory system in general could be found.  By relaxing the respiratory system constraint, information about other aspects of polar bears in the winter could be found.  Because each facet is assumed to be a traditional hierarchical taxonomic structure, the organism facet could be the Taxonomy of Species, enabling clear identification of particular organisms or groups of organisms.  

Faceted classification systems are often used in case-based reasoning systems like travel booking sites, where users are asked to provide information along a number of different relevant axes such as flight duration, date, airport, and time of day to refine the search for the desired flight.  Each search generates multiple instances, filtered by the desired criteria, until the user selects the single instance that best matches their needs.  We call this branch selection mechanism {\em faceted}.  

While the Taxonomy of Species and the LCC are both purely branch select taxonomies, robotics taxonomies take many forms.  They include facet-only structures, structures where the initial node below the root is branch select and all subsequent nodes are faceted (Mixed Type I), structures where the initial node below the root is faceted and all subsequent nodes are branch select (Mixed Type II), structures where the top node is branch select and subsequent decisions include both branch select and faceted nodes (Mixed Type III), and structures where the top node is faceted and subsequent decisions include both branch select and faceted nodes (Mixed Type IV).

\subsection{Tip Categories}
In robotics taxonomies, we differentiate between tips that represent categories, such as behavior types or task properties, and tips that represent specific numbers.  We define tips that represent specific numbers or parameter values as {\em assign value} tips and tips that represent categories as {\em category} tips.  

In the travel booking flight selection example, the tips of the facets are typically specific details about particular flights rather than classes of flights.  For example, you might want to select only from non-stop flights, or from flights with just one stop.  The tip of this branch of the taxonomy is an assigned value for the number of stops.  This is an example of an {\em assign value} tip.  

The original Taxonomy of Species did not support assign value tips.  If you needed to identify a specific individual member of a species (e.g. Lonesome George, the last Galapagos tortoise \citep{caccone99}), you would require additional information beyond the species name.  The Taxonomy of Species only enabled you to classify to the species level, so the assigned value rested outside the taxonomy, and the tip at the end of the branch continued to be a category tip with Lonesome George's species.  However, with the shift to a cladistic scheme for the taxonomy, it became possible to define specific individuals at a genetic level while maintaining the same categorization structure.  In this case, assign value tips can be used when the taxonomy is extended to capture a genome or part of a genome of a specific individual.

Because the end point in robotics taxonomies varies from extremely broad (``Tasks that require multiple systems", Fig.~\ref{fig:MRSDudekTask}) to extremely narrow (``Required Speed", Fig.~\ref{fig:HRITsiakas}), both assign value and category tips are necessary to represent the various taxonomic structures presented here.

In most of the reviewed taxonomies, the node above the tips uses a branch select structure.  Although faceted tips are possible, they do not allow the user to distinguish between specific instances and therefore indicate an incomplete taxonomy, with an undefined lower level category or assign value tips structure.

\section{Existing Robotics Taxonomies}
\label{sec:existingTaxonomies}
Because our purpose is to identify and illustrate patterns in structure and form, the allocation of taxonomies into sections is driven by their structure rather than their content.  However, one of the questions we need to answer is whether there is anything consistent or unique about how structure may relate to content type.  With this in mind, three content types are defined that would contribute to the desired larger taxonomy that is our overall goal.

The three content types captured by the various taxonomies include resource descriptors (including resource properties), task types, and task properties.  
\begin{itemize}
\item A {\em resources} taxonomy categorizes resources in the form of:  material needed by the system, the system itself, and environmental characteristics, as well as resource properties, which include system properties, environmental properties, and properties of objects in the environment.  In this context, ``system'' could be a robot or a human-robot interface or any other element acting as the agent for the purposes of the task.  
\item A {\em task type} taxonomy is an organizational framework that classifies specific job types.  This type of taxonomy categorizes individual specific jobs like ``pick up'' rather than identifying the properties of jobs.  
\item A {\em task properties} taxonomy categorizes different properties of jobs rather than the jobs themselves.   Instead of categorizing ``pick up'' relative to ``search'', these taxonomies are concerned with properties of jobs like ``references an object'' relative to ``references a place''.
\end{itemize}

% Branch-only diagram
\begin{figure*}[t]
\begin{center}
\includegraphics[width=5in, viewport=138in 140.8in 162.7in 149.1in, clip=true]{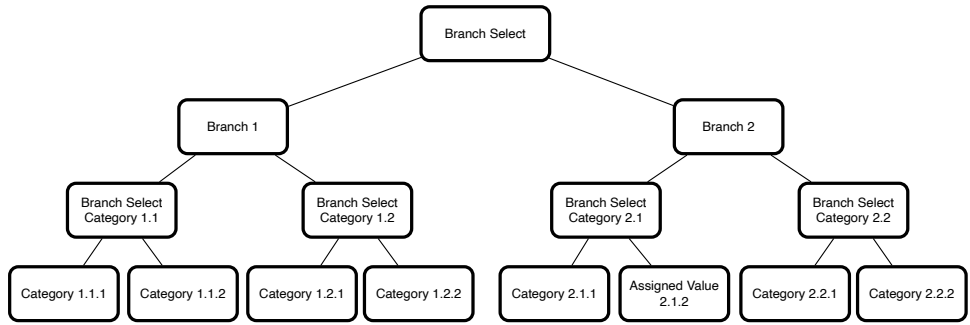}
\caption{Basic structure of branch only taxonomies.}
\label{fig:BranchOnly}
\end{center}
\end{figure*}

Although each taxonomy when taken as a whole typically addresses a single content type, many taxonomies use multiple content type categorizations within the taxonomy to identify instances.  In Section~\ref{sec:analysis}, each taxonomy's content type is captured as its Content, $Ct$. 

This document is organized according to form and structure rather than content, so each section addresses one kind of taxonomic structure.  
In Section~\ref{ssec:branchOnly}, we present taxonomies with an exclusively branch-select structure.  Section~\ref{ssec:facetOnly} holds the taxonomies whose categories are exclusively faceted, down to the last level above the tips (taxonomies where the final level of nodes above the tips are branch select are included here).  Section~\ref{ssec:facetOverBranch} has mixed taxonomies of type I where the top level is faceted and the lower levels take the branch-select form (facet-over-branch), while Section~\ref{ssec:branchOverFacet} has the mixed taxonomies of type II where the top level is branch-select and lower levels are faceted (branch-over-facet).  Finally, Section~\ref{ssec:trueMixed} contains the mixed taxonomies of types III and IV that have faceted and branch selection structures at varying levels.  Within each section, we will highlight where taxonomies use assign value tips instead of categories, as well as the places where one structure or tip could be replaced by a different type.

%%%%%%%    BRANCH ONLY    %%%%%%%%
\subsection{Branch Only Taxonomies}
\label{ssec:branchOnly}
The taxonomies in this section use the standard taxonomic selection structure from the Taxonomy of Species and library cataloging systems, where only one branch can be selected at each node.  The basic structure is illustrated in Fig.~\ref{fig:BranchOnly}, where each node uses a branch selection mechanism to choose only one of its sub-categories. 

The tips of each branch can be either categories or assigned values.  An instance in this structure would consist of the path Branch 1 $\Rightarrow$ (branch select node) Category 1.2 $\Rightarrow$ (branch select node) Branch Select Category 1.2.1, which would result in categorizing the item as Category 1.2.1.  

\subsubsection{\citet{Dudek96}}
\label{sssec:dudek1}
% Dudek task taxonomy
\begin{figure}[t]
\begin{center}
\includegraphics[width=3in, viewport=25.2in 112in 39.2in 118.4in, clip=true]{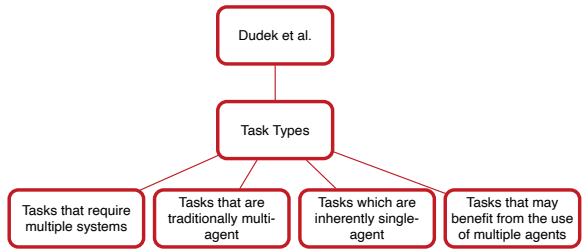}
\caption{Dudek et al.'s taxonomy of task types}
\label{fig:MRSDudekTask}
\end{center}
\end{figure}
Dudek, et al. contains two taxonomies; one for tasks categorized as a function of single vs. multiple agents, and the other to categorize types of multi-agent systems or collectives.  

The taxonomy shown in Fig.~\ref{fig:MRSDudekTask} categorizes tasks on the basis of the number of agents used or needed to accomplish it.  The core design decision that this taxonomy addresses, the descriptor of the root of this taxonomy, is ``what is the relationship between the number of agents available and the needs of the task?''  The resulting categories are:
\begin{itemize}
  \item{Tasks that require multiple agents: these tasks cannot be completed with a single agent}
  \item{Tasks that are traditionally multi-agent:  these tasks are traditionally accomplished with multiple agents, but do not necessarily require more than one agent to accomplish}
  \item{Tasks which are inherently single agent:  these tasks can only be done by a single agent}
  \item{Tasks that may benefit from the use of multiple agents:  these are tasks that could be done by a single agent, but whose performance might improve if additional agents were used}
\end{itemize}
This taxonomy has the simplest structure of all the taxonomies reviewed here, with a root, one branch select node, four category tips, and no intermediate categories.  

\subsubsection{\citet{Bullock11}}
\label{sssec:bullock}
% Bullock and Dollar taxonomy
\begin{figure*}[t]
\begin{center}
\includegraphics[width=6.5in, viewport=11in 24in 59in 38in, clip=true]{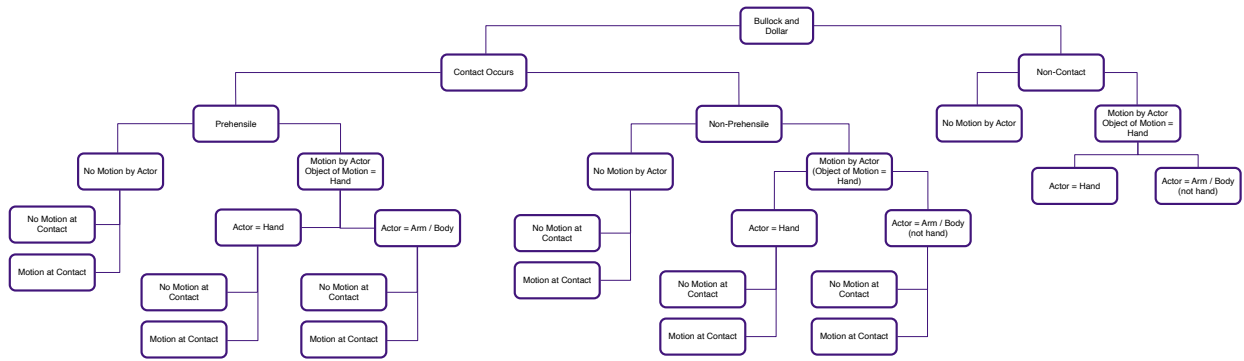}
\caption{Bullock and Dollar's taxonomy of manipulation activities --- branch only}
\label{fig:SBullockDollar}
\end{center}
\end{figure*}
\begin{figure*}[t]
\begin{center}
\includegraphics[width=5.5in, viewport=166in 127in 196in 135in, clip=true]{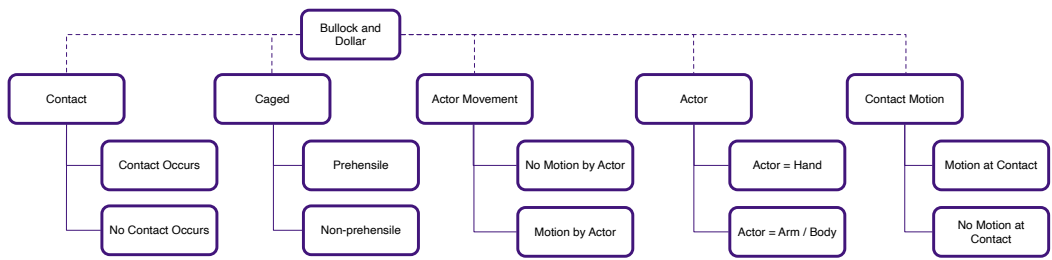}
\caption{Bullock and Dollar's taxonomy of manipulations --- facet only}
\label{fig:SBullockDollar_2}
\end{center}
\end{figure*}The taxonomy of manipulation tasks developed by Bullock and Dollar, on the other hand, is moderately complex.  Shown in Fig.~\ref{fig:SBullockDollar}, its root concern is manipulation based on hand-style grippers, including contact and non-contact manipulations.  It does not differentiate between manipulation where the goal is motion of the robot's arm or hand and manipulation where the goal is object motion, instead focusing on the mechanics of the grasp and the interaction between the agent and the object of motion during the manipulation.  This taxonomy terminates in categories rather than assigned values, and each node is a binary decision point.  

Because our long term goal is the creation of a large, integrated taxonomy, the efficiency with which we can represent this information is important.  This taxonomy could be reframed more compactly as a faceted taxonomy with the following binary facets:
\begin{itemize}
\item {{\em Contact}:  {\em Non-contact} manipulations involve actions where the object being manipulated is not manipulated through contact.  It covers situations where the object is part of the robot and is being moved independently (example = waving one's arm).  {\em Contact} manipulations involve actions where the object being manipulated is manipulated through contact.  It covers situations where the object is independent of the robot and requires robot motion to drive object motion (example = turning a screwdriver).}
\item {{\em Caged}:  {\em Prehensile} manipulations involve a caged grasp and force exerted against the inside of the grasp (example = holding a ball).  {\em Non-prehensile} manipulations rely on the environment to direct the object's motion and are not fully constrained by the grasp (example = flipping a light switch).}
\item {{\em Actor movement}:  {\em No motion by Actor} manipulations involve actively holding the Actor in place (example = holding one's arm in place against the wind).  {\em Motion by Actor} manipulations involve causing one's self to move (example = waving one's arm).}
\item {{\em Actor}:  {\em Not within hand} or {\em Actor=Arm/Body} manipulations involve moving the body or arm of the robot rather than the gripper or hand (example = doing a push-up), while {\em Within hand} or {\em Actor=Hand} manipulations involve moving the robot's gripper (example = grasping a pencil).}
\item {{\em Contact motion}:  {\em No motion at Contact} manipulations involve actively holding an object in place (example = holding the cup in place while pouring a glass of water).  {\em Motion at Contact} manipulations involve causing an object to move (example = using your palm to control the motion of a ball rolling across a table).  {\em ``at contact''} indicates that motion is occurring at the point where the robot makes contact with the object (object moves or deforms relative to contact point or contact point on object shifts), as opposed to the default case where the object is fixed relative to the contact point and moves relative to an external coordinate frame. }
\end{itemize}

This structuring of the taxonomy is shown in Fig.~\ref{fig:SBullockDollar_2}, with faceted nodes represented using dashed lines to connect the parent category to the child categories and branch selection nodes represented using solid lines to connect parent categories to child categories.  

Because combinations of facets that were not captured or were invalid in the initial structure are not explicitly disallowed in this structuring, the possible instances are increased from 15 to 32.  In this case, these combinations involve the motion-at-contact facet and the no-contact facet, which are obviously incompatible.  However, this new structure involves only 1 layer of intermediate categories instead of 5, has only 6 nodes instead of 14, and requires only 15 elements overall instead of 28.  If it becomes necessary to take advantage of this more efficient structure, a mechanism must be developed to identify invalid potential instances defined by a taxonomy.

\subsubsection{\citet{Feix15}}
\label{sssec:feix}
% Feix et al. taxonomy
\begin{figure*}[t]
\begin{center}
\includegraphics[width=4.5in, viewport=50.8in 118in 71.6in 147.7in, clip=true]{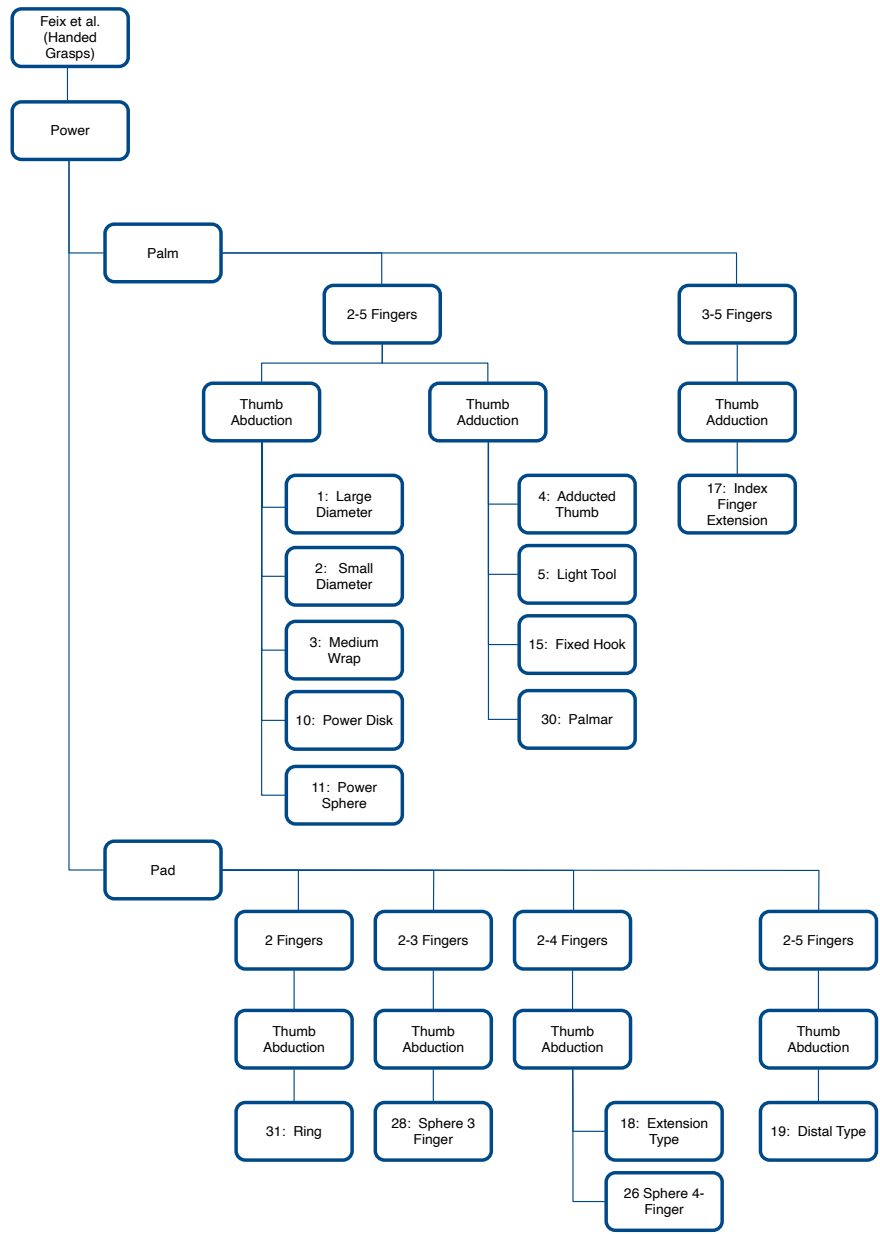}
\caption{Feix et al.'s taxonomy of grasps --- Power branch}
\label{fig:SFeix_1}
\end{center}
\end{figure*}
%Feix Intermediate
\begin{figure*}[t]
\begin{center}
\includegraphics[width=3.5in, viewport=86.9in 113in 103.4in 130.1in, clip=true]{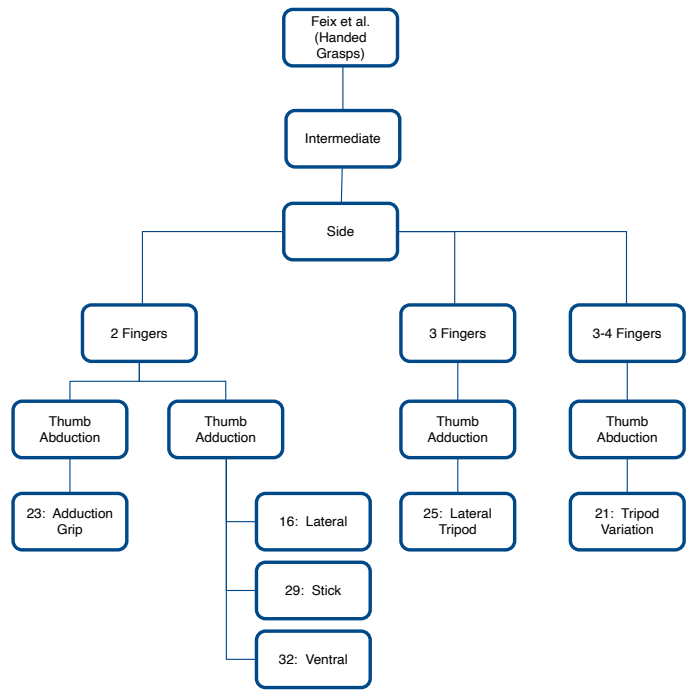}
\caption{Feix et al.'s taxonomy of grasps --- Intermediate branch}
\label{fig:SFeix_3}
\end{center}
\end{figure*}
%Feix Precision
\begin{figure*}[t]
\begin{center}
\includegraphics[width=6.5in, viewport=72.7in 133.8in 110.1in 149.9in, clip=true]{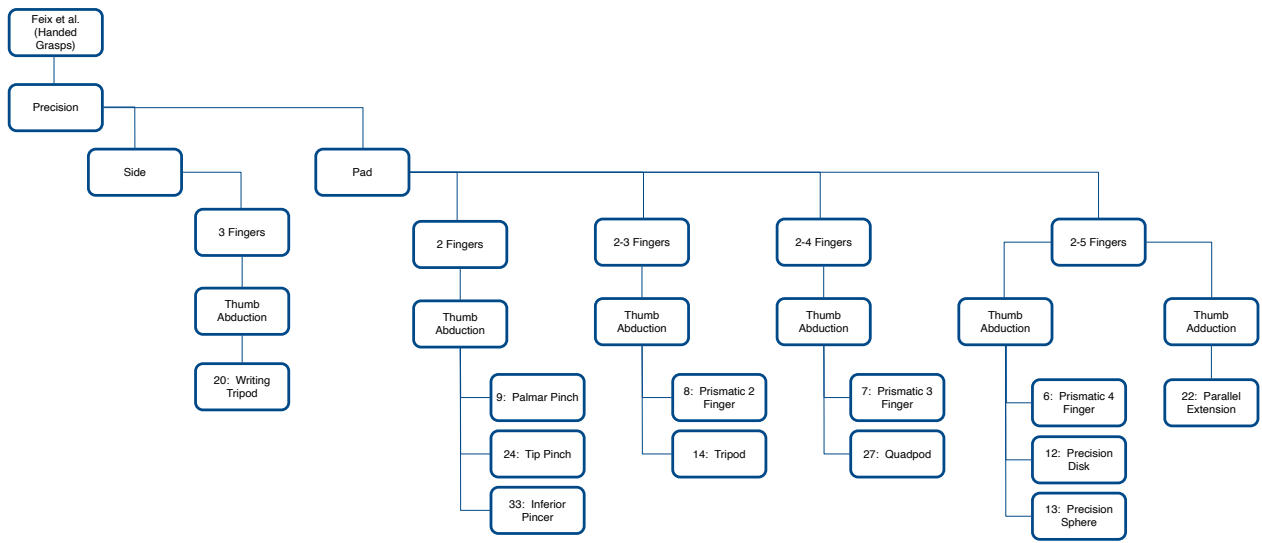}
\caption{Feix et al.'s taxonomy of grasps --- Precision branch}
\label{fig:SFeix_2}
\end{center}
\end{figure*}
%Feix Facet Only
\begin{figure*}[t]
\begin{center}
\includegraphics[width=4in, viewport=166.3in 139.1in 187.5in 149.2in, clip=true]{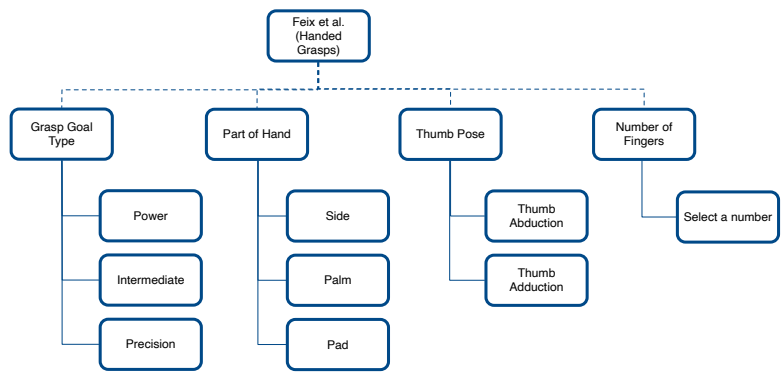}
\caption{Feix et al.'s taxonomy of grasps --- facet only}
\label{fig:SFeix_4}
\end{center}
\end{figure*}
% Fishwick's taxonomy
\begin{figure*}[t]
\begin{center}
\includegraphics[width=6.5in, viewport=70.3in 77.1in 105in 92.6in, clip=true]{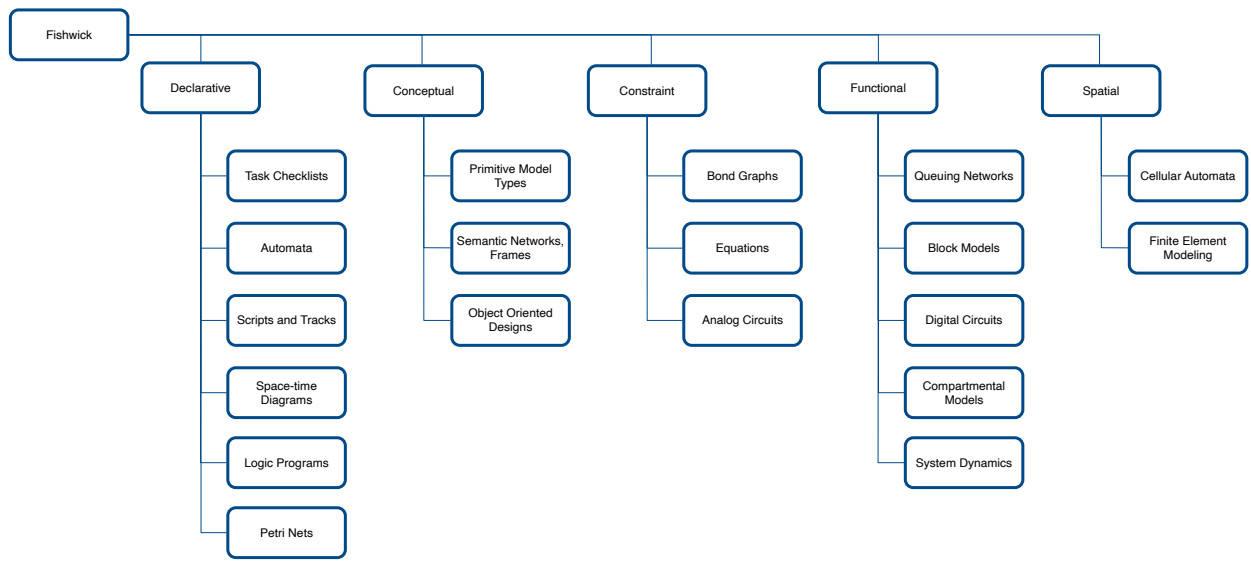}
\caption{Fishwick's taxonomy of simulation design models.}
\label{fig:SFishwick}
\end{center}
\end{figure*}

Feix el al.'s work has defined a grasp taxonomy based on human-style hands.  Due to its size, each top level branch is contained in its own figure.  Fig.~\ref{fig:SFeix_1} contains the Power branch,  Fig.~\ref{fig:SFeix_3} contains the Intermediate branch, and Fig.~\ref{fig:SFeix_2} contains the Precision branch.  

Some human-capable grasp types are excluded, and grasps specific to non-human grippers are out of scope.  Again, this is a strictly branch select structure, but instead of terminating at the category of grasp types (e.g. Power $\Rightarrow$ Palm $\Rightarrow$ 3-5 Fingers $\Rightarrow$ Thumb Adduction), it defines specific grasps as the termination categories (e.g. Power $\Rightarrow$ Palm $\Rightarrow$ 3-5 Fingers $\Rightarrow$ Thumb Adduction $\Rightarrow$ 17: Index Finger Extension).  Each numbered tip indicates a particular grasp type that is categorized according to that path through the taxonomy.  While the specific number referenced may seem to indicate an assign value tip type, these numbered grasps are treated as detailed categories defining specific grasp types rather than assigned values.  We know that multiple grasps can fall into a single instance of this taxonomy because in the above example, grasps involving 3, 4, or 5 fingers all fall into category tip 17: Index Finger Extension grasps.

As with Bullock and Dollar's taxonomy, this strictly branch select structure could be reframed with the following facets, as shown in Fig.~\ref{fig:SFeix_4}:
\begin{itemize}
\item{{\em Interaction Type},  capturing the type of interaction the grasp is intended for, where {\em Power} indicates a grasp that is used for forceful motion, while {\em Precision} grasps are used where precise motion is important and {\em Intermediate} grasps operate in the middle ground between them.}
\item{{\em Part of Hand},  capturing the part of the hand that is the focus of the grasp; it holds the categories {\em Palm}, {\em Pad}, and {\em Side}.}
\item{{\em Number of Fingers}, defining how many fingers are involved.  This is an assign value tip, defining either a specific number or a range of finger values.}
\item{{\em Thumb Pose}, identifying whether the thumb is placed across the palm ({\em Thumb Abduction}) or oriented parallel to the fingers ({\em Thumb Adduction}).}
\end{itemize}

Again, this restructuring introduces additional potential categories excluded by the original taxonomy.  In this case, the number of instances that this taxonomy can represent increases from 33 to 126.  As before, this is because there are certain combinations of facets that are not represented in the original taxonomy.  There are no Power $\Rightarrow$ Side grasps, no Precision $\Rightarrow$ Palm, Intermediate $\Rightarrow$ Palm, and no Intermediate $\Rightarrow$ Pad grasps.  The combinations of number of fingers and thumb pose are also restricted.

However, this new structure is also significantly more efficient, as the number of intermediate layers drops from 5 to 1, the number of elements needed to capture the possibilities drops from 72 to 19, and the number of nodes drops from 41 to 5.  As before, if we want to use the more efficient structure, we need a way to represent invalid combinations of facets.  This time, however, it is not clear that the invalid combinations are invalid because of inherently impossible relationships.  The interaction type/part of hand facet combinations can certainly define invalid instances --- a grasp involving the palm is almost certainly not appropriate for precision work unless there are no alternatives --- but the validity of the potential new relationships between the number of fingers and the thumb pose is less clear, especially if we attempt to generalize this taxonomy to less human-centric manipulators.

% Metzler and Shea's taxonomy
\begin{figure*}[t]
\begin{center}
\includegraphics[width=6.5in, viewport=7.75in 0in 52in 18.5in, clip=true]{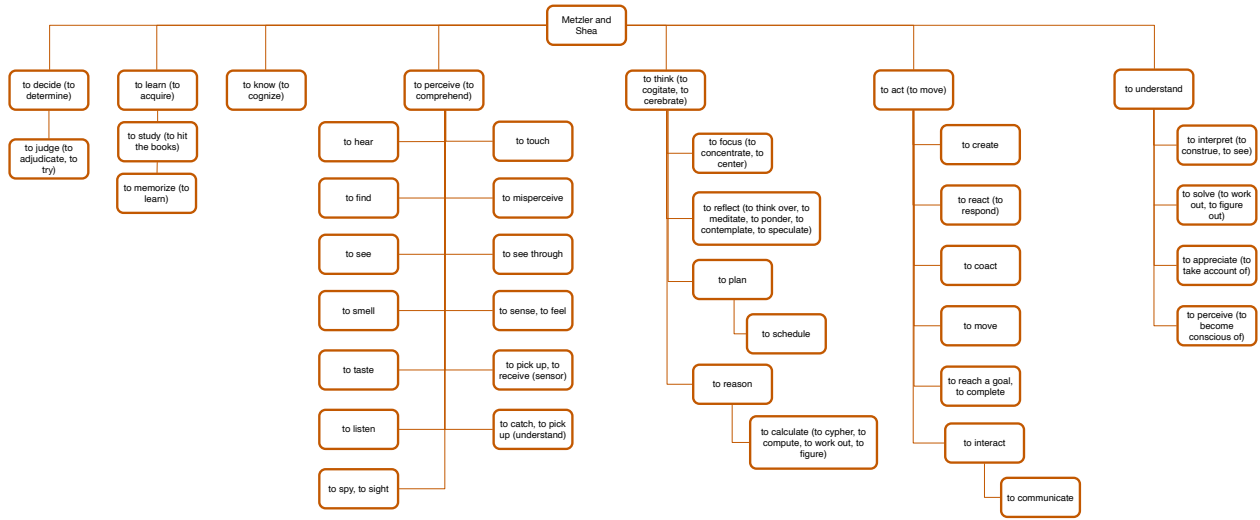}
\caption{Metzler and Shea's taxonomy of cognitive functions.}
\label{fig:SMetzlerShea}
\end{center}
\end{figure*}
% Ab.Acus robot type taxonomy 2 
\begin{figure*}[t]
\begin{center}
\includegraphics[width=3.5in, viewport=139.6in 22.8in 155.1in 28.7in, clip=true]{BySourceFigures.pdf}
\caption{Ab.Acus's taxonomy of robot types from~\citet{Siciliano09}}
\label{fig:SAbAcus_2}
\end{center}
\end{figure*}

\subsubsection{\cite{Fishwick98}}
\label{sssec:fishwick}
The root of Fishwick's taxonomy is the categorization of  simulation model types based on specific modeling techniques.  Each of the potential modeling techniques is associated with a specific simulation domain, but the individual category tips are defined by the modeling technique used in the simulation model design, not the domain in which it is typically found.  The decision to terminate the taxonomy with modeling techniques rather than domains was made because each domain is allocated to a single technique.  The domains do not create new nodes where categorization decisions must be made to generate specific instances.  

For example, if the Conceptual $\Rightarrow$ Semantic Networks instance was extended to include the suggested Artificial Intelligence tip, that would only be useful in defining instances of this taxonomy if other domains also using Semantic Networks were identified as additional tips.

This taxonomy is a relatively simple branch selection taxonomy, but while its structure has fewer layers than Feix et al. or Bullock and Dollar (only 1 intermediate layer, as shown in Fig.~\ref{fig:SFishwick}), the number of choices at each node varies from 2 to 6.  Any given instance only requires two elements to define it (e.g. Constraint $\Rightarrow$ Analog Circuits).

From a content type perspective, this is a resource taxonomy, as it provides a means to describe a desired mechanism required as part of a system rather than a job the system is expected to perform.  

Unlike Bullock and Dollar and Feix's taxonomies, this taxonomy has no redundancy that could be converted into a faceted form to improve its compactness.

\subsubsection{\cite{Metzler11}}
\label{sssec:metzler}
Metzler and Shea developed a very complex taxonomy of cognitive functions (shown in Fig.~\ref{fig:SMetzlerShea}) based on the various cognitive processes required by their coffee robot waiter.  This taxonomy is also defined as a resource taxonomy since it captures required capabilities rather than jobs to be done or properties of jobs to be done.  
 
Although it is a purely branch selection structure and terminates in category tips like Fishwick, its internal structure is significantly more complex.  Instead of a single layer of intermediate categories, this taxonomy has a variable number of decisions at each node, and incorporates two layers of intermediate categories in some areas (e.g. to act $\Rightarrow$ to interact $\Rightarrow$ to communicate) and only one in others (e.g. to act $\Rightarrow$ to move).  In some places, more detailed concepts are captured below more abstract concepts, implying that the taxonomy may be able to be terminated at any level of abstraction --- the cognitive processes for learning include but are not limited to the cognitive processes for memorization or studying, so the required cognitive capability may be better expressed as ``learning'' rather than as ``memorizing''.

% Ab.Acus robot type taxonomy 3 (repositioned for layout reasons)
\begin{figure*}[t]
\begin{center}
\includegraphics[width=4.5in, viewport=156.25in 22.8in 176in 39.8in, clip=true]{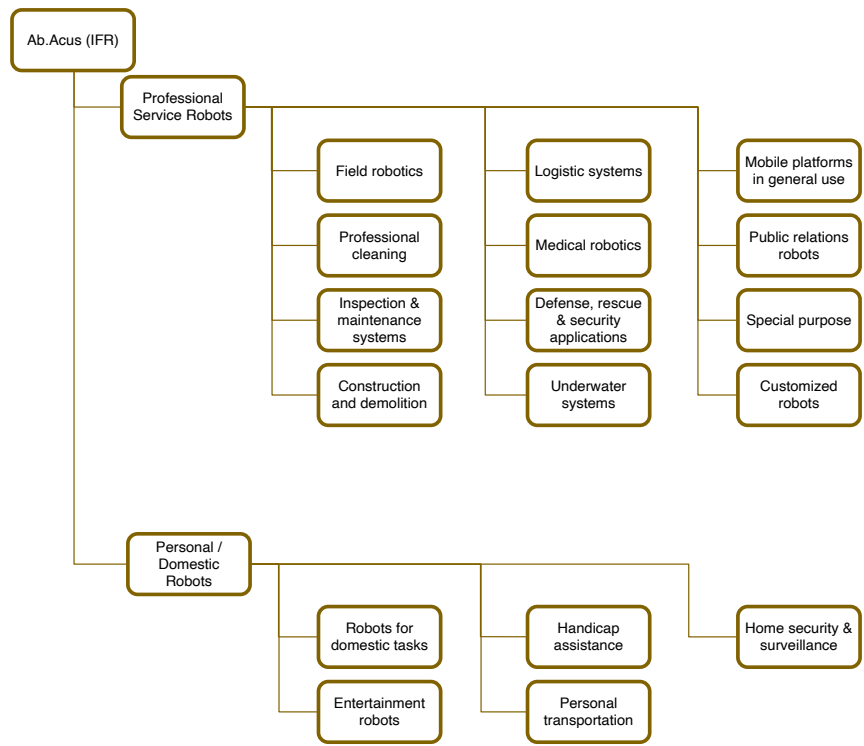}
\caption{Ab.Acus's taxonomy of robot types from the International Federation of Robotics}
\label{fig:SAbAcus_3}
\end{center}
\end{figure*}
% Robin and Lacroix's taxonomy
\begin{figure*}[t]
\begin{center}
\includegraphics[width=5.5in, viewport=41.9in 51.2in 64in 61.2in, clip=true]{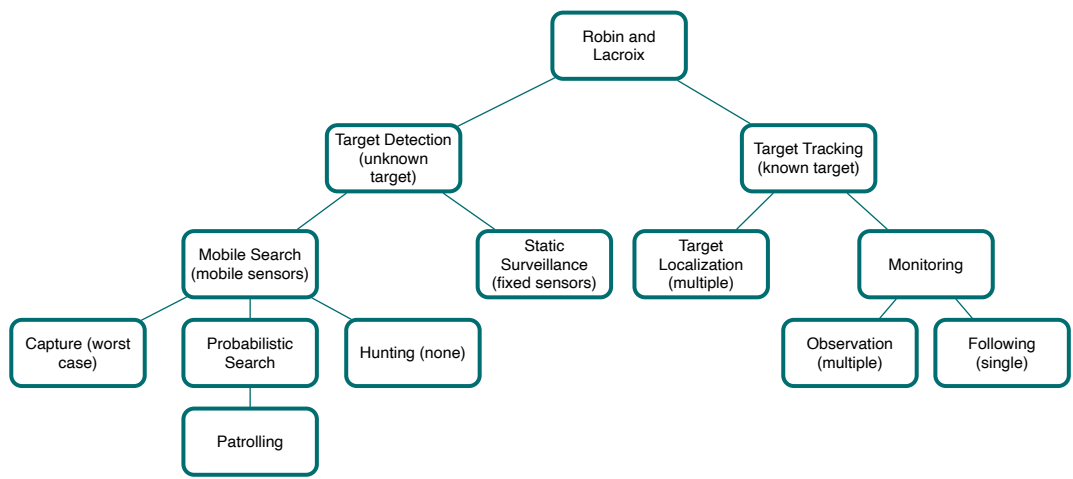}
\caption{Robin and Lacroix's taxonomy of target detection and tracking tasks.}
\label{fig:TDRobinLacroix}
\end{center}
\end{figure*}

\subsubsection{\cite{AbAcus17}}
An unpublished taxonomy document attributed to Ab.Acus identifies three different robot taxonomies.  Two of these are of the branch select form; both categorize robots according to the types of tasks they are designed to accomplish.  Because they categorize robots rather than tasks, they are captured here as resource taxonomies.  Both of these taxonomies have category tips rather than assign value tips.

The taxonomy shown in Fig.~\ref{fig:SAbAcus_2}, which they drew from \cite{Siciliano09} \underline{Robotics Modelling, Planning and Control}, is another extremely simple taxonomy, defining categories of robot tasks at a very high level.  The {\em Industrial} intermediate category captures tasks that occur in highly regulated environments and categorizes them based on specific types of industrial jobs related to the task goals (e.g. manipulation, measurement).  

The {\em Advanced} intermediate category, on the other hand, assumes an unstructured environment and further divides these tasks into a field category, covering tasks with unsafe environments, and a service category, covering tasks where the goal is to improve quality of life.

The taxonomy shown in Fig.~\ref{fig:SAbAcus_3}, drawn from the International Federation of Robotics' website in 2015, captures a more detailed categorization of robots according to the task they are designed for.  This is also a simple taxonomy, with only only two intermediate categories and only three nodes, but it has many more choices per node than their other taxonomy.  Integrating these two taxonomies would be difficult, even though they address categorization along the same axis, because they capture two different concepts with the same words.  Both taxonomies use the word ``service,'' but one uses it to identify robots that perform quality-of-life related tasks, while the other uses it to differentiate between robots used in an industrial setting and robots used in any other setting.  It is not clear whether they are using ``field'' to refer to the same underlying concept or not, but in one, it references a high level categorization that distinguishes it from service robots, while in the other, it is a subset of service robots.  

As we integrate the concepts from these individual taxonomies to develop the larger taxonomy, it will be important to ensure that the concepts from each are clearly defined and any inconsistencies are addressed, with new terminology assigned as needed to ensure that the various types of systems and approaches to distinguishing between them are included.

%%% Korsah et al. Figure
\begin{figure*}[t]
\begin{center}
\includegraphics[width=5.5in, viewport=111.5in 5.3in 136.9in 23.7in, clip=true]{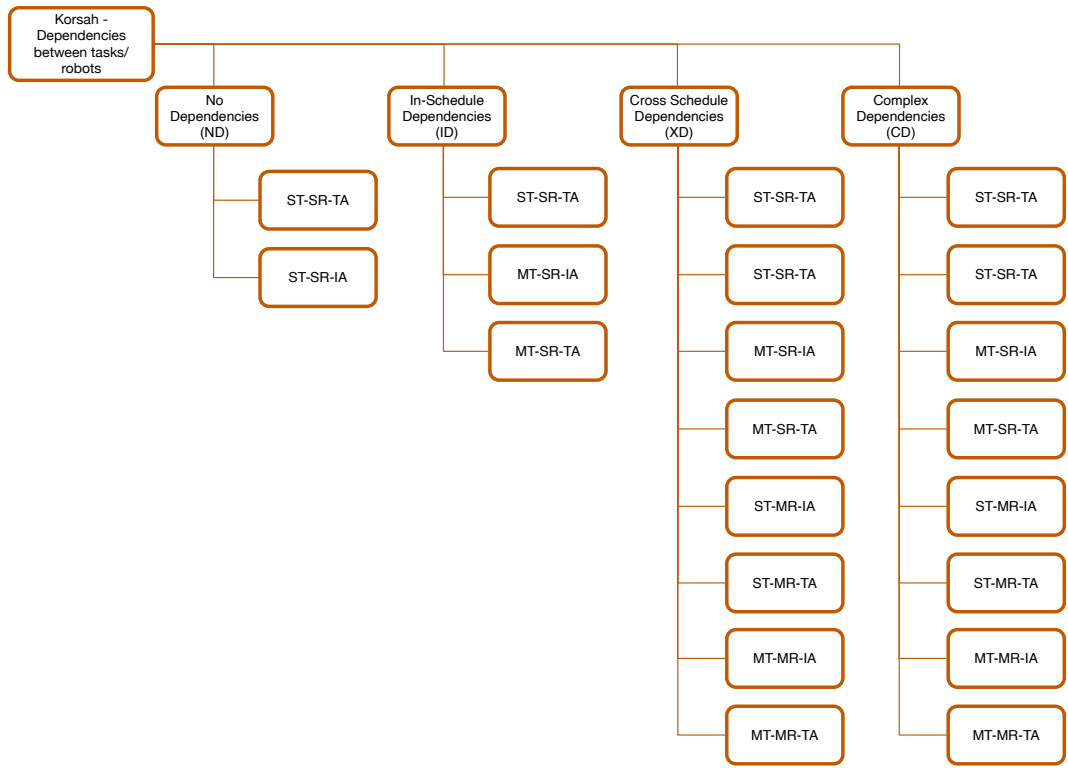}
\caption{Korsah et al.'s multi-robot systems taxonomy.}
\label{fig:MRSKorash}
\end{center}
\end{figure*}

\subsubsection{\cite{Robin16}}
\label{sssec:robin}
Robin and Lacroix's taxonomy of target detection and tracking tasks (shown in Fig.~\ref{fig:TDRobinLacroix}) is particularly challenging to categorize into task, task property, or resources.  While the Target Detection vs. Target Tracking categories clearly define different task types, the Mobile Search vs. Static Surveillance categories are based on available resource types and the Target Localization vs. Monitoring categories require knowledge of the number of targets and the number of viewpoints from which they may be observed (task properties).  Fundamentally, the purpose of this taxonomy is to define task type instances in the context of these other types of information, so it is categorized as a task type taxonomy.

It has a purely branch selection structure, and includes both high level and detailed terminating categories.  To separate the task type categories from the resources and task properties decision criteria, each category is described below.
\begin{itemize}
\item{{\em Target Management node (root)}:  This node defines selection on the basis of resource properties, where ``known'' implies a previously defined target type while ``unknown'' indicates that the target type has not been defined as of interest
 \begin{itemize}
 \item{{\em Target Detection}: This task goal category is determined by resources, and the node below it selects among the task types necessary to accomplish this goal. The goal is focused on the detection of a specific type of target.
   \begin{itemize}
   \item{{\em Mobile Search}:  This task type category is broken down into the various subtasks that can be activated during a target detection operation using mobile sensors
     \begin{itemize}
     \item{{\em Capture (worst case)}:  This task type is a category tip describing a mobile search task against a known target that terminates in a {\em Capture} event}
     \item{{\em Probabilistic Search}:  This is a high level cyclic capability, with no defined start and end point --- it continues operating until some external event intervenes.  This task type is defined in part by the existence of some guarantees of finding the target.  
       \begin{itemize}
       \item{{\em Patrolling}:  This is a category tip for a specific type of {\em Probabilistic Search}.  Others could be defined but are not specified as part of this taxonomy.}
       \end{itemize}}
     \item{{\em Hunting (none)}:  This is another mobile search task type that involves searching for an object in cases where there is no guarantee that the object exists in the search area}
     \end{itemize}}
   \item{{\em Static Surveillance}:  This section addresses the various task types that could be activated during a target detection operation using fixed sensors; however, no sub-categories are defined, so this becomes a category tip.}
   \end{itemize}}
 \item{{\em Target Tracking}:  This intermediate category enables classification of the different capabilities that can be used when a known target's location over time is the desired information.  The node below this category defines branches based on what resources are available to gather this information.  
   \begin{itemize}
   \item{{\em Target Localization}:  This category tip terminates the instance with a task involving localization of a target through multiple points of view.}
   \item{{\em Monitoring}:  This intermediate category addresses the case where there is only one viewpoint of one or more targets
     \begin{itemize}
     \item{{\em Observation (multiple)}:  This category covers the situation where there are multiple targets but only one point of view}
     \item{{\em Following (single)}:  This category covers the situation where there is one target and one point of view}
     \end{itemize}}
   \end{itemize}}
 \end{itemize}}
\end{itemize}

% Facet Only Diagram figure (repositioned to support layout considerations)
\begin{figure*}[t]
\begin{center}
\includegraphics[width=6in, viewport=138.2in 130.5in 163in 138.9in, clip=true]{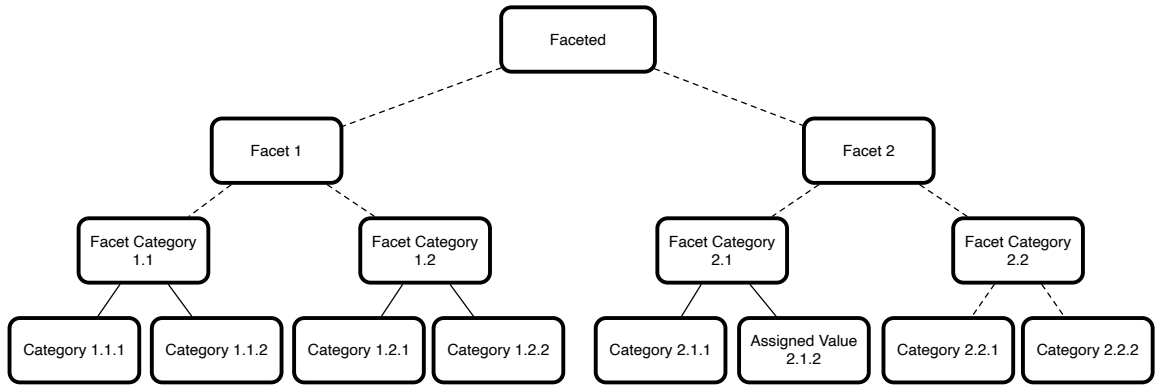}
\caption{Basic structure of facet only taxonomies.}
\label{fig:FacetOnly}
\end{center}
\end{figure*}
% Facet Only Diagram figure
% Gerkey and Mataric
\begin{figure*}[t]
\begin{center}
\includegraphics[width=5in, viewport=40.4in 67.5in 59.8in 73.3in, clip=true]{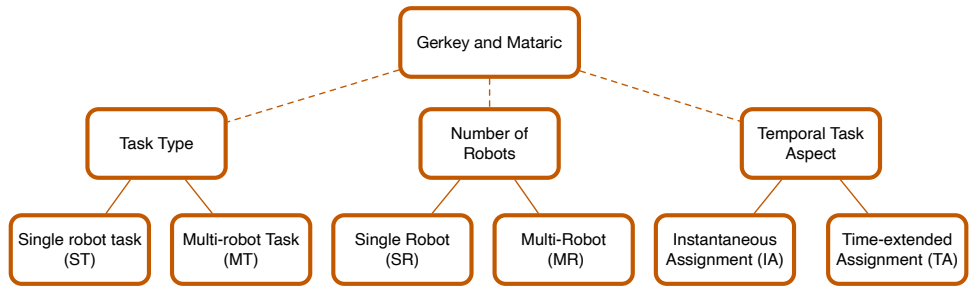}
\caption{Gerkey and Matari\'{c}'s multi-robot systems taxonomy.}
\label{fig:MRSGerkey}
\end{center}
\end{figure*}

\subsubsection{\cite{Korsah13}}
\label{sssec:korsah}
The iTax taxonomy developed by Korsah, et al. builds on the faceted taxonomy developed by  \cite{Gerkey04} (described in Section~\ref{sssec:gerkey}) to incorporate the degree of independence of agent-task utilities. They introduce a top-level category capturing dependency, with specific instances of Gerkey and Matari\'{c}'s taxonomy of robot type defining the assign value tips at the second level. This dependency category may apply to the system capabilities or to the task implementation, but fundamentally it captures the degree of interrelatedness (either required or provided) between the agents engaged in a specific task.  This taxonomy is illustrated in Fig.~\ref{fig:MRSKorash}.

Looking at the individual intermediate categories, we can see that determining the content type of this taxonomy is challenging.
\begin{itemize}
\item{Degree of interrelatedness
  \begin{itemize}
  \item{{\em No dependencies (ND)}:  How well an agent can perform the task is independent of both the status of this or any other agent and of any other tasks.}
  \item{{\em In-Schedule dependencies (ID)}:  How well an agent can perform the task depends on what other tasks it is performing --- task performance is affected by the status of this agent but not by the status of other agents.}
  \item{{\em Cross-Schedule dependencies (XD)}:  How well an agent can perform the task depends on the status of this agent and other agents in the system.}
  \item{{\em Complex dependencies (CD)}:  How well an agent can perform the task depends on the status of this agent, the status of the other agents, and the specific task decomposition chosen.}
  \end{itemize}}
\end{itemize}
Each instance is defined in terms of the relationship between the resource and the task.  The actual thing being categorized depends on the focus within which the taxonomy is used.  If the taxonomy is being used to classify robot teams in terms of their performance, then it is a resource taxonomy, defining a robot team property.  If the taxonomy is being used in a task context, then it is a task property taxonomy, defining the task in terms of what kind of team coordination is necessary to accomplish it.

% Tan et al.'s taxonomy
\begin{figure}[t]
\begin{center}
\includegraphics[width=3in, viewport=45.25in 42in 60in 49.25in, clip=true]{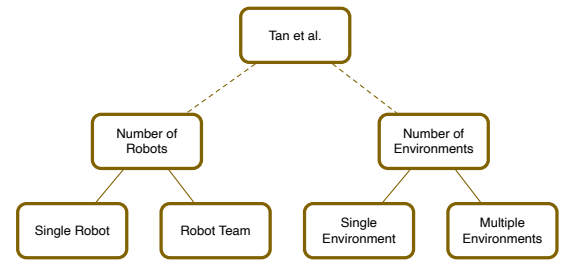}
\caption{Tan et al.'s taxonomy of robot-environment relationships.}
\label{fig:STan}
\end{center}
\end{figure}

%% Zech Top
\begin{figure}[t]
\begin{center}
\includegraphics[width=3in, viewport=103.2in 101.7in 118.9in 108.4in, clip=true]{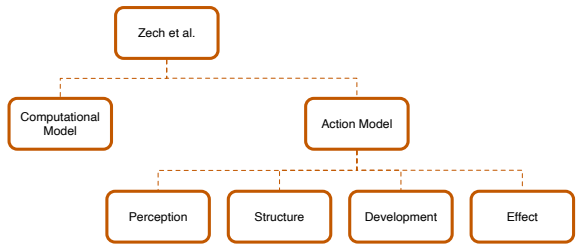}
\caption{Zech et al.'s taxonomy of cognitive functions --- Top level facet structure.}
\label{fig:SZech_0}
\end{center}
\end{figure}
% Zech Computation
\begin{figure}[t]
\begin{center}
\includegraphics[width=3in, viewport=117.3in 126.7in 133in 147.8in, clip=true]{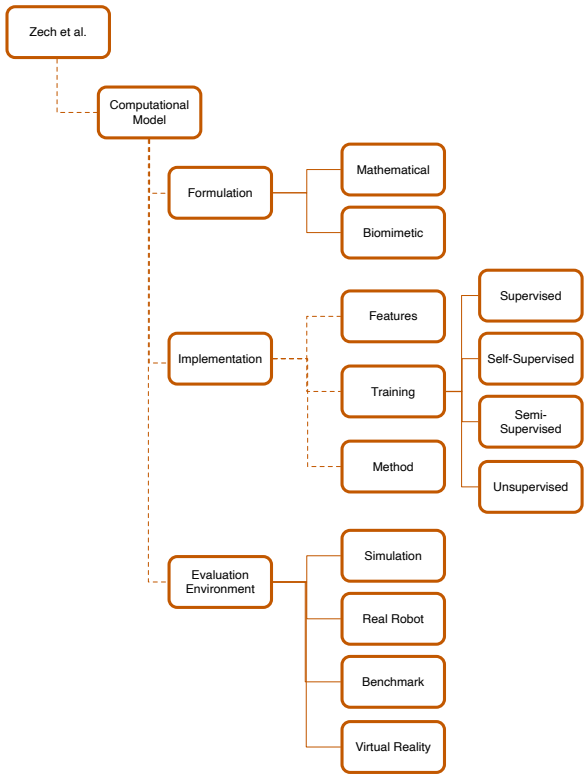}
\caption{Zech et al.'s taxonomy of cognitive functions --- Computational Model facet.}
\label{fig:SZech_1}
\end{center}
\end{figure}
%% Zech Action Perception
\begin{figure}[t]
\begin{center}
\includegraphics[width=3in, viewport=118.2in 105.9in 133in 122.9in, clip=true]{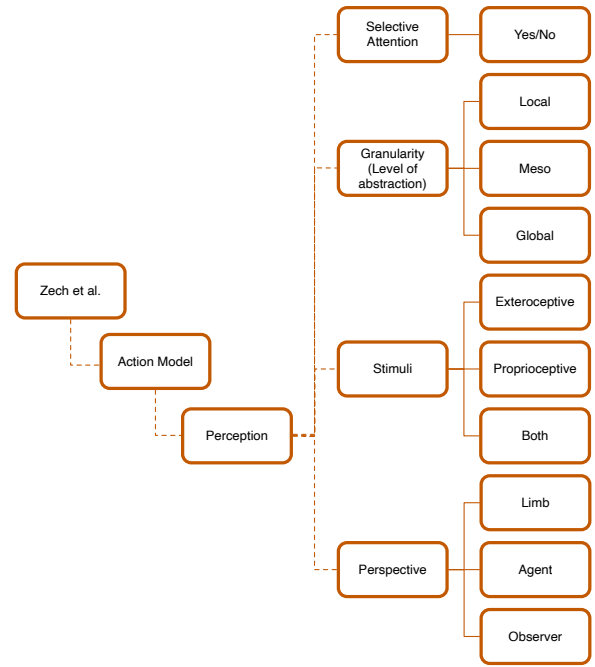}
\caption{Zech et al.'s taxonomy of cognitive functions --- Action Model facet, Perception category.}
\label{fig:SZech_2}
\end{center}
\end{figure}
%% Zech Action Structure
\begin{figure}[t]
\begin{center}
\includegraphics[width=3in, viewport=117.6in 85.4in 133in 96in, clip=true]{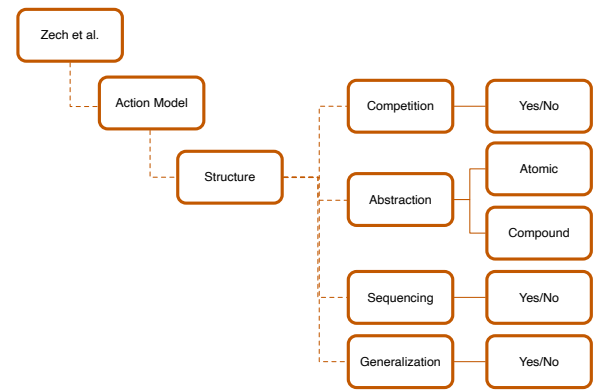}
\caption{Zech et al.'s taxonomy of cognitive functions --- Action Model facet, Structure category.}
\label{fig:SZech_3}
\end{center}
\end{figure}

%%%%%%%    FACET ONLY    %%%%%%%%
\subsection{Facet Only Taxonomies}
\label{ssec:facetOnly}
The facet only taxonomies have a faceted structure down to the final intermediate categories.  Where tips are defined, the final selection of terminating element may use a branch select structure, but everything above that is faceted.  They may use category or assign value tips to define a specific instance, but do not contain branch select structures whose children are intermediate categories.  Where the final selection mechanism is faceted, assign value tips imply a branch selection decision regarding the value assigned, while category tips imply the existence of further detail leading to a branch selection mechanism.
 
In Fig.~\ref{fig:FacetOnly}, one instance of this taxonomy (using colons to demarcate facet boundaries and square brackets to demarcate collections of facets) would be defined as [Category 1.1.1 : Category 1.2.2 : Assigned Value 2.1.2 : [Category 2.2.1 : Category 2.2.2]].  As before, faceted nodes are represented with dashed lines while branching nodes are represented with solid lines.  

Facet-only taxonomies are distinct from the facet-over-branch taxonomies in Section~\ref{ssec:facetOverBranch}, where there are branch selection decision points that lead to intermediate categories above the tips.

\subsubsection{\citet{Gerkey04}}
\label{sssec:gerkey}
Gerkey and Matari\'{c}'s simple taxonomy of multi-robot systems (shown in Fig.~\ref{fig:MRSGerkey}) has three facets, capturing single-task robots (SR) vs. multiple-task robots (MR), single robot tasks (ST) vs. multiple-robot tasks (MT), and instantaneous assignment (IA) vs time-extended assignment (TA) in category tips.  Although it is defined as a resource taxonomy by the authors, this incorporates mixed type facets, since the Task Type facet and the Temporal Task Aspect facet are both task property categories, while the Number of Robots facet is a resource property category.  

\subsubsection{\citet{Tan16}}
The root concern of Tan et al.'s faceted task type taxonomy is the relationship between robot groups and environments.  One facet addresses the number of robots and the other addresses the number of environments, producing the taxonomic structure shown in Fig.~\ref{fig:STan}.  This structure captures the associated 2x2 array of possible combinations of single and multiple robots and environments.

For example, the [single robot : single environment] instance describes a task type where there is only one robot and it is expected to operate in only one environment (e.g. a robot vacuum in a house).  Similarly, the [robot team : single environment] instance would describe a task that specifies a team of robot package handlers in a warehouse.  A task involving a robot vacuum and a robot window washer in the same house would be an example of a [robot team : single environment] task, and a single robot vacuum used in both a house and a workshop would be categorized as a [single robot : multiple environments] task type.  Two robots coordinating to clean several different rooms, with the robots operating in parallel in different rooms, would be an example of a [multiple robots : multiple environments] task.

From a structural perspective, this taxonomy is very simple, capable of representing only 4 different instances and capturing one very abstract or high level aspect of task categorization.

\subsubsection{\citet{Zech17, Zech19}}
Zech et al. have developed a comprehensive taxonomy for the classification of action representations.  The final taxonomy is very large and is separated here into figures by facet at the top two levels.  The top two levels of facets are illustrated in Fig.~\ref{fig:SZech_0}.

Effectively, this taxonomy as a whole focuses on the description and categorization of a specific defined action as provided, rather than on the properties of a specific task to which that action might be applied, and is therefore a resource property taxonomy.  This holds across the sub-facets of the taxonomy with the exception of the Computational Model $\Rightarrow$ Evaluation Environment facet, which captures resources rather than resource properties.  Because we are lumping resources and resource properties into one group, this entire taxonomy fits cleanly into our resources group.

The top layer contains two facets: the Computational Model and the Action Model.  The Computational Model assigns categories related to the design principles, implementation mechanisms, and evaluation environments for the action, while the Action Model categorizes actions based on design criteria.  These design criteria include the properties of the structures used to represent the action, the development mechanism properties, the properties of the perception system, and effects that can be applied to the action mechanism.

%% Zech Action Effect
\begin{figure}[t]
\begin{center}
\includegraphics[width=3in, viewport=118.2in 27.5in 133in 43.2in, clip=true]{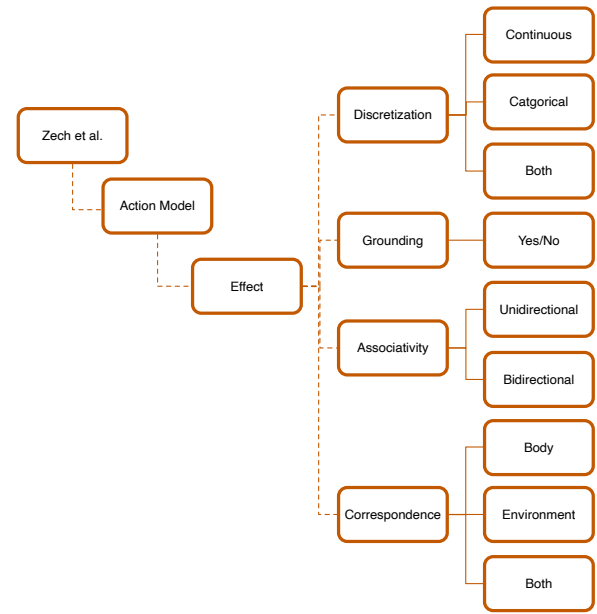}
\caption{Zech et al.'s taxonomy of cognitive functions --- Action Model facet, Effect category.}
\label{fig:SZech_5}
\end{center}
\end{figure}
%% Zech Action Development
\begin{figure}[t]
\begin{center}
\includegraphics[width=3in, viewport=116.6in 46.3in 133.1in 80.2in, clip=true]{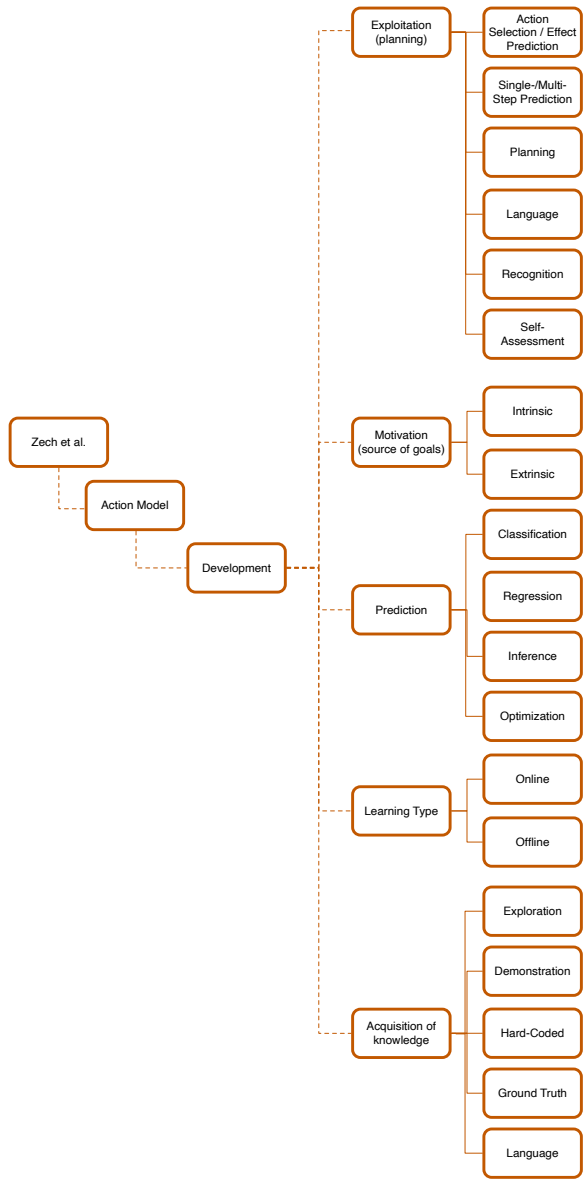}
\caption{Zech et al.'s taxonomy of cognitive functions --- Action Model facet, Development category.}
\label{fig:SZech_4}
\end{center}
\end{figure}

% Tsiakis et al
\begin{figure*}[t]
\begin{center}
\includegraphics[width=6in, viewport=1.0in 0in 33.5in 49.1in, clip=true]{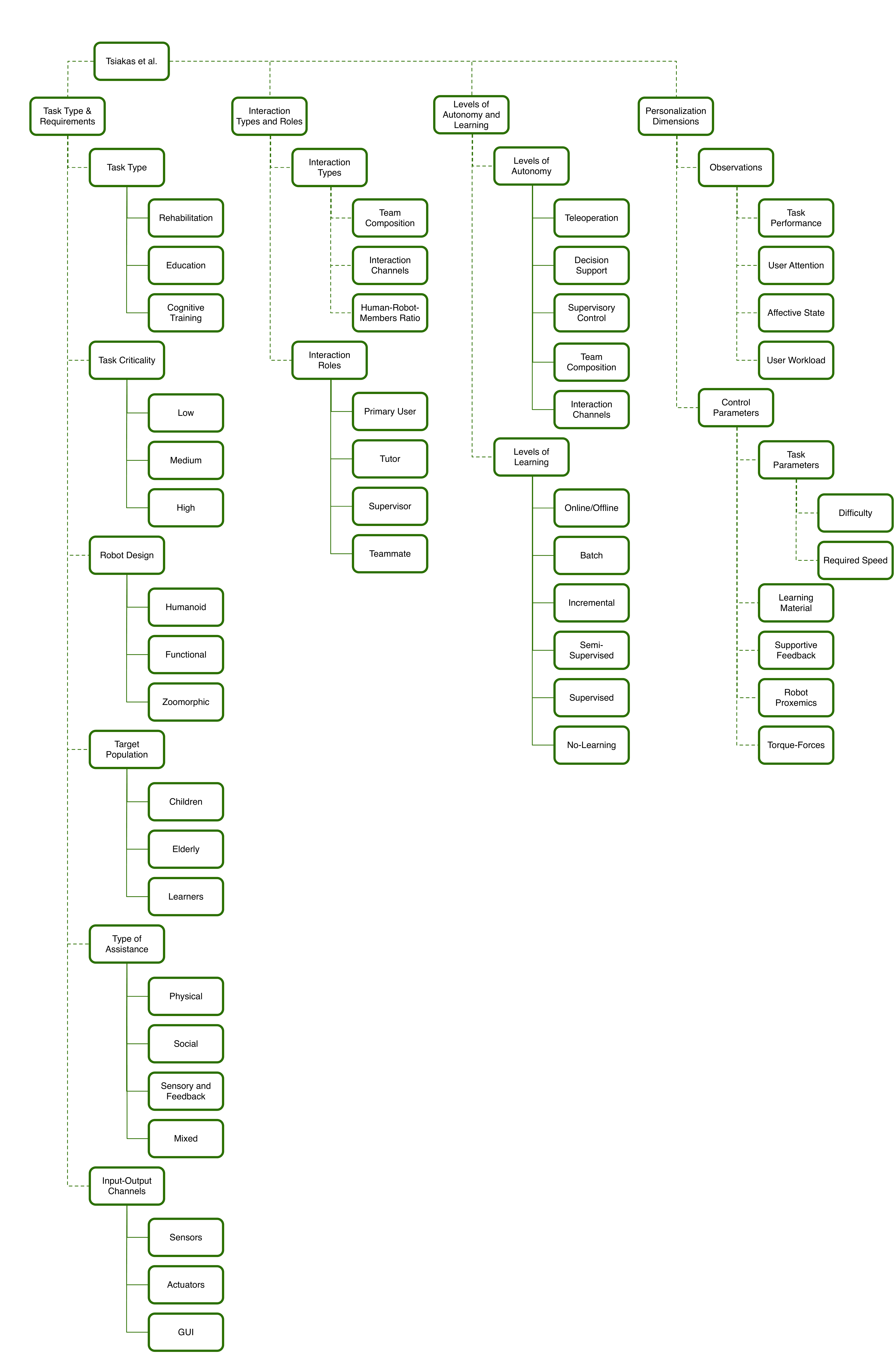}
\caption{Tsiakas et al.'s taxonomy of robot-assisted training systems.}
\label{fig:HRITsiakas}
\end{center}
\end{figure*}

While the taxonomy itself primarily uses category tips, specific potential assigned values (e.g. Yes/No) are identified for some of the categories.

This taxonomy illustrates the case where the category tips are themselves faceted.  It is assumed that there are lower level branch select category or assign value tips below the {\em Features} and {\em Method} facets in the {\em Computational Model} facet shown in Fig.~\ref{fig:SZech_1}, but as these are undefined, each faceted category tip only contributes a single instance to the complexity metrics.  

The result is that while this taxonomy has the highest structural complexity (as discussed in Section~\ref{ssec:structural}) and the highest representational complexity (as discussed in Section~\ref{ssec:representational}) of all the reviewed taxonomies, the metrics are actually undercounting the complexity this taxonomy is capable of representing.

\subsubsection{\citet{Tsiakas18}}
\label{sssec:tsiakis}
Tsiakas et al.'s taxonomy of robot-assisted training systems is shown in Fig.~\ref{fig:HRITsiakas}.  This is a purely faceted taxonomy, and the Task Type and Relationships and Levels of Autonomy and Learning facets end in branch selection category tips.  However, in this taxonomy some facets end with a faceted tip selection mechanism that could connect to assigned value options or to other taxonomies.  The Interaction Types and Roles $\Rightarrow$ Interaction Types $\Rightarrow$ Team Composition facet could lead to Dudek's (Section~\ref{sssec:dudek2}) or Farinelli's (Section~\ref{sssec:farinelli}) Team Composition intermediate categories, while the Personalization $\Rightarrow$ Control Parameters $\Rightarrow$ Task Parameters facet points to assign value tips Difficulty and Speed. 

Furthermore, all three types of content are mixed within a single facet.  As with Gerkey and Matari\'{c}'s taxonomy (Section~\ref{sssec:gerkey}), although the taxonomy as a whole is intended to categorize specific resources (robot-assisted training systems), the Task Type and Requirements facet mixes resources (e.g. {\em Robot Design}), tasks (e.g. {\em Task Type}), and task properties (e.g. {\em Task Criticality}) as individual facets within a facet.  

%% Beer et al.
\begin{figure*}[t]
\begin{center}
\includegraphics[width=5.5in, viewport=67.7in 5.4in 95.7in 11.2in, clip=true]{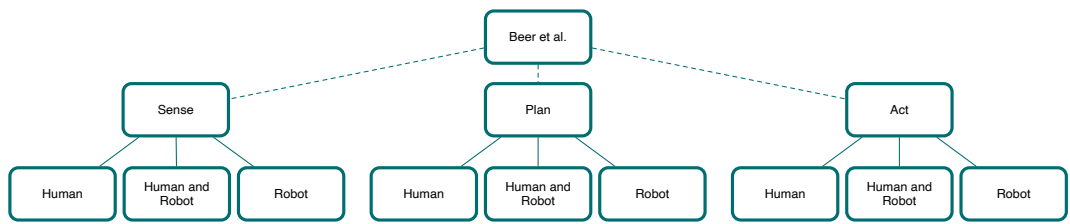}
\caption{Beer et al.'s levels of autonomy structured by human, human-robot, or robot sensing, planning, and acting.}
\label{fig:HRIBeer}
\end{center}
\end{figure*}

\subsubsection{\citet{Beer14}}
\label{sssec:beer}
Beer et al.'s levels of autonomy for human-robot interaction (LORA) is another faceted resource taxonomy.  In this case, it categorizes the combination of autonomy design element and assumed responsibility aspects of a system in the service of categorizing levels of autonomy.  It assumes a Sense-Plan-Act autonomy design and combines this with categorization of who holds the responsibility for decisions and outcomes.  This taxonomy, as shown in Fig.~\ref{fig:HRIBeer}, is represented as a two tier structure with a faceted Sense, Plan, Act top layer, each of which has a branch selection structure containing subcategories of Human (H), Robot (R), and Shared (H/R) responsibility.  Each level of autonomy, below, represents an instance of the taxonomy. 
\begin{itemize}
\item{{\em Manual}:  Sense-H, Plan-H, Act-H}
\item{{\em Teleoperation}:  Sense-H/R, Plan-H, Act-H/R}
\item{{\em Assisted Teleoperation}:  Sense-H/R, Plan-H, Act-H/R (this includes reactive obstacle avoidance)}
\item{{\em Batch Processing}:  Sense-H/R, Plan-H, Act-R (human determines goals and defines plan; robot implements plan)}
\item{{\em Decision Support}:  Sense-H/R, Plan-H/R, Act-R (robot and human generate plans; human selects plan)}
\item{{\em Shared Control with Human Initiative}:  Sense-H/R, Plan-H/R, Act-R (supervised --- human monitoring)}
\item{{\em Shared Control with Robot Initiative}:  Sense-H/R, Plan-H/R, Act-R (robot asks for help)}
\item{{\em Executive Control}:  Sense-R, Plan-H/R, Act-R (human provides goal; robot acts autonomously to achieve goal)}
\item{{\em Supervisory Control}:  Sense-H/R, Plan-R, Act-R (human continuously monitors and may override at any time)}
\item{{\em Full Autonomy}:  Sense-R, Plan-R, Act-R (no human intervention)}
\end{itemize}
Although this list has ten levels of autonomy to be categorized, Fig.~\ref{fig:HRIBeer} shows only nine categories.  This is because the two Shared Control levels have the same encoding in the taxonomy and are represented by the same instance.  In order to differentiate them, it would be necessary to add another facet representing whether one agent imposes help or the other requests help.

\begin{figure}[t]
\begin{center}
\includegraphics[width=3in, viewport=166in 114.9in 181.6in 123.9in, clip=true]{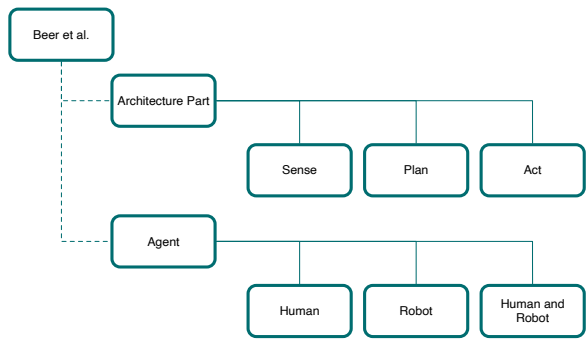}
\caption{Beer et al.'s levels of autonomy --- incorrect faceted structure.}
\label{fig:HRIBeer_3}
\end{center}
\end{figure}

Given the redundancy in the lower levels, it is tempting to think that Fig.~\ref{fig:HRIBeer_3} would provide a more efficient facet-only approach to this structure, but since a valid level of autonomy in this framework requires a matched Agent element for each Architecture Part facet, it is incorrect.  The faceted taxonomy shown in Fig.~\ref{fig:HRIBeer_3} generates invalid instances of the form [Sense : Human], not correct instances of the form [Sense-Human : Plan-Robot : Act-Human].  
%%% Beer et al. figure
%%% Shim and Arkin figure
\begin{figure*}[t]
\begin{center}
\includegraphics[width=4.5in, viewport=1in 127.4in 19.9in 133.3in, clip=true]{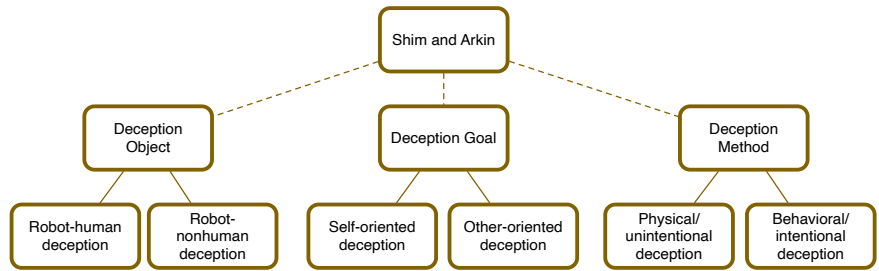}
\caption{Shim and Arkin's taxonomy of deception tasks.}
\label{fig:HRIShimArkin}
\end{center}
\end{figure*}
%%% Jiang and Arkin figure
\begin{figure*}[t]
\begin{center}
\includegraphics[width=5.5in, viewport=23.4in 121.6in 45.6in 133.3in, clip=true]{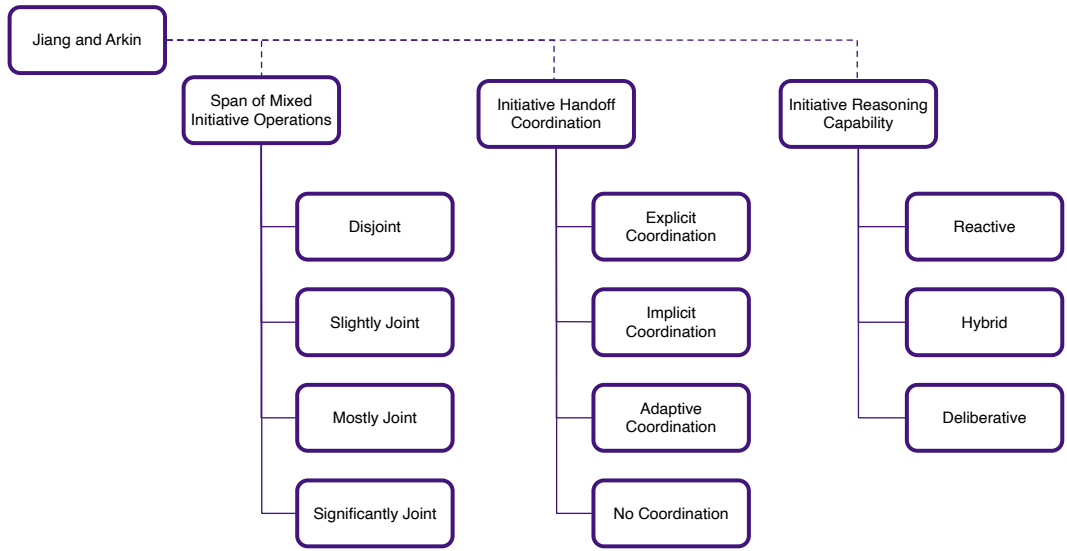}
\caption{Jiang and Arkin's taxonomy of mixed-initiative human-robot interactions.}
\label{fig:HRIJiangArkin}
\end{center}
\end{figure*}

%%%%%%%  
\subsubsection{\citet{Shim13}}
\label{sssec:shim}
Shim and Arkin's taxonomy categorizing tasks requiring deception identifies three main facets:  deception object, goal, and method.  Within the facets, the deception object and deception goal categories contribute to the definition of task goals, while the deception method categories define a resource taxonomy of behavior descriptors.  This taxonomy (illustrated in Fig.~\ref{fig:HRIShimArkin}), like many of the taxonomies in this section, has a very simple structure, with a single layer of intermediate categories between the root and the category tips.  Because this taxonomy, like many of the taxonomies in this section, has only one layer of intermediate categories, it is impossible to know whether this is inherently a facet-over-branch structure, where all intervening layers would consist of branch select nodes, or a mixed type III structure, where subsequent nodes could include both branch select and faceted nodes.

As we can see in the following descriptions, the taxonomy is limited to categorizing the context in which the deception is practiced and the high level method used to accomplish it, making this a task type taxonomy:
\begin{itemize}
\item{Deception Object:
  \begin{itemize}
  \item{Robot-human deception: the robot is attempting to deceive a human}
  \item{Robot-nonhuman deception:  the robot is attempting to deceive a nonhuman (other robot, animal, etc.)}
  \end{itemize}}
\item{Deception Goal:
  \begin{itemize}
  \item{Self-oriented deception: deception is practiced for the robot's benefit}
  \item{Other-oriented deception: deception is practiced for some other's benefit}
  \end{itemize}}
\item{Deception Method:
  \begin{itemize}
  \item{Physical/unintentional deception:  deception is instantiated through embodiment or low cognitive or behavioral complexity}
  \item{Behavioral/intentional deception:  deception is instantiated through mental representations and behaviors}
  \end{itemize}}
\end{itemize}
This taxonomy doesn't capture resource information related to things that the robot might want to be deceptive about, like goals, position/status, or environmental state.

Interestingly, the Deception Object category only differentiates between human and non-human targets, not between teammates and external agents or sensors and thinking agents.  This facet could clearly be expanded to address more subtle differences in the targets of deception, since the approaches suitable for deception of a non-human target like a dog will be very different from the approaches suitable for a non-human target like an AI perception algorithm.

\subsubsection{\cite{Jiang15}}
\label{sssec:jiang}
The taxonomy of mixed-initiative human-robot interactions developed by Jiang and Arkin (Fig.~\ref{fig:HRIJiangArkin}) has a faceted node at the top level and branch selection nodes just above the tips.  While this taxonomy may appear to be a task property taxonomy, capturing aspects of tasks, it is actually defined to categories the mixed initiative mechanisms used by the members of the team to interact across robot/human boundaries.  As shown below, it is fundamentally capturing properties of how the team works, and is therefore a resource taxonomy:
\begin{itemize}
\item{Span of Mixed-Initiative Facet: degree to which tasks are initiated by different types of team members - these are evaluated based on how many of the three mission initiative elements are joint (the three mission initiative elements are goal setting, planning, and execution).
  \begin{itemize} 
  \item{Disjoint:  robot and human are not authorized to intervene or take control from each other}
  \item{Slightly Joint: only one element of the mission supports intervention from either the human or the robot}
  \item{Mostly Joint: two elements of the mission support intervention from either the human or the robot}
  \item{Significantly Joint:  all three mission initiatives support intervention from either the human or the robot}
  \end{itemize}}
\item{Initiative Reasoning Capacity Facet: when intervention or change of control can occur; covers both interleaved interaction (when the humans and robots work simultaneously)  and opportunistic interaction (when the humans and robots take turns working on the task)
  \begin{itemize}
  \item{Reactive:  decisions are made based on incoming sensor data}
  \item{Deliberative:  decisions are made based on deliberations around the status of a world model and how it compares to a desired world model}
  \item{Hybrid:  some decisions are Reactive and some decisions are Deliberative}
  \end{itemize}}
\item{Initiative Handoff Coordination Facet: how to determine when a change in control is warranted and accepted
  \begin{itemize}
  \item{Explicit Coordination:  human and robot explicitly state and agree to changes in control authority}
  \item{Implicit Coordination:  human and robot infer when changes in control authority have or should happen}
  \item{Adaptive Coordination:  different coordination mechanisms are used at different times}
  \item{No Coordination:  no changes in control authority occur}
  \end{itemize}}
\end{itemize}
Because of the way the category tips are defined, this taxonomy explicitly defines the category tips as tips and cannot be expanded into more layers.

%%% Dudek et al. 2 figure
\begin{figure*}[t]
\begin{center}
\includegraphics[width=5.5in, viewport=26in 98.9in 54in 105.8in, clip=true]{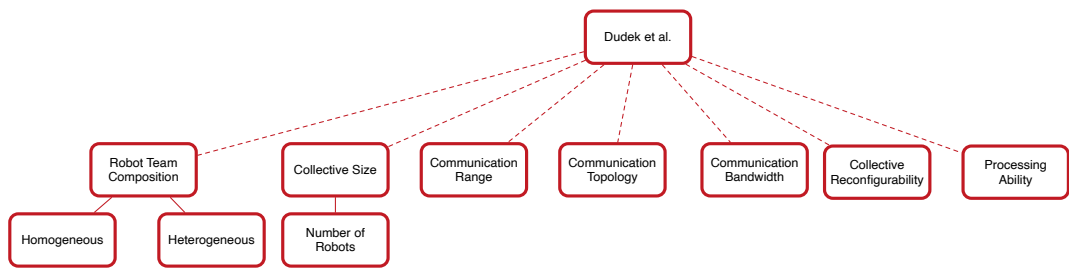}
\caption{Dudek et al.'s taxonomies of multi-agent system configuration properties.}
\label{fig:MRSDudekSystem}
\end{center}
\end{figure*}
%%% Farinelli et al. figure
\begin{figure*}[t]
\begin{center}
\includegraphics[width=6.5in, viewport=24.2in 78.5in 56.5in 97.9in, clip=true]{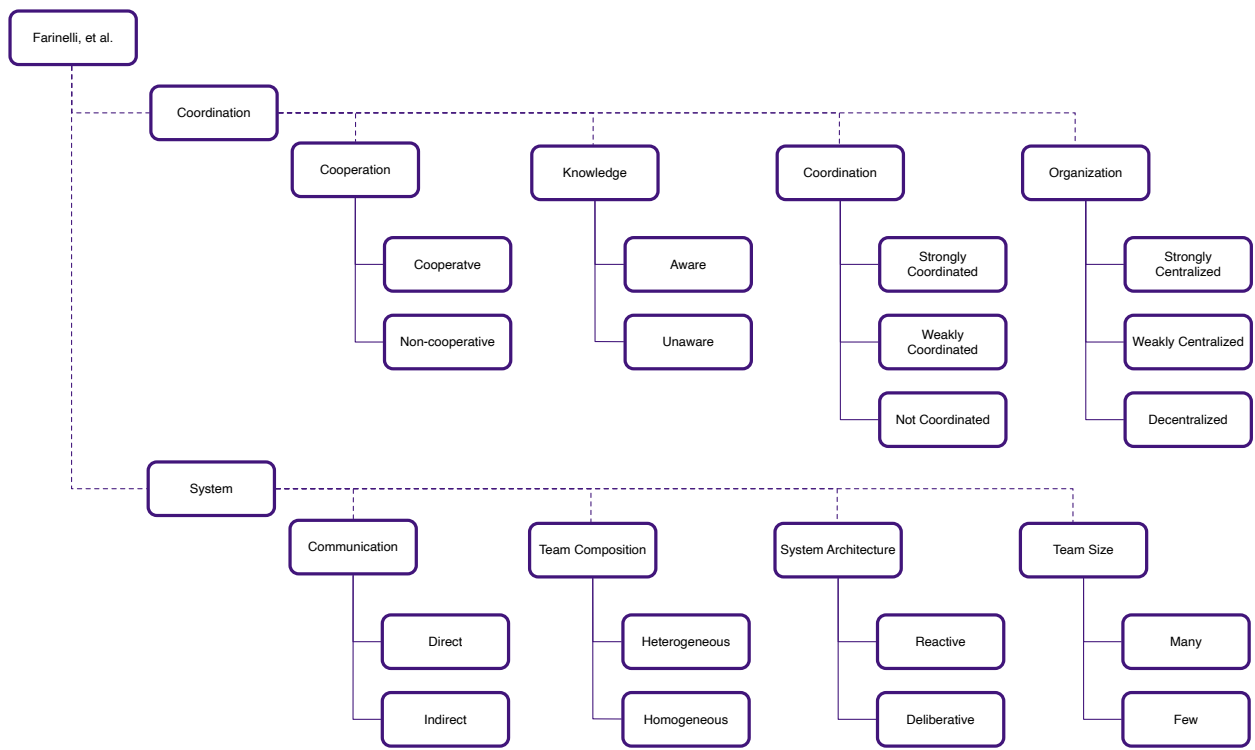}
\caption{Farinelli et al.'s taxonomy of coordination and system properties.}
\label{fig:MRSFarinelli}
\end{center}
\end{figure*}

\subsubsection{\cite{Dudek96}}
\label{sssec:dudek2}
The robot collective taxonomy from Dudek et al., shown in Fig.~\ref{fig:MRSDudekSystem}, addresses the different aspects of group operation of robotic systems.  It constructs eight categories and defines the values they may take.  This is a faceted resource taxonomy with some assigned value tips and some categorization tips.   The eight categories are 
\begin{itemize}
\item{Collective Size:  value of how many agents are in the group}
\item{Communication Range:  value of how large a distance they can communicate across}
\item{Communication Topology:  value of which robots can be communicated with}
\item{Communication Bandwidth:  value of how much information can be communicated}
\item{Collective Reconfigurability:  value of the rate at which the collective's organization can change}
\item{Processing Ability:  value defined as the computational model used by the individual agents in the collective --- note that this a value selector in the same way as Yanco and Drury's {\em available sensor information} category}
\item{Collective Composition:  branch selection of heterogeneity/homogeneity categories}
\end{itemize}
Because of this taxonomy's use of facets as assign value tips for most categories, the resulting number of instances it can represent in our representational complexity metric is low relative to its actual potential representational complexity.  If each facet defined lower level branch select categories, such as "Near", "Mid", and "Far" for the Communication Range facet, the number of potential instances would grow exponentially.  Complexity metrics for taxonomies containing faceted tips are addressed in more detail in Section~\ref{sssec:facetedTips}.

%%% Winfield figure
\begin{figure*}[t]
\begin{center}
\includegraphics[width=6.5in, viewport=68in 21.5in 98.8in 61.8in, clip=true]{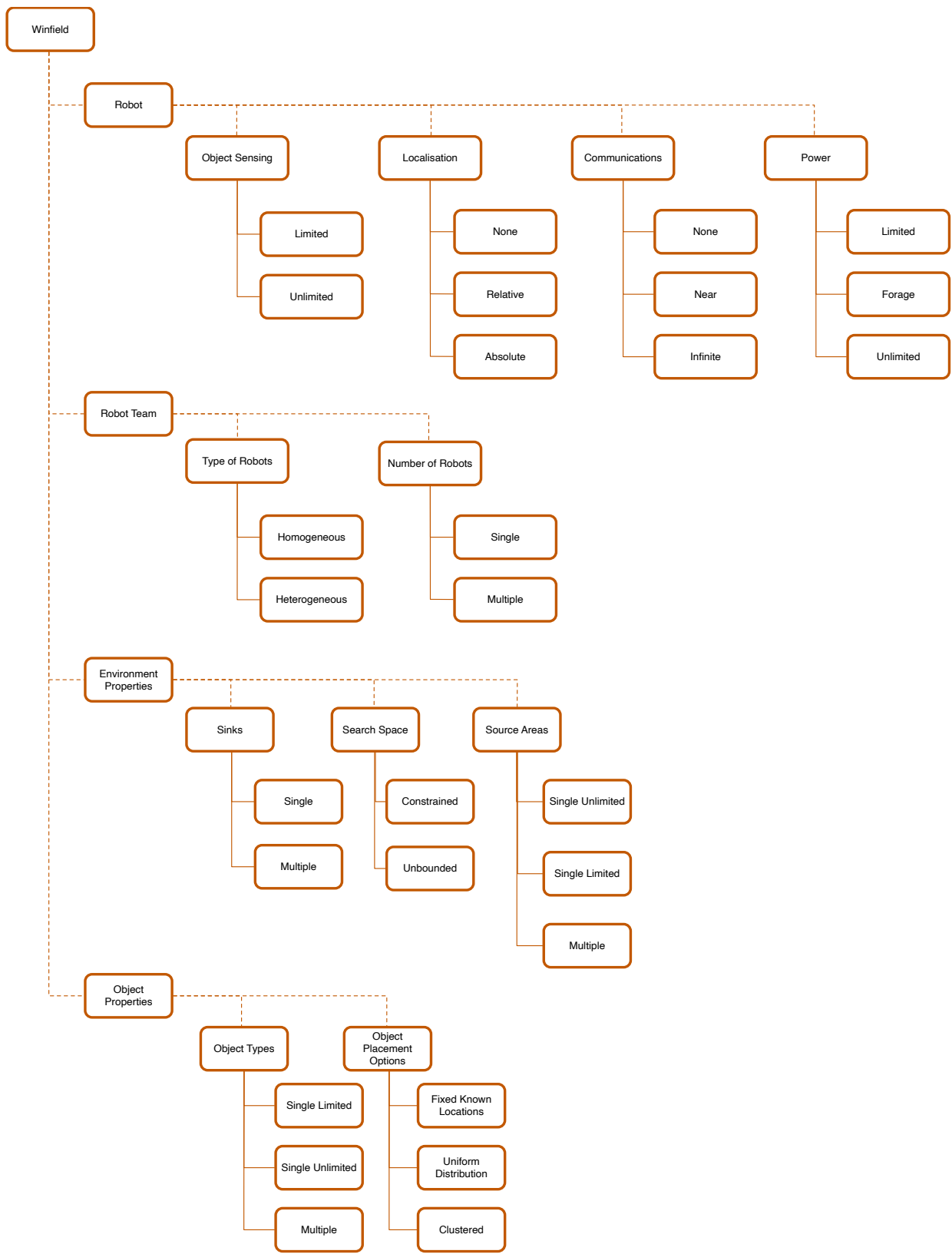}
\caption{Winfield's taxonomy of robot foraging.}
\label{fig:SWinfield}
\end{center}
\end{figure*}
%%% Pulford figure
\begin{figure*}[h]
\begin{center}
\includegraphics[width=6.5in, viewport=5.2in 41.5in 34in 76.5in, clip=true]{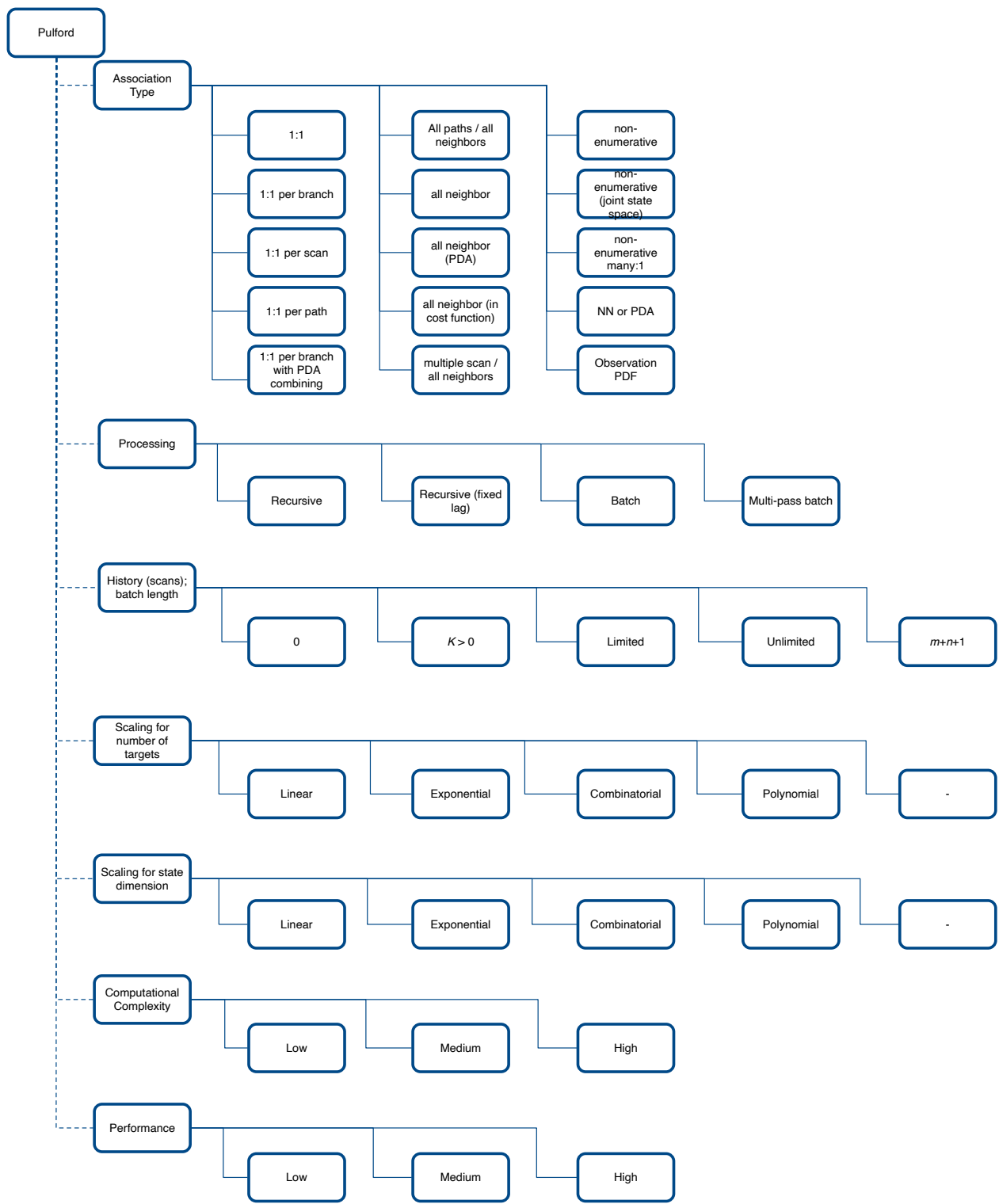}
\caption{Pulford's multi-target tracking taxonomy.}
\label{fig:TDPulford}
\end{center}
\end{figure*}

\subsubsection{\citet{Farinelli04}}
\label{sssec:farinelli}
Farinelli et al.'s resource properties taxonomy (shown in Fig.~\ref{fig:MRSFarinelli}) consists of two facets: coordination and system properties. Both facets consist of their own internally faceted structure, which in each case is further broken down via terminating branch selection nodes into category tips:
\begin{itemize}
\item{Coordination Facet:
  \begin{itemize}
  \item{Cooperation:  cooperative and non-cooperative (there is no option for variety of cooperation)}
  \item{Knowledge:  aware (acting with knowledge of other agents) and unaware (acting without knowledge of other agents)}
  \item{Coordination:  strongly coordinated (explicit coordination relying on a coordination protocol), weakly coordinated (implicit coordination that does not rely on an established protocol), not coordinated}
  \item{Organization: strongly centralized (there exists a single group leader who makes decisions for the team; non-leaders can only make decisions for themselves), weakly centralized (responsibilities of group leader can be allocated to different team members over the course of the operation), decentralized (team members make decisions independently}
  \end{itemize}}
\item{System Facet:
  \begin{itemize}
  \item{Communication:  direct (using normal communication channels) and indirect (implicit communication through the environment)}
  \item{Team Composition:  heterogeneous and homogeneous}
  \item{System Architecture:  this captures team architecture rather than robot architecture; reactive implies that the team responds on an individual level while deliberative implies that the team has a defined strategy for reconfiguring iteslf}
  \item{Team Size:  many or few, but any actual numbers or thresholds are undefined}
  \end{itemize}}
\end{itemize}

This taxonomy offers a particularly good illustration of the case where the distinction between category tips and assign value tips is difficult to make.  {\em Reactive} and {\em Deliberative} are obviously categories of system architecture that may each include many lower level categories and instances.  {\em Many} and {\em Few} are clearly range-defined numeric values.  The {\em Team Size} intermediate category could become an assign value tip without loss of specificity.  The definitions under {\em Organization}, however, define very specific team organizations.  Teams either do or do not have one single group leader who makes decisions for the group as a whole at all times.  There is no room for definition of lower level categories related to team organization below this category.  Instead of defining the choice in terms of centralization, the choice could be defined as an assign value tip for {\em Number of Leaders} --- decentralized would correspond to no leader, strongly centralized would correspond to one leader, and weakly centralized would correspond to more than one time-varying leaders.

For the purposes of this review, the definitions and structure from the original paper are retained.  However, as a new integrated taxonomy is developed, consistent rationales for the decisions about how to frame these concepts and how to relate them to the larger structure will need to be developed.

\subsubsection{\citet{Winfield09}}
\label{sssec:winfield}
Winfield's taxonomy of robot foraging defines properties of the environment and robot rather than task types or task properties.  This taxonomy is shown in Fig.~\ref{fig:SWinfield} and has a faceted structure, where the only branch select nodes are immediately above the tips.  This taxonomy terminates in category tips rather than assigned value tips.

Even though the maximum number of elements at a branch selection node is only three, the large number of faceted nodes means that this structure supports representation of large numbers of different robot foraging resource properties.

This taxonomy also brings up another aspect of categorization and representation that will need to be addressed in a larger taxonomy.  At the top level, there is a facet for {\em Robot} and a facet for {\em Robot Team}.  If the [Robot Team $\Rightarrow$ Type of Robots $\Rightarrow$  Heterogeneous] and [Robot $\Rightarrow$  Number of Robots $\Rightarrow$ Multiple] options are selected, then it may not be possible to adequately capture the team members' properties with a single instance from the Robot facet.  Instead, the team may consist of one robot with limited sensing, relative localization, and near communications that can forage for power and one robot with limited sensing, absolute localization, and near communications that has limited power.    In order to represent this team, more than one instance of the Robot facet must be able to be selected as part of an instance of this taxonomy.

%%% Daas figure
\begin{figure*}[t]
\begin{center}
\includegraphics[width=5in, viewport=141.6in 43.3in 162.1in 72.2in, clip=true]{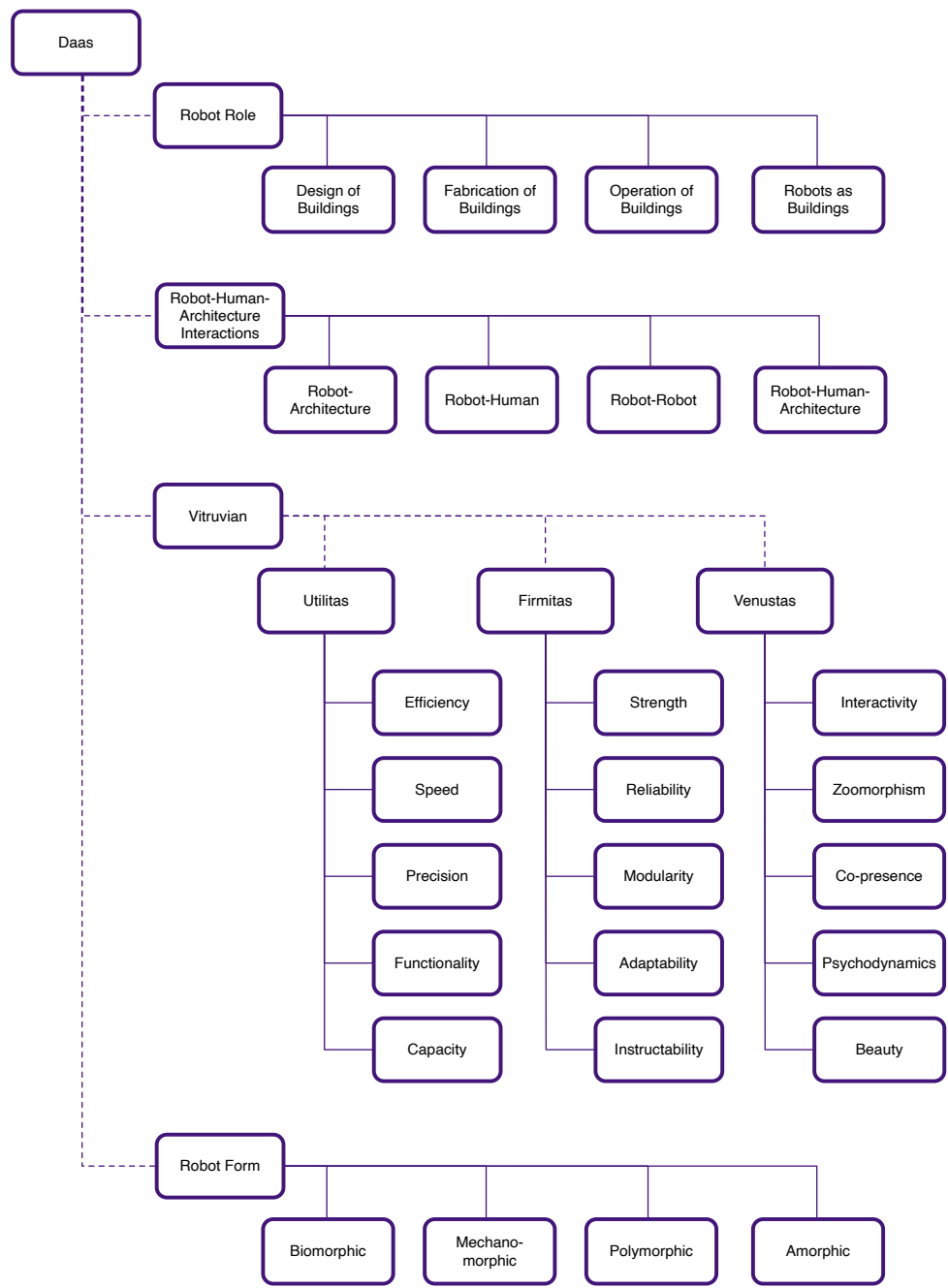}
\caption{Daas' taxonomy of robot-human-architecture interactions.}
\label{fig:SDaas}
\end{center}
\end{figure*}

\subsubsection{\citet{Pulford05}}
\label{sssec:pulford}
Pulford's multi-target tracking taxonomy (shown in Fig.~\ref{fig:TDPulford}) abstracts and categorizes different mechanisms for solving target tracking problems.  Because it has only one intermediate layer, there is one faceted node at the top level and branch select nodes above the tips.
\begin{itemize}
\item{Processing:  this resource category uses a branch selection structure to define different ways in which a given algorithm handles optimization across multiple criteria}
\item{History (scans); batch length:  this is also a resource branch selection structure that defines how complex the processing step is.}
\item{Association type:  this defines how the competing elements are processed.  This is a branch selection structure of a list of categories and illustrates the list-based approach to option selection, in contrast to Yanco and Drury's assigned value approach.}
\item{Scaling for number of targets:  this is another branch selection structure to define one type of scaling}
\item{Scaling for state dimension:  this is another branch selection structure to define one representation of dimensional state}
\item{Computational complexity:  this is another branch selection structure to define the relative level of computational complexity}
\item{Performance: this is another branch selection structure to define the relative performance evaluation mechanism}
\end{itemize}

In some cases (e.g. {\em Performance}), the tips are straightforward categories.  In others, however, category tips and assign value tips are mixed.  In the {\em History (scans): batch length} facet, some tips are assign value (e.g. {\em 0} and {\em K $>$ 0}), while others are category type (e.g. {\em Limited} and {\em Unlimited}).  The {\em m+n+1} tip could be an assign value tip if the resulting value is defined by specific values for {\em m} and {\em n}, or it could be a category tip capturing a number of different approaches to batch length depending on how {\em m} and {\em n} are defined.  Furthermore, {\em Limited} and {\em Unlimited} would appear to fully cover the space of possible batch lengths and overlap with the assign value tips, with {\em 0} under {\em Limited} and {\em K $>$ 0} and {\em m+n+1} under {\em Unlimited}.  If this change was made, either additional generic categories would then need to be created under both {\em Limited} and {\em Unlimited} to capture the non-specific options not currently captured by the {\em 0}, {\em K $>$ 0}, and {\em m+n+1} tips, or some mechanism to support early termination of those branches would be necessary to capture the increased abstraction.  If this change were made, the introduction of a new branch select node between the top level faceted node and the lowest level branch select node would transition this taxonomy to a facet-over-branch structure.

%%% Branch over Facet figure
\begin{figure*}[t]
\begin{center}
\includegraphics[width=6in, viewport=138in 110.6in 163in 119in, clip=true]{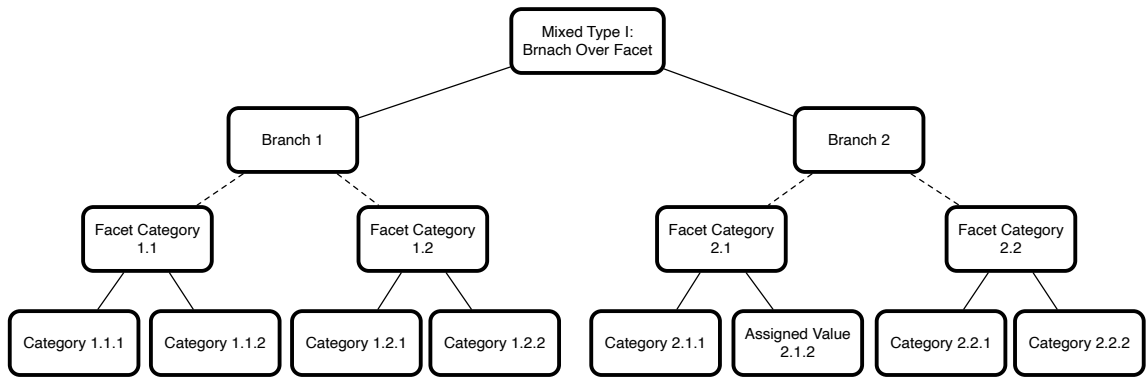}
\caption{Basic structure of taxonomies with branching structures above faceted structures.}
\label{fig:BranchOverFacet}
\end{center}
\end{figure*}
%%% Yim figure
\begin{figure*}[t]
\begin{center}
\includegraphics[width=4.25in, viewport=58.5in 99.7in 77.2in 111.5in, clip=true]{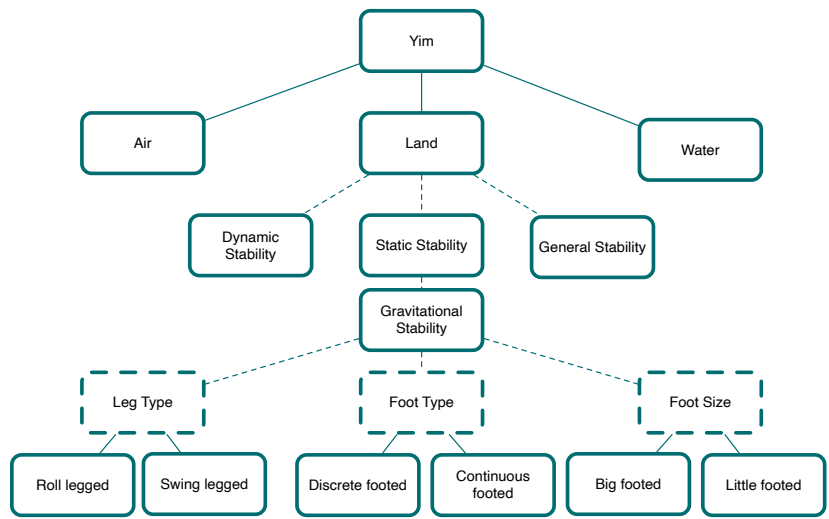}
\caption{Yim's taxonomy of simple locomotion.}
\label{fig:SYim}
\end{center}
\end{figure*}

\subsubsection{\cite{Daas14}}
This taxonomy has a faceted structure that combines a task type taxonomy with a resources taxonomy.  The robot role and robot-human-architecture interactions facets proposed by Daas create a taxonomy of tasks, while the vitruvian and robot form facets describe the robot or its properties and would be categorized as describing resources.  Dass' taxonomy is shown in Fig.~\ref{fig:SDaas}.
\begin{itemize}
\item{Robot Role Frame:  this intermediate category captures the role that the robot plays with respect to the building - a given robot will only play one of these roles
  \begin{itemize}
  \item{Design of Buildings:  desktop or prototyping robots that assist in building design}
  \item{Fabrication of Buildings:  construction robots}
  \item{Operation of Buildings:  robots that form components of buildings (e.g. skins)}
  \item{Robots as Buildings (mobile autonomous or semi-autonomous systems that people live or work inside)}
  \end{itemize}}
\item{Robot-Human-Architecture Interactions Frame:  this intermediate category captures the way in which the robot interacts with humans in relation to the building --- this is also a branch select node, although it could be represented as a faceted node if the {\em Robot-Human-Architecture (All)} tip were removed.
  \begin{itemize}
  \item{Robot-Architecture:  robot participation in the design process or directly interacting with buildings}
  \item{Robot-Human:  robot interacts with human to enhance usability of building (e.g. give directions)}
  \item{Robot-Robot:  robots interacting with other robots in the service of architecture (e.g. self-assembling systems)}
  \item{Robot-Human-Architecture (All):  three way interactions}
  \end{itemize}}
\item{Vitruvian Triad Frame:  the faceted node under this intermediate category enables the user to capture properties of the robot.  The tips are mixed, with some (e.g., {\em Speed}) potentially referring to an assigned value and others (e.g. {\em Functionality}), potentially referring to a further taxonomic substructure.
  \begin{itemize}
  \item{Utilitas, tips:  efficiency, speed, precision, functionality, capacity}
  \item{Firmitas, tips:  strength, reliability, modularity, adaptability, instructability}
  \item{Venustas, tips:  interactivity, zoomorphism, co-presence, psychodynamics, beauty}
  \end{itemize}}
\item{Robot Form Frame:  this intermediate category captures the form of the individual robot, where ``form'' refers to the observed physical structure
  \begin{itemize}
  \item{Biomorphic:  robots that resemble living things}
  \item{Mechanomorphic:  robots that resemble machines or embody mechanical characteristics in their form}
  \item{Polymorphic:  robots that assume different forms}
  \item{Amorphic:  robots with no identifiable form}
  \end{itemize}}
\end{itemize}
 
The purpose of this taxonomy is to classify the ways in which robots may be involved with the design, construction, and use of physical buildings, and it is therefore primarily a task type taxonomy.  Each instance can be used to describe the parameters of a job the robot must be capable of (e.g. [Design of Buildings : Robot-Architecture : [Precision : Modularity : Interactivity] : Mechanomorphic]). 

From a structural standpoint, the {\em Vitruvian} facet is the most interesting.  While for some systems, all three lower level facets will contribute useful information to the description of the task, in other cases only one may be needed.  In some cases, more than one of the tips in each facet may be needed to properly describe the job.  Some jobs will need to emphasize precision, speed, and reliability, while others will need to focus on interactivity and instructability.  The choice of branch select nodes above the category tips was essentially arbitrary, as it does not affect the metrics, but will need to be revisited to ensure the associated concepts are properly integrated into a larger taxonomy.
 
%%% Facet over branch
\begin{figure*}[t]
\begin{center}
\includegraphics[width=6in, viewport=138.4in 120.4in 163in 128.6in, clip=true]{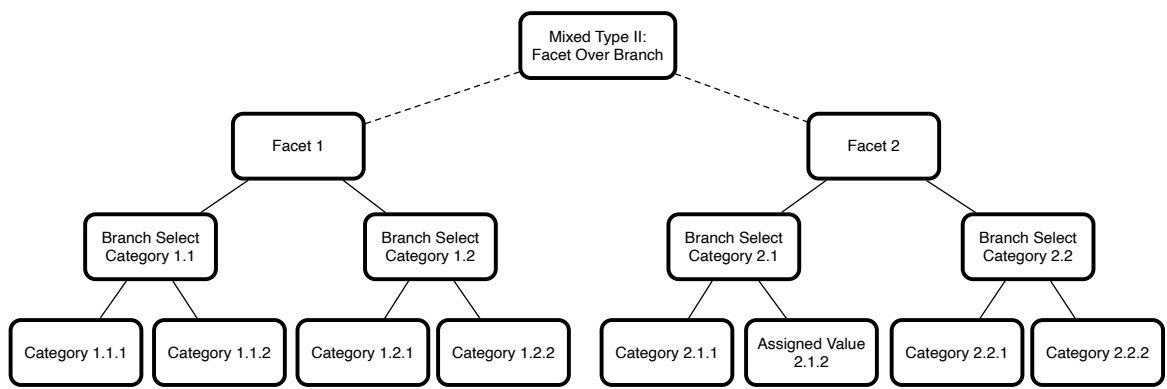}
\caption{Basic structure of taxonomies with faceted structures above branching structures.}
\label{fig:FacetOverBranch}
\end{center}
\end{figure*}
%%% Ab.Acus 3
% Ab.Acus taxonomy 1 (robot structure)
\begin{figure*}[t]
\begin{center}
\includegraphics[width=4.25in, viewport=170.2in 57.6in 187.6in 68.5in, clip=true]{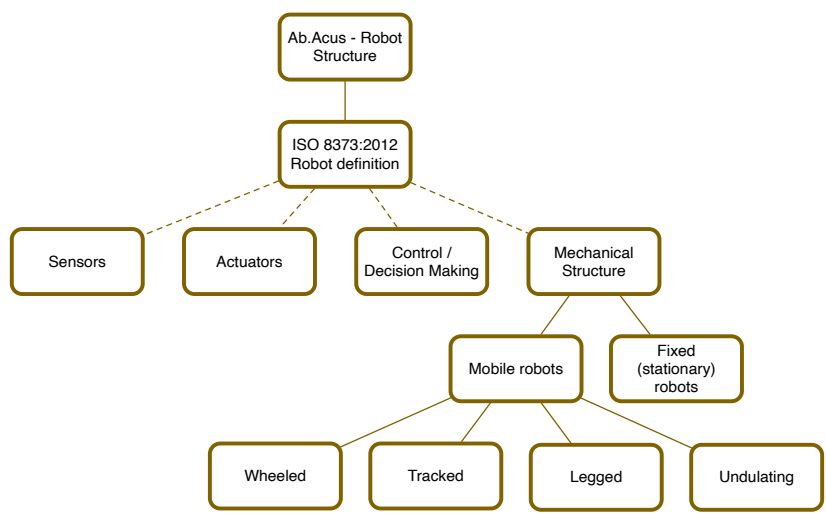}
\caption{Ab.Acus's taxonomy of robot structure from ISO 8373:2012}
\label{fig:SAbAcus_1}
\end{center}
\end{figure*}

%%%%%%%    BRANCH OVER FACET    %%%%%%%%
\subsection{Branch Over Facet (Mixed Type I) Taxonomies}
\label{ssec:branchOverFacet}
The branch-over-facet taxonomies have a branching structure at the top of the structure and faceted structures at the lower levels.  The tips for these taxonomies still use a branch selection structure as shown in Fig.~\ref{fig:BranchOverFacet}.

An instance of this taxonomy would be [Category 2.1.1 : Category 2.2.2], because only one branch can be taken, but both facet categories within that branch must be evaluated.  There is only one reviewed robotics taxonomy with this structure.

\subsubsection{\citet{Yim94}}
\label{sssec:yim}
Yim's taxonomy of simple locomotion is a resources taxonomy with a mixed structure.  Instead of the common arrangement with a faceted structure above a branch selection structure, however, this taxonomy has branching structures on top and a faceted structure at lower levels (again, terminating in branch select nodes immediately above the tips).

Only the Land branch is expanded, although similar constructs may exist for the Air and Water branches.  As we follow the potential facets down, through branches that categorize the locomotion type through type of stability, we find three facets representing the physical structure of the robot:  the leg type, the foot type, and the foot size.  

This taxonomy is demonstrably incomplete, only addressing detailed categories for the {\em Gravitational Stability} aspect of the problem.  It captures an important aspect of locomotion for systems that operate in approximately uniform gravitational fields on solid surfaces, and could easily be extended to capture more types of locomotion.  Presumably locomotion by systems that operate in narrow tunnels via pressure could be captured as something like {\em Pressure Stability} under {\em Static Stability} and systems that move like sidewinders could be captured under {\em Dynamic Stability}, while spacecraft locomotion could be captured under a new {\em Vacuum} facet that presupposes both complex gravitational fields and the absence of a surrounding medium.

%%% Cao et al.'s figure
\begin{figure*}[t]
\begin{center}
\includegraphics[width=5.5in, viewport=165.9in 72.8in 190.5in 112.3in, clip=true]{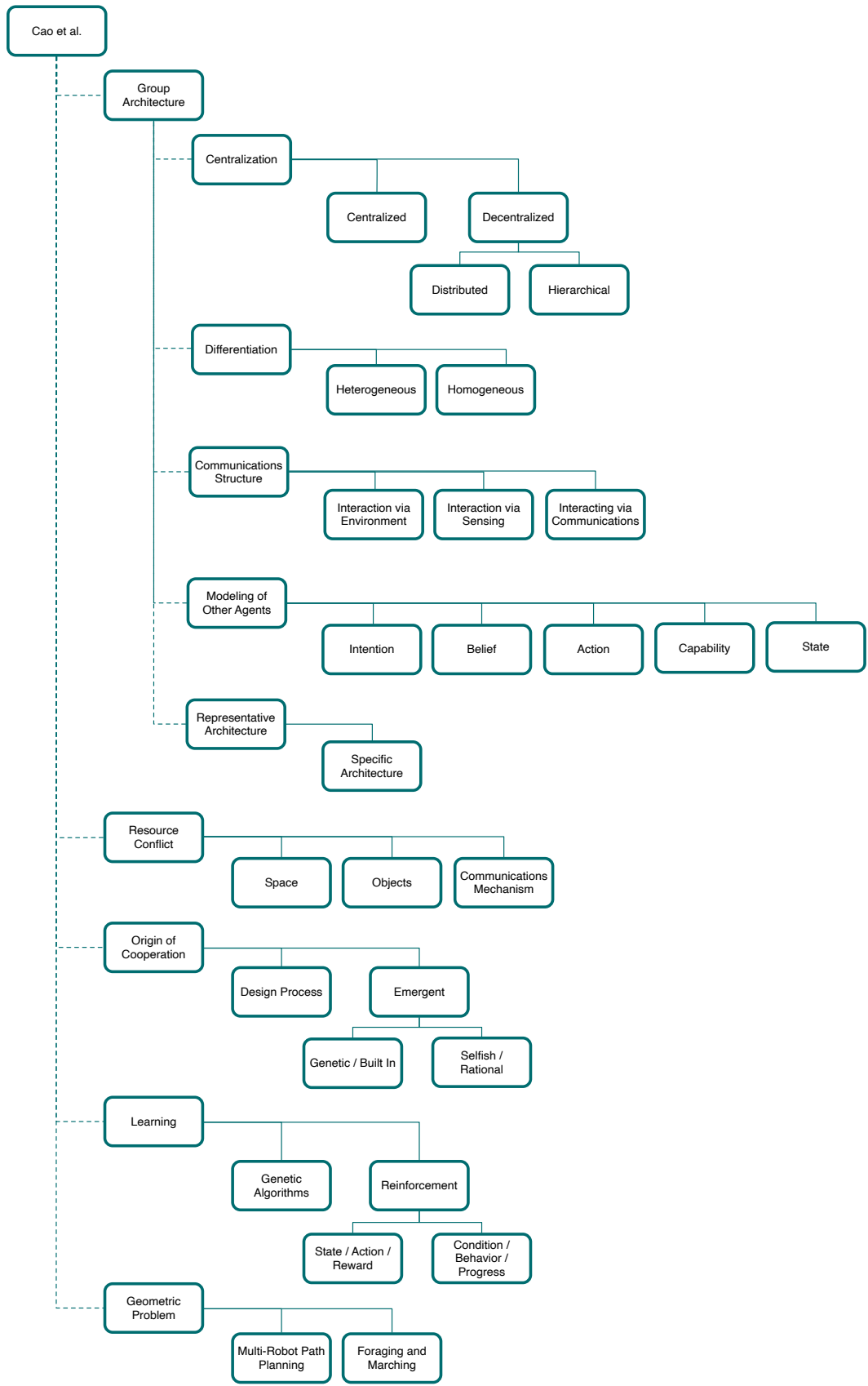}
\caption{Cao et al.'s cooperative robot problem taxonomy}
\label{fig:MRSCao}
\end{center}
\end{figure*}
%%% True Mixed type III
\begin{figure*}[t]
\begin{center}
\includegraphics[width=6in, viewport=138in 99.2in 163.1in 107.3in, clip=true]{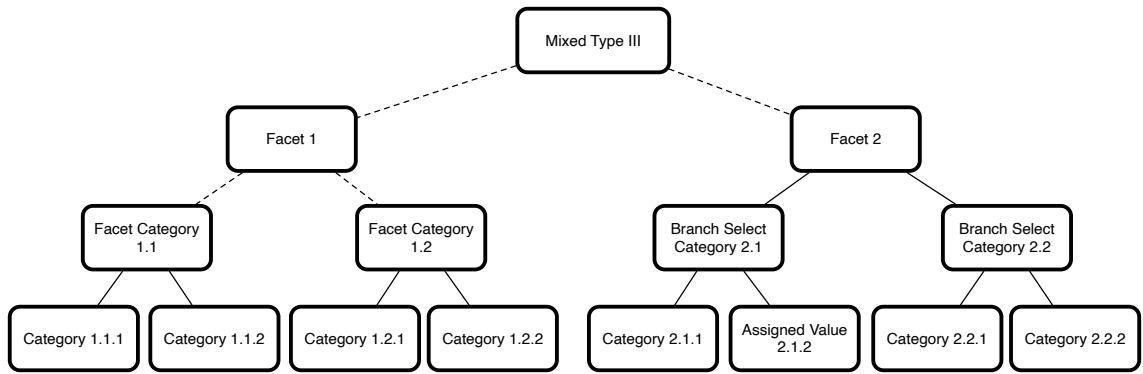}
\caption{Mixed taxonomy structure, Type III, with a top level faceted structure and mixed lower level structures.}
\label{fig:MixedTypeIII}
\end{center}
\end{figure*}
%%% True Mixed type IV
\begin{figure*}[t]
\begin{center}
\includegraphics[width=6in, viewport=138in 88.2in 162.6in 96.5in, clip=true]{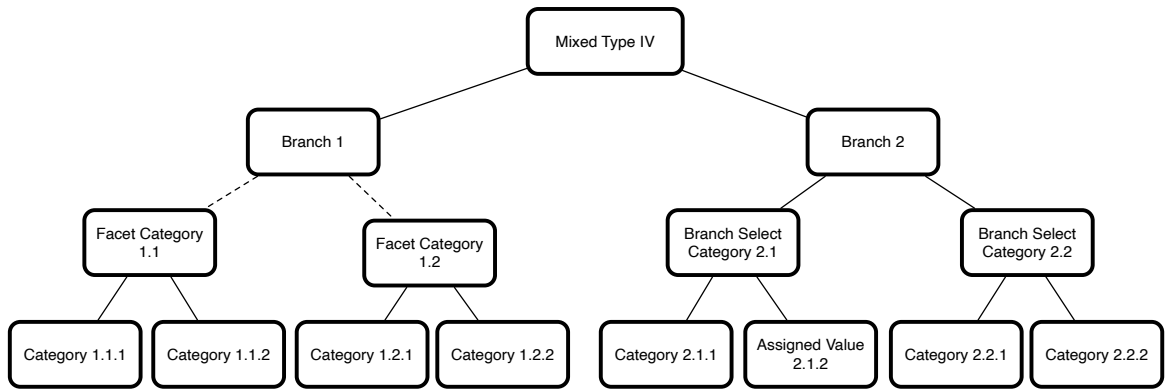}
\caption{Mixed taxonomy structure, Type IV, with a top level branch selection structure and mixed lower level structures.}
\label{fig:MixedTypeIV}
\end{center}
\end{figure*}

%%%%%%%    FACET OVER BRANCH    %%%%%%%%
\subsection{Facet Over Branch (Mixed Type II) Taxonomies}
\label{ssec:facetOverBranch}
The facet-over-branch taxonomies have a faceted structure at the top of the structure and branch selection in the lower level categories, but unlike the facet-only taxonomies in Section~\ref{ssec:facetOnly}, there must be at least one branch select node that leads to an intermediate category.  Category or assign value tips may be used to define a specific instance.  In Fig.~\ref{fig:FacetOverBranch}, one instance of this taxonomy would be defined as [Category 1.1.2 : Assigned Value 2.1.2].

\subsubsection{\citet{AbAcus17}}
\label{sssec:abacus3}
The Ab.Acus taxonomy document identifies one facet-over-branch taxonomy as shown in Fig.~\ref{fig:SAbAcus_1}.  This taxonomy breaks up the robot's physical structure into various component types (the top level faceted node), and then categorizes the system further based on capability defined by mechanical components (the intermediate branch select node), eventually categorizing based on the specific physical motion components (the category tips).

Again, the faceted tips imply an incomplete taxonomic structure, as the categories under {\em Sensors}, {\em Actuators}, and {\em Control / Decision Making} are likely to be even more complex than the limited categories under {\em Mechanical Structure}.

Note also that the {\em Mechanical Structure} category is categorizing similar properties to Yim's simple locomotion taxonomy (Section~\ref{sssec:yim}).  While Yim is attempting to classify locomotion types (behavior resources) based on physical/mechanical properties of the system, this taxonomy is attempting to classify robot bodies based on physical mechanical properties of the system.  This overlap indicates that it may not be possible to define a single taxonomy for resources that includes both behaviors and physical structure unless it also includes the ability to provide references that can connect one to the other.  A larger taxonomy will need to be able to generate descriptors for specific platforms and be able to use them to further classify behaviors --- in other words, the larger taxonomy will need a platform-specific facet and a behavior-specific facet.

\subsubsection{\citet{Cao97}}
Cao et al. focused their taxonomy on the problems the multi-agent systems community was trying to solve rather than from a perspective of the application domains or the solution developed.  This would seem to imply that the result would be a taxonomy of tasks.  The net result, however, is a faceted taxonomy that has more in common with the resources taxonomies than with either the task property taxonomies or with the taxonomies of tasks.  

As shown in Fig.~\ref{fig:MRSCao}, the {\em Group Architecture} facet contains facets that capture the different decisions that must be made during the design phase of a system, rather than decisions that must be made in order to define a task.  The only portion of the taxonomy that is focused more on the task to be accomplished than on the resources used to accomplish it is the {\em Geometric Problem} facet, which identifies {\em Multi-Robot Path Planning} and {\em Foraging and Marching} as potential task types.  In the context of the broader taxonomy, however, it becomes clear that the task types are defined for the purpose of describing a system design in terms of its behaviors, a resource specification problem.

It contains both category tips (e.g. Group Architecture $\Rightarrow$ Modeling of Other Agents $\Rightarrow$ Intention/Belief/Action/...) and assign value tips (e.g. Group Architecture $\Rightarrow$ Representative Architecture $\Rightarrow$ Specific Architecture).  However, because each terminating node is a branch select node, only the assign value tips contribute additional representational complexity.

%%% Yanco and Drury figure
\begin{figure*}[t]
\begin{center}
\includegraphics[width=6.75in, viewport=4.3in 136.1in 39.7in 149.3in, clip=true]{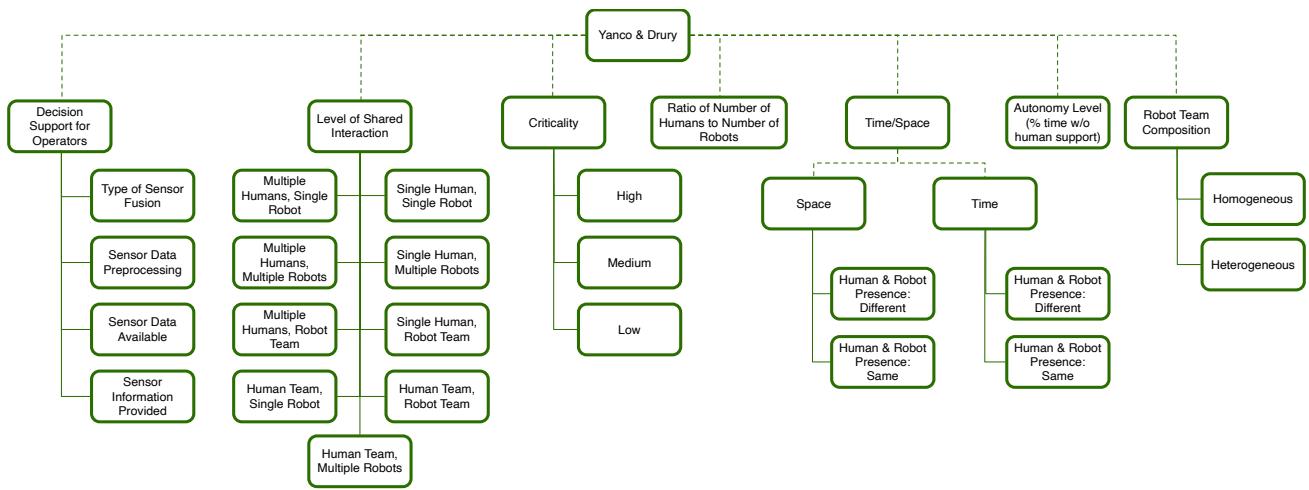}
\caption{Yanco and Drury's taxonomy of types of human-robot interaction.}
\label{fig:HRIYanco}
\end{center}
\end{figure*}
%%% Carlson figure
\begin{figure*}[t]
\begin{center}
\includegraphics[width=6.5in, viewport=75.1in 61.8in 102.9in 76.2in, clip=true]{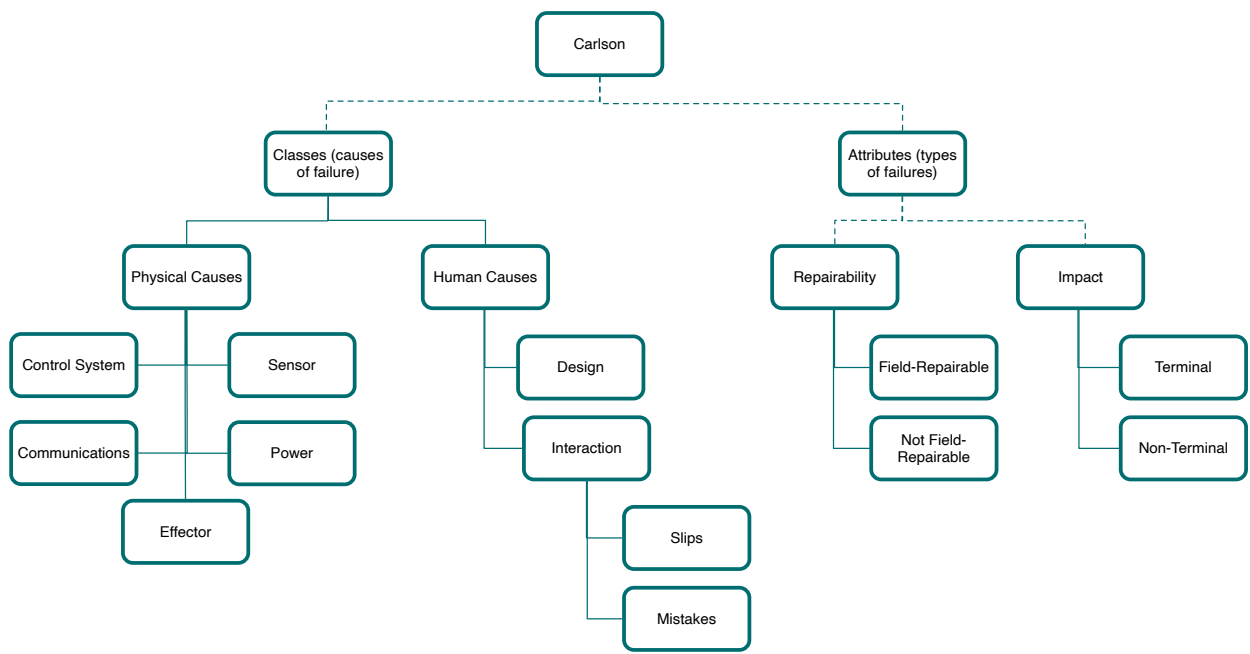}
\caption{Carlson et al.'s taxonomy of robot failure modes.}
\label{fig:SCarlson}
\end{center}
\end{figure*}

%%%%%%%    TRUE MIXED    %%%%%%%%
\subsection{Mixed (Types III and IV) Taxonomies}
\label{ssec:trueMixed}
The mixed taxonomies in this section do not have a clear structural relationship between faceted nodes and branch selection nodes.  They may have branches where a faceted decision point occurs above another faceted decision point along one branch and above a branch selection decision point along another branch, as shown in Fig.~\ref{fig:MixedTypeIII}, or they may have some branches with faceted decisions above branch selection decisions, as shown in Fig.~\ref{fig:MixedTypeIV} and other branches where branch selection occurs above faceted decisions.

The Type IV taxonomy shown in Fig.~\ref{fig:MixedTypeIV} can have instances whose structure differs, since both [Category 1.1.1 : Category 1.2.1] and [Assigned Value 2.1.2] are valid instances.  An instance of the relatively simple Type III taxonomy shown in Fig.~\ref{fig:MixedTypeIII} will always have the same structure because all the branch select nodes lie beneath the faceted nodes:  [Category 1.1.2 : Category 1.2.2 : Assigned Value 2.1.2], but this is not true in general.  A more complex Type III taxonomy may have arbitrary layers of faceted and branch select nodes and the structure of the instances will vary accordingly.

\subsubsection{\citet{Yanco04}}
This Human-Robot Interaction taxonomy by Yanco and Drury is a faceted descriptive taxonomy that defines types of human robot interaction along seven primary facets.  This taxonomy is shown in Fig.~\ref{fig:HRIYanco}.  The facets capture:
\begin{itemize}
\item{Decision support for operators --- this is a faceted resource-based categorization mechanism that has four subcategories, each addressing different aspects of what kind of information is available to the operator:
  \begin{itemize}
  \item{available sensor information --- this is a resource value that is defined as a list of sensing types available on the robot platform, and may also contain location of sensors. If the taxonomy identified a comprehensive list of potential sensors, it would require both a faceted approach (to address multiple sensor types) and a multiple-instance approach (to address multiple instances of the same sensor type). It provides a baseline for understanding the other categories}
  \item{sensor information provided --- this is another resource value that consists of a subset of the previous available sensor information list.  It defines the information the robot provides to the operator}
  \item{type of sensor fusion --- this is a resource category that establishes the software or capability-based resources available to accomplish the task; the categories available under this item include functions that take in multiple items from the sensor information provided list.}
  \item{pre-processing ---  this consists of the amount of pre-processing applied to sensor data for decision support [sensorA goes to processing output]}
  \end{itemize}}
\item{Level of shared interaction among teams --- this is another resource category broken into eight possible configurations based on the structure and composition of the teams and how they interact.}
\item{Criticality --- this is a task property category that uses a traditional branch selection mechanism; it is defined as the importance of getting a task done correctly in terms of negative effects if there's a problem (high, medium, low)}
\item{Ratio of people to robots --- this is a resource value that is defined as a non-reduced fraction of the number of humans over the number of robots.}
\item{Time/space taxonomy --- these task properties are pulled from \citet{Ellis91}.  They are defined via four subcategories, where time is represented as synchronous or asynchronous and space is represented as collocated and non-collocated.
  \begin{itemize}
  \item{same space, same time --- example = robot wheelchair, where the robot's role in the job is done in the same place at the same time as the human's role in the job}
  \item{same space, different time --- example = manufacturing robots, where the robot performs part of the job at one time and the human performs another part of the job at a different time but in the same space}
  \item{different space, same time --- example = urban search and rescue, where the human and robot work together to accomplish the job at the same time but are located in different places}
  \item{different space, different time --- example = mars rover, where the robot performs its job in a different place and at a different time than the human performs theirs}
  \end{itemize}}
\item{Autonomy level / Amount of intervention --- this is a faceted resource value category that defines the amount of intervention needed to control a robot measured along an axis from teleoperation to full autonomy.  The Autonomy subclass is measured by the percentage of time that the robot is carrying out its task without human support; the Intervention subclass is measured by the percentage of time that the human operator is controlling the robot (Autonomy + Intervention = 100 percent).}
\item{Composition of robot teams --- this is a selection-based resource category that can take the values homogeneous or heterogeneous and only applies to robot teams.  Heterogeneous may be further specified as a resource value like the {\em available sensor information} element with a list of the types of robots and numbers of each type of robot on the team.}
\end{itemize}

Just within this one taxonomy, we see great diversity in structure and purpose.  Some of these categories describe task properties and some define resources, some are faceted and some are branch selection based, and it includes both category and assign value tips.  Even though it has relatively few elements, the complexity of the concepts it can represent is high.

\subsubsection{\citet{Carlson04}}
Carlson et al.'s taxonomy of robot failures (shown in Fig.~\ref{fig:SCarlson}) categorizes failure modes that a given system needs to be robust to in the context of a desired task.  It is a mixed structure, with a top level faceted structure that captures failure causes and attributes and lower levels that include both branch select nodes and faceted nodes.  It has category tips rather than assign value tips, and has no faceted terminating nodes.  The {\em Physical Causes} intermediate category in particular has category tips that map onto robot components and could lead to a more comprehensive taxonomy of robot parts.  The taxonomy as a whole describes the properties of a failure, and could therefore be used to describe either a task requirement (task type) or a description of provided system robustness (resource property).

%%%%%%%    EXCLUDED    %%%%%%%%
\subsection{Excluded Taxonomies}
While this review contains a comprehensive survey of taxonomies in robotics, some were deemed out of scope.  

Two taxonomies did not focus on the categorization of robots, robot properties, tasks, or task properties.  The \citet{Salter10} environmental property taxonomy of child-robot interaction `wildness' is designed purely to characterize {\em experiments} (rather than robots or tasks) involving interactions between children and robots, and is not properly a resource or a task type taxonomy.    The \citet{Adamides15} taxonomy of user interface design properties enables users to identify {\em drawbacks} of a given user interface design, but is not intended to describe the interface itself. 

Several proposed taxonomies, including \citet{Gottschlich94}, \citet{Hollerbach96}, \citet{Kowadlo08}, and \citet{Gustafson03}, were excluded because this review focuses on the form and structure of taxonomies and they defined terms but not the structure relating those terms to each other.   

Other taxonomies were excluded because they were subsumed in more recent taxonomies.  \citet{Cutkosky89}, for example, was subsumed in \citet{Feix15}'s taxonomy of grasps.  

\section{Common Properties}
\label{sec:analysis}
The following variables are used to develop metrics for the structure and form of the taxonomies:  
 \begin{itemize}
 \item {{\em Layers} is the total number of intermediate categories and tips along a single path through the taxonomy, starting at (but not including) the root.
   \begin{itemize}
   \item {{\em Maximum} ($L_M$) is the maximum number of layers it is possible to traverse through the taxonomy along a single path.}
   \item{{\em Sum} ($L_S$)is the sum of all the maximum layer paths you need to take through a single taxonomy in order to create a valid instance.  For a branch-only taxonomy, this number is equal to the maximum number of layers.}
   \end{itemize}}
\item {{\em Elements} ($E$) is the total number of categories, not including the root but including all the tips, in the taxonomy.}
\item {{\em Nodes} ($N$) is the total number of nodes in the taxonomy.}
\item {{\em Choices} is the total number of options at a given node in the taxonomy.
  \begin{itemize}
  \item{{\em Maximum} (${C}_{x}$) is the maximum number of choices possible in the taxonomy.}
  \item{{\em Average} (${C}_{A}$) is the average number of choices across each node in the taxonomy.}
  \item{{\em Minimum} (${C}_{n}$) is the minimum number of choices possible in the taxonomy.}
  \end{itemize}}
\item{{\em Branch Structure} ($B$) is which category of branch selection structure the taxonomy falls into:  branch-only (BO), facet-only (FO), branch-over-facet (BOF), facet-over-branch (FOB), or mixed type III or IV (MTIII or MTIV).}
\item{{\em Assign Value Tips} ($AT$) captures the number of non-binary assign value tips present in the taxonomy; taxonomies with only category tips are represented by ``-''; a number in parentheses indicates the number of binary yes/no assign value tips, each of which is counted as two options.}
\item{{\em Faceted Tips} ($FT$) captures the number of tips that have a faceted node as their parent node.}
\item{{\em Content} (${T}$) is the type of content the taxonomy is intended to categorize: resources, tasks or task properties.}
\item{{\em Instances} ($I$) is the number of possible instances that the taxonomy can be used to express.  Where a taxonomy has a faceted node above a tip, that tip is counted as contributing one instance.}
\end{itemize}
The raw data for all of these variables is shown in Table~\ref{tab:rawdata1}.

\begin{table*}
\small\sf\centering
\caption{Taxonomy property comparison data, organized by structure.}
\label{tab:rawdata1}
\begin{tabular}{ | c || c | c | c | c | c | c | c | c | c | c | c | c |}
\toprule
\multicolumn{1}{| c ||}{\bfseries Taxonomy} & {$\bm {L_M}$} & {$\bm {L_S}$} & {$\bm {E}$} & {$\bm {N}$} & {$\bm {{C}_{x}}$} & {$\bm {{C}_{A}}$} & {$\bm {{C}_{n}}$} & {$\bm {I}$} & {$\bm {B}$} & {$\bm {T}$} & {$\bm {AT}$} & {$\bm {FT}$} \\ 
\midrule
Ab.Acus (Siciliano) & 2 & 2 & 7 & 3 & 3 & 2.33 & 2 & 5 & BO & R & - & 0 \\ 
Ab.Acus (IFR) & 2 & 2 & 19 & 3 & 12 & 6.33 & 2 & 17 & BO & R & - & 0  \\ 
Bullock \& Dollar & 5 & 5 & 28 & 14 & 2 & 2.00 & 2 & 15 & BO & T & - & 0 \\ 
Dudek et al. (Task) & 1 & 1 & 4 & 1 & 4 & 4.00 & 4 & 4 & BO & T & - & 0 \\ 
Feix et al. & 5 & 5 & 72 & 41 & 5 & 1.80 & 1 & 33 & BO & T & - & 0 \\ 
Fishwick & 2 & 2 & 24 & 6 & 6 & 4.00 & 2 & 19 & BO & R & - & 0 \\ 
Korsah et al. & 2 & 2 & 25 & 5 & 8 & 5.00 & 2 & 21 & BO & R, T & - & 0 \\ 
Metzler \& Shea & 3 & 3 & 40 & 11 & 13 & 3.64 & 1 & 30 & BO & R & - & 0 \\ 
Robin \& Lacroix & 4 & 4 & 12 & 6 & 3 & 2.00 & 1 & 7 & BO & R & - & 0 \\ \midrule
Beer et al. & 2 & 6 & 12 & 4 & 3 & 3.00 & 3 & 27 & FO & R & - & 0  \\ 
Bullock \& Dollar (eff.) & 1 & 10 & 15 & 6 & 2 & 2.50 & 2 & {\bfseries 32*} & FO & T & - & 0 \\ 
Daas & 3 & 13 & 34 & 8 & 4 & 4.25 & 4 & 8,000 & FO & T & - & 0 \\ 
Dudek et al. (Collective) & 2 & 9 &10 & 3 & 7 & 3.33 & 1 & 2  & FO & R & 5 & 5 \\ 
Farinelli et al. & 3 & 18 & 28 & 11 & 4 & 2.55 & 2 & 576 & FO & R & - & 0 \\ 
Feix et al. (eff.) & 1 & 8 & 13 & 5 & 6 & 2.60 & 2 & {\bfseries 126*} & FO & T & 1 & 0 \\ 
Gerkey \& Matari\'{c} & 2 & 6 & 9 & 4 & 3 & 2.25 & 2 & 8 & FO & R & - & 0 \\ 
Jiang \& Arkin & 2 & 6 & 14 & 4 & 4 & 3.50 & 3 & 48 & FO & R & - & 0 \\ 
Pulford & 2 & 14 & 47 & 8 & 15 & 5.88 & 3 & 67,500 & FO & R & - & 0 \\ 
Shim \& Arkin & 2 & 6 & 9 & 4 & 2 & 2.25 & 2 & 8 & FO & T & - & 0 \\ 
Tan et al. & 2 & 4 & 6 & 3 & 2 & 2.00 & 2 & 4 & FO & T & - & 0 \\ 
Tsiakas et al. & 4 & 29 & 64 & 39 & 6 & 1.64 & 2 & 116,640 & FO & R & 4 & 13 \\ 
Winfield & 3 & 26 & 43 & 16 & 4 & 2.69 & 2 & 23,328 & FO & R & - & 0 \\ 
Zech et al. & 4 & 49 & 87 & 28 & 6 & 3.11 & 1 & 477,757,440 & FO & R & (5) & 0 \\ \midrule
Ab.Acus (ISO Robot) & 3 & 6 & 10 & 3 & 4 & 3.33 & 2 & 5 & FOB & R & - & 3 \\ 
Cao et al. & 4 & 22 & 38 & 14 & 5 & 2.71 & 1 & 4,860 & FOB & R & 1 & 0 \\ \midrule
Yim & 5 & 11 & 16 & 7 & 3 & 2.29 & 1 & 8 & BOF & R & - & 2 \\ \midrule
Carlson et al. & 4 & 9 & 19 & 8 & 5 & 2.38 & 2 & 32 & MTIII & T & - & 0 \\ 
Yanco \& Drury & 3 & 15 & 31 & 8 & 9 & 3.88 & 2 & 864 & MTIII & R, TP & 2 & 0 \\ 
\bottomrule 
\multicolumn{13}{l}{\footnotesize{*Efficient faceted taxonomy variants.}}
\end{tabular}
\end{table*}

\subsection{Implications of Faceted and Assign Value Tips}
\label{sssec:facetedTips}
Where a taxonomy uses binary assign value tips (e.g. yes/no options), the value of that branch is counted as though there were two category tips.  Where a taxonomy uses non-binary assign value tips, the value of that branch is counted as though there were one category tip.  The number of these tips in each taxonomy is captured as the $AT$ value.  Assign value tips can be continuous numbers, discrete numbers, or booleans.  Continuous- and discrete-valued assign value tips can be further constrained within ranges.  The number of elements needs to include category tips because they contribute to the number of instances that can be defined.  If we include the category tips, we need to also include the places where value assignments must be made or we will be undercounting the potential instances even further as those assign value tips are defining categories of information that must be provided to create a valid instance.  

Cases where the tips are themselves facets is captured as the $FT$ value.  This occurs in only four taxonomies.  Tsiakis et al. (see Section~\ref{sssec:tsiakis}) and Dudek's multi-robot taxonomy (see Section~\ref{sssec:dudek2}) are particularly notable.  In Tsiakis et al.'s case, the rest of the taxonomy is sufficiently complex that it is already the second-most complex taxonomy reviewed.  In Dudek's taxonomy, however, the simplicity and smallness of the taxonomic structure belies the actual complexity of the possible representations.  Because each of these faceted tips is an assign value tip, rather than a category tip, the complexity is further increased.  Dudek's multi-robot system taxonomy is an excellent example of balancing the relationship between compactness of representation and diversity of instances.  However, because the resulting output is so customized to the particular instance via assign value tips, it is more difficult to categorize and group like systems at higher levels of abstraction.

In cases where faceted tips lie below other faceted nodes, another aspect of the problem needs to be addressed.  It is possible for some facets to be inapplicable in some instances.  If facets are allowed to represent null or not applicable values as well as specific categories within the facet, then each facet effectively consists of a binary exists/does not exist branch above the tip.  For the case where there is a node with two faceted tips, this would now represent three possible cases:  [A exists, B does not], [A does not exist, B exists], and [A and B both exist].  Each of these three cases would be joined by a parent branch select node, and each of these three cases would consist of a faceted tip for whichever facet (A or B) exists.

For the purposes of this document, we assume that the published version is the desired structure and these potential further refinements are not intended to be captured in a structural analysis like the one presented here.

%%% Flatness Figure
\begin{figure*}[t]
\begin{center}
\includegraphics[width=6.5in, viewport=0.2in 5.5in 11in 8.5in, clip=true]{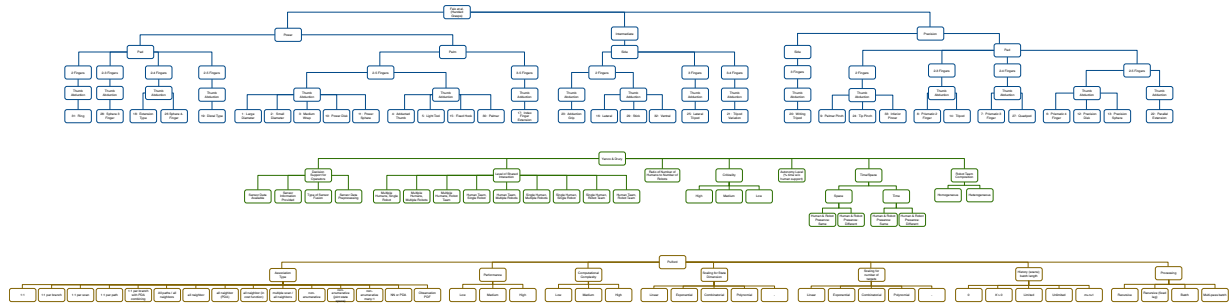}
\caption{Illustration of the {\em flatness} property.  Feix et al. ({\em tall}, top), Yanco \& Drury ({\em square}, middle), and Pulford ({\em flat}, bottom).}
\label{fig:flatnessillustration}
\end{center}
\end{figure*}
%%% Flatness Table
\begin{table}
\small\sf\centering
\caption{Flatness metric and supporting data - flatness type (Type) categorized into Flat (F), Square (S) both small (s) and medium sized (m), and Tall (T).}
\label{tab:ftdata}
\begin{tabular}{  | c || c | c | c | c |}
\toprule
\multicolumn{1}{| c ||}{\bfseries Taxonomy} & {$\bm {{C}_{A}}$} & {$\bm {L_M}$} & {$\bm {F}$} & {\bfseries Type} \\
\midrule
Feix et al. & 1.80 & 5 & 0.36 & T \\ 
Bullock \& Dollar & 2.00 & 5 & 0.40 & T \\ 
Tsiakas et al. & 1.64 & 4 & 0.41 & T \\ 
Yim & 2.29 & 5 & 0.46 & T \\ 
Robin \& Lacroix & 2.00 & 4 & 0.50 & T \\ 
Carlson et al. & 2.38 & 4 & 0.59 & T \\ 
Cao et al. & 2.71 & 4 & 0.68 & T \\ 
Zech et al. & 3.11 & 4 & 0.78 & T \\ 
Farinelli et al. & 2.55 & 3 & 0.85 & T \\ 
Winfield & 2.69 & 3 & 0.90 & T \\ \midrule
Tan et al. & 2.00 & 2 & 1.00 & S (s) \\ 
Ab.Acus (ISO) & 3.33 & 3 & 1.11 & S (m) \\ 
Gerkey \& Matari\'{c} & 2.25 & 2 & 1.13 & S (s) \\ 
Shim \& Arkin & 2.25 & 2 & 1.13 & S (s) \\ 
Ab.Acus (Siciliano) & 2.33 & 2 & 1.17 & S (s) \\ 
Metzler \& Shea & 3.64 & 3 & 1.21 & S (m) \\ 
Bullock \& Dollar (eff.) & 2.50 & 2 & 1.25 & S (m) \\ 
Yanco \& Drury & 3.88 & 3 & 1.29 & S (m) \\ 
Feix et al. (eff.) & 2.60 & 2 & 1.30 & S (m) \\ \midrule
Daas & 4.25 & 3 & 1.42 & F \\ 
Beer et al. & 3.00 & 2 & 1.50 & F \\ 
Dudek et al. (Collective) & 3.33 & 2 & 1.67 & F \\ 
Jiang \& Arkin & 3.50 & 2 & 1.75 & F \\ 
Fishwick & 4.00 & 2 & 2.00 & F \\ 
Korsah et al. & 5.00 & 2 & 2.50 & F \\ 
Pulford & 5.88 & 2 & 2.94 & F \\ 
Ab.Acus (IFR) & 6.33 & 2 & 3.17 & F \\ 
Dudek et al. (Task) & 4.00 & 1 & 4.00 & F \\ 
\bottomrule
\end{tabular}
\end{table}
%%% Structural Complexity Table
\begin{table*}
\small\sf\centering
\caption{Structural complexity metric and supporting data.}
\label{tab:scdata}
\begin{tabular}{ | c || c | c || c | c || c | c || c |}
\toprule
\multicolumn{1}{| c ||}{\bfseries Taxonomy} & {$\bm {E}$} & {$\bm {E_R}$} & {$\bm {N}$} & {$\bm {N_R}$} & {$\bm {{C}_{A}}$} & {$\bm {{C}_{R}}$} & {$\bm {SC}$} \\ 
\midrule
Ab.Acus (Siciliano) & 7 & 1 & 3 & 1 & 2.33 & 2 & 1.33 \\ 
Tan et al. & 6 & 1 & 3 & 1 & 2.00 & 2 & 1.33 \\ 
Ab.Acus (ISO) & 10 & 1 & 3 & 1 & 3.33 & 3 & 1.33 \\ 
Dudek et al. (Collective) & 10 & 1 & 3 & 1 & 3.33 & 3 & 1.67 \\ 
Gerkey \& Matari\'{c} & 9 & 1 & 4 & 2 & 2.25 & 2 & 1.67 \\ 
Shim \& Arkin & 9 & 1 & 4 & 2 & 2.25 & 2 & 1.67 \\ 
Bullock \& Dollar (eff.) & 15 & 2 & 6 & 2 & 2.50 & 2 & 2.00 \\ 
Dudek et al. (Task) & 4 & 1 & 1 & 1 & 4.00 & 4 & 2.00 \\ 
Robin \& Lacroix & 12 & 2 & 6 & 2 & 2.00 & 2 & 2.00 \\ 
Yim & 16 & 2 & 7 & 2 & 2.29 & 2 & 2.00 \\ 
Beer et al. & 12 & 2 & 4 & 2 & 3.00 & 3 & 2.33 \\ 
Carlson et al. & 19 & 2 & 8 & 3 & 2.38 & 2 & 2.33 \\ 
Daas & 34 & 2 & 8 & 3 & 4.25 & 2 & 2.33 \\ 
Feix et al. (eff.) & 13 & 2 & 5 & 2 & 2.60 & 3 & 2.33 \\ 
Jiang \& Arkin & 14 & 2 & 4 & 2 & 3.50 & 3 & 2.33 \\ 
Winfield & 43 & 2 & 16 & 4 & 2.69 & 1 & 2.33 \\ \midrule
Ab.Acus (IFR) & 19 & 2 & 3 & 1 & 6.33 & 5 & 2.67 \\ 
Farinelli et al. & 28 & 3 & 11 & 3 & 2.55 & 2 & 2.67 \\ 
Bullock \& Dollar & 28 & 3 & 14 & 4 & 2.00 & 2 & 3.00\\ 
Fishwick & 24 & 3 & 6 & 2 & 4.00 & 4 & 3.00 \\ 
Cao et al. & 38 & 4 & 14 & 4 & 2.71 & 2 & 3.33 \\ 
Korsah et al. & 25 & 3 & 5 & 2 & 5.00 & 5 & 3.33 \\ 
Metzler \& Shea & 40 & 4 & 11 & 3 & 3.64 & 3 & 3.33 \\ 
Yanco \& Drury & 31 & 4 & 8 & 3 & 3.88 & 3 & 3.33 \\ \midrule
Feix et al. & 72 & 5 & 41 & 5 & 1.80 & 1 & 3.67 \\ 
Pulford & 47 & 4 & 8 & 3 & 5.88 & 5 & 4.00 \\ 
Tsiakas et al. & 64 & 5 & 39 & 4 & 1.64 & 3 & 4.00 \\ 
Zech et al. & 87 & 5 & 28 & 5 & 3.11 & 3 & 4.33 \\ 
\bottomrule
\end{tabular}
\end{table*}

\subsection{Metrics}
\label{ssec:metrics}
Three metrics are used to compare the taxonomies and identify any patterns within their structure or form. {\em Flatness} is used to describe the structural type of the taxonomy.  It captures whether the taxonomy focuses more on drilling down into the details of a small number of categories or on providing a large number of high level categories with relatively little detail.  {\em Structural Complexity} is used to describe whether the structure of the taxonomy is simple or complicated.  Some taxonomies consist entirely of hierarchical binary choices, while others have complex combinations of faceted and branch select nodes and wildly varying numbers of choices available at each node.  {\em Representational Complexity} is used to describe the complexity of concept that a given taxonomy can be used to represent.  This includes both the number of instances that can be generated and the complexity of the individual instances.

All of these metrics are based on ranking the variables presented in Table~\ref{tab:rawdata1} into broad regimes rather than on the variables themselves.  Because the variables' values cover broad ranges of values, it might seem appropriate to develop weighted averages based on normalized values.  Alternative metrics (such as taking the average of the normalized $I$ and $L_S$ variables instead of averaging ranked values) can vary the ordering of the taxonomies somewhat and introduce slightly different natural dividing lines between low, moderate, and high, but using these normalized metrics does not shift taxonomies' relative positions very far in terms of these broad assessments.  

The broad problem with the normalized approach to these metrics is that it introduces spurious accuracy when all that is needed is a rough categorization to investigate whether there is any clearly and obviously evident relationship between the different structures, forms, and content types.  

First, the additional level of detail is unnecessary.  Whether or not a difference of $0.001$ on some weighted, normalized representational complexity metric puts a given taxonomy into a low or moderate complexity category is less relevant than whether there is a correlation between low representational complexity and flatness. 

Second, these normalized metrics are heavily dependent on the largest values in the sample --- a sample that doesn't include Zech et al.'s taxonomy will have significantly different normalized values for the $I$ variable but less different normalized values for the $L_S$ variable.  This means that for the same weighted average function, different rankings are possible based on the composition of the sample set. Having a continuum of possible values rather than an integer or fixed subset of possible values does not provide an improved measure of the relative rankings of the taxonomies if the relative rankings of the remaining taxonomies change when one sample is added or removed.  The rank-based approach to these metrics prevents this volatility.

Third, the rank-based approach is extensible.  As more complex and larger taxonomies are developed, additional ranks can be created to accommodate them without affecting either the absolute metric values for or the comparisons between the smaller and less complex taxonomies captured here.

\subsubsection{Flatness}
The {\em flatness} property metric $F$ is the ratio of the average number of choices per node ${C}_{A}$ to the maximum number of layers in the taxonomy $L_{M}$.  A relatively wide and shallow ({\em flat}) taxonomy like Pulford et al. will have a high number of choices per node and a low number of layers.  A relatively tall and narrow taxonomy ({\em tall}) taxonomy like Feix et al. will have few choices per node but many layers.  A relatively {\em square} taxonomy like Yanco and Drury will have roughly equal numbers of choices per node and layers.  If we reconfigure the taxonomy diagrams so the orientation and layout is consistent, Fig.~\ref{fig:flatnessillustration} shows the difference in flatness across these three taxonomies.  Table~\ref{tab:ftdata} shows the input data and flatness metric values for all the taxonomies.

\subsubsection{Structural Complexity}
\label{ssec:structural}
The {\em structural complexity} metric $SC$ captures the internal complexity of the taxonomy.  The number of nodes $N$, the average number of choices per node ${C}_{A}$, and the number of elements $E$ are each ranked from 1 to 5, with 1 being the smallest and 5 being the largest.  These rankings ($N_R$, ${C}_{R}$, and $E_R$) are shown in Table~\ref{tab:scdata}.  The structural complexity ${SC}$ is the average of these three rankings and is intended to capture both the relative size of the taxonomy (in terms of number of categories and tips) and the relative complexity of the taxonomy (in terms of the number of nodes and decisions per node).  A taxonomy with a relatively small number of nodes and a relatively small number of choices per node will have a lower structural complexity than a taxonomy with a relatively large number of categories and a relatively large number of choices per node. This number does not address complexity introduced via branch vs. facet selection methods, which is captured by the branch structure and instances data.  However, when we compare the branch selection versions of Feix's and Bullock and Dollar's taxonomies to their equivalent faceted versions (marked ``(eff.)''), it is clear that shifting to a faceted representation does reduce the structural complexity of the taxonomy.
Divisions between low, moderate, and high structural complexity are marked with horizontal lines in Table~\ref{tab:scdata}.

Both the flatness and the structural complexity numbers are needed because some taxonomies have a large number of categories per node but have a very simple structure with only a couple of nodes (flat, but moderate structural complexity, e.g. Fishwick), while others have a complex structure with many decision points but relatively few categories (tall, but moderate structural complexity, e.g. Cao et al.).

%%% Representational Complexity Table
\begin{table*}
\small\sf\centering
\caption{Representational complexity metric and supporting data.}
\label{tab:rcdata}
\begin{tabular}{ | c || c | c || c | c || c || c | c |}
\toprule
\multicolumn{1}{| c ||}{\bfseries Taxonomy} & {$\bm {I}$} & {$\bm {I_R}$} & {$\bm {L_S}$} & {$\bm {{L_S}_R}$} & {$\bm {RC}$} & {$\bm{AT}$} & {$\bm{FT}$} \\ 
\midrule
Ab.Acus (Siciliano) & 5 & 1 & 2 & 1 & 1.0 & - & 0 \\ 
Dudek et al. (Task) & 4 & 1 & 1 & 1 & 1.0 & - & 0 \\ 
Robin \& Lacroix & 7 & 1 & 4 & 1  & 1.0 & - & 0 \\ 
Tan et al. & 4 & 1 & 4 & 1 & 1.0 & - & 0  \\ 
Ab.Acus (IFR) & 17 & 2 & 2 & 1 & 1.5 & - & 0  \\ 
Ab.Acus (ISO Robot) & 5 & 1 & 6 & 2 & 1.5  & - & 3  \\ 
Fishwick & 19 & 2 & 2 & 1 & 1.5 & - & 0 \\ 
Gerkey \& Matari\'{c} & 8 & 1 & 6 & 2 & 1.5  & - & 0  \\ 
Korsah et al. & 21 & 2 & 2 & 1 & 1.5 & - & 0\\ 
Metzler \& Shea & 30 & 2 & 3 & 1 & 1.5 & - & 0\\ 
Shim \& Arkin & 8 & 1 & 6 & 2 & 1.5 & - & 0 \\ 
Beer et al. & 27 & 2 & 6 & 2 & 2.0 & - & 0\\ 
Bullock \& Dollar & 15 & 2 & 5 & 2 & 2.0 & - & 0 \\ 
Dudek et al. (Collective) & 2 & 1 & 9 & 3 & 2.0 & 5 & 5 \\ 
Feix et al. & 33 & 2 & 5 & 2 & 2.0 & - & 0 \\ 
Jiang \& Arkin & 48 & 2 & 6 & 2 & 2.0 & - & 0 \\ 
Yim & 8 & 1 & 11 & 3 & 2.0 & 1 & 0 \\ 
Bullock \& Dollar (eff.) & 32 & 2 & 10 & 3 & 2.5 & - & 0 \\ 
Feix et al. (eff.) & 90 & 2 & 8 & 3 & 2.5 & 1 & 0 \\ 
Carlson et al. & 32 & 2 & 9 & 3 & 2.5 & - & 0 \\ \midrule
Farinelli et al. & 576 & 3 & 18 & 4 & 3.5 & - & 0  \\ 
Yanco \& Drury & 864 & 3 & 15 & 4 & 3.5 & 2 & 0 \\ 
Daas & 8,000 & 4 & 13 & 4 & 4.0 & - & 0 \\ 
Cao et al. & 4,860 & 4 & 22 & 5 & 4.5 & 1 & 0 \\ 
Pulford & 67,500 & 5 & 14 & 4 & 4.5 & - & 0 \\ 
Winfield & 23,328 & 5 & 26 & 5 & 5.0 & - & 0 \\ 
Tsiakas et al. & 116,640 & 6 & 29 & 5 & 5.5 & 4 & 13 \\ \midrule
Zech et al. & 477,757,440 & 9 & 49 & 6 & 7.5 & - & 0 \\ 
\bottomrule
\end{tabular}
\end{table*}

\subsubsection{Representational Complexity}
\label{ssec:representational}
The {\em representational complexity} metric $RC$ provides a measure for the complexity of the concepts that the taxonomy is capable of representing.  The different branch selection methods and the different tip types each affect this metric.

The core of representational complexity is the sheer number of different instances ($I$) that are possible --- the more different instances, the more complex the potential representation.  However, representational complexity also needs to capture the complexity of the content of the individual instances, so the ranked layers sum ($L_S$) is also incorporated into the metric to capture how many different levels and categories of information are represented in a given instance.  

But there are additional factors beyond the simple instance computation and the number of layers that need to be considered.  In the current instance computation, non-binary assign value tips, especially continuous assign value tips ($AT$), are assumed to contribute a single tip.  Similarly, tips whose parent node uses the faceted branch selection mechanism ($FT$) are assumed to represent a single category tip.  Because the total instances of a faceted taxonomy are the result of multiplying the number of instances in each facet, a branch that consists of a single tip doesn't contribute to the total number of possible instances --- it represents the same value in all instances.  In practice, however, these faceted tips are more likely to represent entire sub-taxonomies than to represent a single instance.  The raw data, with the $AT$ and $FT$ values, are shown in Table~\ref{tab:rcdata}.  
(Where a table contains an $AT$ column, a dash (-) indicates that there are no assign value tips in that taxonomy and a number in parentheses refers to the number of binary yes/no assign value tips in the taxonomy.)

In both the assign value tip and the faceted tip cases, the number of instances that could be represented is extremely large rather than extremely small, introducing problems with the representational complexity computation.  Because these elements are placed at the tips of the taxonomies, the additional number of potential instances they represent only affects the detail with which a given instance can be defined and does not increase the types of things that can be defined.  

This would be a simple problem to solve if we could use a metric like the number of instances as a proxy for this complexity, but unfortunately the taxonomies with these $AT$ and $FT$ tips are not the taxonomies with the largest number of instances $I$ or the largest layers sum $L_S$.  This subset of taxonomies spans almost the full range of represented instance values, from a taxonomy with 2 instances to a taxonomy with 116,640 instances, and ranges across taxonomies representing over half of the layers sum values.  

We have three options for handling these taxonomies:  
\begin{enumerate}
\item We can ignore this problem and evaluate all the taxonomies on the basis of a metric that assumes the level of detail provided by the paper's authors is the intended level of representational complexity
\item  We can develop a metric for representational complexity that integrates the explicit number of instances as well as the effect of the $AT$ and $FT$ elements
\item  We can treat the taxonomies with $AT$ and $FT$ elements as a separate category and assess them independently
\end{enumerate}

For the purposes of this paper, the point of the representational complexity metric is to improve our understanding of taxonomy structures and forms that enable either representation of a larger set of tasks/behaviors or the ability to represent a task or behavior in more detail.  Increased representation at the tips contributes to increased detail in the representation of a single existing instance and does not expand the scope of the taxonomy to include new tasks or behaviors, affecting our ability to represent a task or behavior in more detail.  However, since we currently have no way of bounding or categorizing that increased detail, we are forced to take option 1 with some additional discussion of the results for the affected taxonomies.

What the taxonomies with assign value tips prove is that in order to adequately define particular aspects of robotic systems and tasks, we will need a mechanism to capture these continuous and discrete assign value tips at varying levels of abstraction that does not affect the rest of the structure.  The taxonomies with faceted tips indicate an incomplete taxonomic structure.

\begin{figure}[t]
\begin{center}
\includegraphics[width=3in, viewport=0.8in 2.6in 7.5in 8.1in, clip=true]{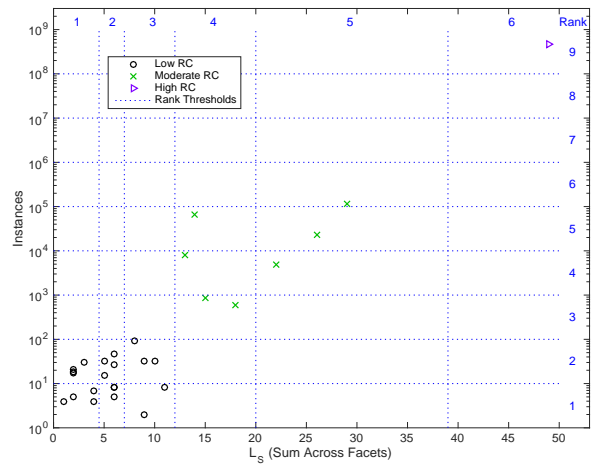}
\caption{Relationship between number of instances and maximum number of categories required to define a valid instance.  Each taxonomy is represented as a single point on the graph.  The dashed lines indicate the boundaries between rankings along the Instances and $L_S$ axes; the ranks themselves are indicated along the outside edges of the graph.}
\label{fig:layerinstance}
\end{center}
\end{figure}

Representational complexity $RC$ is therefore computed by ranking the number of instances $I$ and the sum of the number of layers $L_S$ that must be traversed to define a specific instance and then taking the average of these rankings.  The instances are ranked from 1 to 9 on a log scale as shown in Fig.~\ref{fig:layerinstance}.  This enables us to accommodate Zech et al.'s taxonomy in the upper right corner, which due to its size and structure can represent over 450 million instances.  The summed layers are ranked from 1 to 6, which separates Zech et al. from Winfield and Cao et al..  The $L_S$ number integrates the branch vs. facet structural element without requiring relative rankings for the representational complexity of the different branch structure types independent of specific taxonomies.  

The $RC$ values and the rankings of the raw instance and $L_S$ data are also shown in Table~\ref{tab:rcdata}.  In the table, divisions between low, moderate, and high representational complexity are marked with horizontal lines.

Fig.~\ref{fig:layerinstance} clearly illustrates these three clusters of representational complexity.  Low complexity taxonomies, represented by circles, have both low numbers of instances and low sums across the layers.  Taxonomies with moderate representational complexity (represented by x's) have moderately high numbers of elements that must be traversed to define an instance and can represent a moderate number of different instances, and the taxonomy that can represent orders of magnitude more instances also requires traversing a large number of elements to define an instance.

This relationship is expected, since more complex paths should be correlated with a greater number of possible instances.

\subsection{Content-Based Structural Patterns}
In addition to analyzing the taxonomies on the basis of flatness, structural complexity, and representational complexity, we must also address whether there is any correlation between the form and structure of the taxonomies and the type of categories they classify things into.  Relationships between content type and the metrics defined in Section~\ref{ssec:metrics} will be addressed in Section~\ref{ssec:complexityform}.

%%% Task Type Taxonomy Table
\begin{table*}
\small\sf\centering
\caption{Task type taxonomy data.}
\label{tab:tasktaxes}
\begin{tabular}{| c || c | c | c | c | c | c | c | c | c | c | c |}
\toprule
\multicolumn{1}{| c ||}{\bfseries Taxonomy} & {$\bm {L_M}$} & {$\bm {L_S}$} & {$\bm {E}$} & {$\bm {N}$} & {$\bm {{C}_{x}}$} & {$\bm {{C}_{n}}$} & {$\bm {I}$} & \multicolumn{1}{| c |}{$\bm {B}$} & \multicolumn{1}{| c |}{$\bm {T}$} &  \multicolumn{1}{| c |}{$\bm {AT}$} & \multicolumn{1}{| c |}{$\bm {FT}$} \\ 
\midrule
Bullock \& Dollar & 5 & 5 & 28 & 14 & 2 & 2 & 15 & BO & T & - & 0 \\ 
Bullock \& Dollar (eff.) & 1 & 10 & 15 & 6 & 2 & 2 & {\bfseries 32*} & FO & T & - & 0 \\ 
Carlson et al. & 4 & 9 & 19 & 8 & 5 & 2 & 32 & MTIII & T & - & 0 \\ 
Daas & 3 & 13 & 34 & 8 & 4 & 4 & 8,000 & FO & T & - & 0 \\ 
Dudek et al. (Task) & 1 & 1 & 4 & 1 & 4 & 4 & 4 & BO & T & - & 0 \\ 
Feix et al. & 5 & 5 & 72 & 41 & 5 & 1 & 33 & BO & T & - & 0 \\ 
Feix et al. (eff.) & 1 & 8 & 13 & 5 & 6 & 2 & {\bfseries 126*} & FO & T & 1 & 0 \\ 
Korsah et al. & 2 & 2 & 25 & 5 & 8 & 2 & 21 & BO & R, T & - & 0 \\ 
Robin \& Lacroix & 4 & 4 & 12 & 6 & 3 & 1 & 7 & BO & T & - & 0 \\ 
Shim \& Arkin & 2 & 6 & 9 & 4 & 2 & 2 & 8 & FO & T & - & 0 \\ 
Tan et al. & 2 & 4 & 6 & 3 & 2 & 2 & 4 & FO & T & - & 0 \\
\bottomrule
\end{tabular}
\end{table*}

\subsubsection{Task Type Taxonomies}
Task type taxonomies, shown in Table~\ref{tab:tasktaxes}, are distributed across the different structure types and include branch-only, facet-only, and mixed type III structures.  Although this category appears to be balanced in terms of branch-only and facet-only taxonomies, it includes two taxonomies with both branch-only and more efficient facet-only representations.  As published, branch-only taxonomies dominate this category.  In general they capture relatively low numbers of instances, indicating highly abstract or limited tasks, with one exception. Daas' taxonomy is more complex and is capable of differentiating between significantly more different instances than the others due to its faceted structure and relatively large number of elements.  As shown in Table~\ref{tab:tasktaxes}, only one taxonomy has any assign value tips and none have faceted tips.

%%% Task Property Type Taxonomy Table
\begin{table*}
\small\sf\centering
\caption{Task property type taxonomy data.}
\label{tab:taskpropertytaxes}
\begin{tabular}{| c || c | c | c | c | c | c | c | c | c | c | c |}
\toprule
\multicolumn{1}{| c ||}{\bfseries Taxonomy} & {$\bm {L_M}$} & {$\bm {L_S}$} & {$\bm {E}$} & {$\bm {N}$} & {$\bm {{C}_{x}}$} & {$\bm {{C}_{n}}$} & {$\bm {I}$} & \multicolumn{1}{| c |}{$\bm {B}$} & \multicolumn{1}{| c |}{$\bm {T}$} & \multicolumn{1}{| c |}{$\bm {AT}$} & \multicolumn{1}{| c |}{$\bm {FT}$} \\ 
\midrule
Yanco \& Drury & 3 & 15 & 31 & 8 & 9 & 1& 768 & MTIII & R, TP & 2 & 0  \\ 
\bottomrule
\end{tabular}
\end{table*}

\subsubsection{Task Property-type Taxonomies}
There is only one taxonomy that categorizes task properties, and it uses the complex mixed type III structure, as shown in Table~\ref{tab:taskpropertytaxes}.  It can also be used to capture resources, and includes assign value tips as well as category tips.  

%%% Resource Type Taxonomy Table
\begin{table*}
\small\sf\centering
\caption{Resource type taxonomy data.}
\label{tab:resourcetaxes}
\begin{tabular}{| c || c | c | c | c | c | c | c | c | c | c | c |}
\toprule
\multicolumn{1}{| c ||}{\bfseries Taxonomy} & {$\bm {L_M}$} & {$\bm {L_S}$} & {$\bm {E}$} & {$\bm {N}$} & {$\bm {{C}_{x}}$} & {$\bm {{C}_{n}}$} & {$\bm {I}$} & \multicolumn{1}{| c |}{$\bm {B}$} & \multicolumn{1}{| c |}{$\bm {T}$} & \multicolumn{1}{| c |}{$\bm {AT}$} & \multicolumn{1}{| c |}{$\bm {FT}$} \\ 
\midrule
Ab.Acus (ISO Robot) & 3 & 6 & 10 & 3 & 4 & 2 & 5 & FOB & R & - & 3  \\ 
Ab.Acus (Siciliano) & 2 & 2 & 7 & 3 & 3 & 2 & 5 & BO & R & - & 0 \\ 
Ab.Acus (IFR) & 2 & 2 & 19 & 3 & 12 & 2 & 17 & BO & R & - & 0 \\ 
Beer et al. & 2 & 6 & 12 & 4 & 3 & 3 & 27 & FO & R & - & 0 \\ 
Cao et al. & 4 & 22 & 38 & 14 & 5 & 1 & 4,860 & FOB & R & 1 & 0 \\ 
Dudek et al. (Collective) & 2 & 9 &10 & 3 & 7 & 1 & 2 & FO & R & 5 & 5  \\ 
Farinelli et al. & 3 & 18 & 28 & 11 & 4 & 2 & 576 & FO & R & - & 0 \\ 
Fishwick & 2 & 2 & 24 & 6 & 6 & 2 & 19 & BO & R & - & 0 \\ 
Gerkey \& Matari\'{c} & 2 & 6 & 9 & 4 & 3 & 2 & 8 & FO & R & - & 0 \\ 
Jiang \& Arkin & 2 & 6 & 14 & 4 & 4 & 3 & 48 & FO & R & - & 0 \\ 
Korsah et al. & 2 & 2 & 25 & 5 & 8 & 2 & 21 & BO & R, T & - & 0 \\ 
Metzler \& Shea & 3 & 3 & 40 & 11 & 13 & 1 & 30 & BO & R & - & 0 \\ 
Pulford & 2 & 14 & 47 & 8 & 15 & 3 & 67,500 & FO & R & - & 0 \\ 
Tsiakas et al. & 4 & 29 & 64 & 39 & 6 & 2 & 116,640 & FO & R & 4 & 13 \\ 
Winfield & 3 & 26 & 43 & 16 & 4 & 2 & 23,328 & FO & R & - & 0 \\ 
Yanco \& Drury & 3 & 15 & 31 & 8 & 9 & 2 & 864 & MTIII & R, TP & 2 & 0 \\ 
Yim & 5 & 11 & 16 & 7 & 3 & 1 & 8 & BOF & R & - & 2  \\ 
Zech et al. & 4 & 49 & 87 & 28 & 6 & 1 & 477,757,440 & FO & R & (5) & 0  \\  
\bottomrule
\end{tabular}
\end{table*}

\subsubsection{Resource-type Taxonomies}
There are many resources taxonomies.  In contrast to the more balanced structure representation in the task taxonomies, the resource-centric taxonomies are weighted heavily towards structures that include faceted nodes, as shown in Table~\ref{tab:resourcetaxes}.  Again, the variable values cover a wide range, from low to high, without an obvious pattern.

\subsection{Scale-Based Structural Patterns}
\label{ssec:sizepatterns}
This section addresses relationships between taxonomy scale and taxonomy structure.  Table~\ref{tab:numElements} includes the variables relevant to a discussion of relationships between scale and structure:  the number of elements ($E$), the maximum number of layers that can be traversed along a single branch ($L_M$), the maximum number of choices possible at a single node (${C}_{x}$), and the average number of choices per node (${C}_{A}$).  These size-related variables are addressed in the context of the structure type ($B$) and the category type ($T$).
\begin{table*}
\small\sf\centering
\caption{Scale-related taxonomy data.}
\label{tab:numElements}
\begin{tabular}{ | c || c | c | c | c | c | c | c | c |}
\toprule
\multicolumn{1}{| c ||}{\bfseries Taxonomy} & {$\bm {L_M}$} & {$\bm {E}$} & {$\bm {{C}_{x}}$} & {$\bm {{C}_{A}}$} & \multicolumn{1}{| c |}{$\bm {B}$} & \multicolumn{1}{| c |}{$\bm {T}$} & \multicolumn{1}{| c |}{$\bm {AT}$} & \multicolumn{1}{| c |}{$\bm {FT}$} \\ 
\midrule
Dudek et al. (Task) & 1 & 4 & 4 & 4.00 & BO & T & - & 0 \\ 
Tan et al. & 2 & 6 & 2 & 2.00 & FO & T & - & 0 \\ 
Ab.Acus (Siciliano) & 2  & 7 & 3 & 2.33 & BO & R & - & 0 \\ 
Gerkey \& Matari\'{c} & 2 & 9 & 3 & 2.25 & FO & R & - & 0 \\ 
Shim \& Arkin & 2 & 9 & 2 & 2.25 & FO & T & - & 0  \\ 
Ab.Acus (ISO Robot) & 3 & 10 & 4 & 3.33 & FOB & R & - & 3 \\ 
Dudek et al. (Collective) & 2 &10 & 7 & 3.33 & FO & R & 5 & 5  \\ 
Beer et al. & 2 & 12 & 3 & 3.00 & FO & R & - & 0  \\ 
Robin \& Lacroix & 4 & 12 & 3 & 2.00 & BO & T & - & 0  \\ 
Feix et al. (eff.) & 1 & 13 & 6 & 2.60 & FO & T & 1 & 0  \\ 
Jiang \& Arkin & 2 & 14 & 4 & 3.50 & FO & R & - & 0  \\ 
Bullock \& Dollar (eff.) & 1 & 15 & 2 & 2.50 & FO & T & - & 0  \\ 
Yim & 5 & 16 & 3 & 2.29 & BOF & R & - & 2   \\ 
Ab.Acus (IFR) & 2 & 19 & 12 & 6.33 & BO & R & - & 0 \\ 
Carlson et al. & 4 & 19 & 5 & 2.38 & MTIII & T & - & 0 \\ 
Fishwick & 2 & 24 & 6 & 4.00 & BO & R & - & 0  \\ 
Korsah et al. & 2 & 25 & 8 & 5.00 & BO & R, T & - & 0  \\ 
Bullock \& Dollar & 5 & 28 & 2 & 2.00 & BO & T & - & 0  \\ 
Farinelli et al. & 3 & 28 & 4 & 2.55 & FO & R & - & 0  \\ 
Yanco \& Drury & 3 & 31 & 9 & 3.88 & MTIII & R, TP & 2 & 0  \\ 
Daas & 3 & 34 & 4 & 4.25 & FO & T & - & 0  \\ 
Cao et al. & 4 & 38 & 5 & 2.71 & FOB & R & 1 & 0  \\ 
Metzler \& Shea & 3 & 40 & 13 & 3.64 & BO & R & - & 0  \\ 
Winfield & 3 & 43 & 4 & 2.69 & FO & R & - & 0  \\ 
Pulford & 2 & 47 & 15 & 5.88 & FO & R & - & 0  \\ 
Tsiakas et al. & 4 & 64 & 6 & 1.64 & FO & R & 4 & 13  \\ 
Feix et al. & 5 & 72 & 5 & 1.80 & BO & T & - & 0  \\ 
Zech et al. & 4 & 87 & 6 & 3.11 & FO & R & (5) & 0  \\   
\bottomrule
\end{tabular}
\end{table*}

\subsubsection{Number of Elements}
These are all small taxonomies relative to a large-scale taxonomy like today's Taxonomy of Species or a large-scale classification system like the Library of Congress classification system.  They each address one small aspect of the field of robotics and average about 26 elements, ranging from 4 to 87.  Larger scale robotics taxonomies that attempt to capture more of these elements into an internally consistent whole were not found in the literature.

In Table~\ref{tab:numElements}, there is no obvious relationship between structural organization ($B$) and the number of individual categories represented within a given taxonomy ($E$).  Branch-only and facet-only structures are represented by taxonomies with both high and low numbers of elements.  While the mixed types seem to skew towards the larger taxonomies, that can be explained by the fact that you cannot have a mixed structure in a taxonomy that only has 2 levels.  The data does not support drawing further conclusions about the relationship between the mixed type taxonomies and taxonomy size.

Even within the taxonomies with limited selection options (low ${C}_{x}$), we still see mixed structures ($B$) like facet-over-branch and branch-over-facet.  Less than a third of the taxonomies use a strict branch selection approach to categorization.  The majority use some form of faceted classification, where the user is expected to select one item from each branch rather than to select a branch.  This is independent of size; it holds for a taxonomy with 6 elements all the way up to a taxonomy with 47 elements.  

\subsubsection{Number of Layers}
We see a similar lack of correlation between the number of layers ($L_M$) in a taxonomy and its structure ($B$).  Again, branch-only and facet-only taxonomies are represented in both the smallest and largest number of layers.

\subsubsection{Number of Choices}
Table~\ref{tab:numElements} also shows no correlation between the average number of choices at any given node (${C}_{A}$) and the structure ($B$) or content ($T$) or number of elements ($E$) in a given taxonomy.

Most of the taxonomies do not ask us to make complex or subtle distinctions at any given node, but to make simple choices between two or three options.  The ${C}_{x}$ and ${C}_{A}$ columns capture the maximum and average number of possible categories at a given node within that taxonomy (the Assign Value selection type contributes 1 in these columns).   About half (15/28) of the taxonomies average below 3 choices per node, and half of those (8/15) have at least one node with 4 or more choices, indicating a skew towards smaller selection opportunities even in taxonomies with larger maximum node choices.

The taxonomies with assign value ($AT$) or faceted tips ($FT$), with potentially the largest number of choices, likewise do not skew towards or away from specific structures.  Taxonomies with faceted tips include facet-only, facet-over-branch, and branch-over-facet structures.  Taxonomies with assign value tips include facet-only, facet-over-branch, and mixed type III structures.

Since the faceted taxonomies skew towards resource content, it is unsurprising that the assign value and faceted tip taxonomies (which also all contain facets) are likewise dominated by categorizing resources rather than tasks or task properties.  Only one of these taxonomies categorizes tasks, rather than resources:  the version of Feix et al. that was revised to be more efficient.

 \subsection{Structural Patterns Characteristic of Robotics Taxonomies}
 \label{ssec:structuralpatterns}
Table~\ref{tab:rawdata1} shows the complete list of the various taxonomy properties sorted by structure type ($B$).  There are no mixed type IV taxonomies.  This is consistent with other domains where faceted classification is used, such as real estate listings (faceted), airplane flights (branch-over-facet or facet-only), or parts catalogs (branch-over-facet).  While we do have one branch-over-facet taxonomy and two each of the facet-over-branch and mixed type III taxonomies, the more complicated mixed type IV variant has not been needed in the robotics community.

In general, the community has gravitated towards faceted taxonomic structures to represent robot information, as the kinds of categories that are useful to the community often involve the intersection of several different types of information.  A robot, for example, is defined at the intersection of its sensors, its motors, its processing capability, its power, and its behaviors.

\subsection{Complexity and Form}
\label{ssec:complexityform}
\begin{figure}[t]
\begin{center}
\includegraphics[width=3in, viewport=0.4in 2.6in 7.5in 8.1in, clip=true]{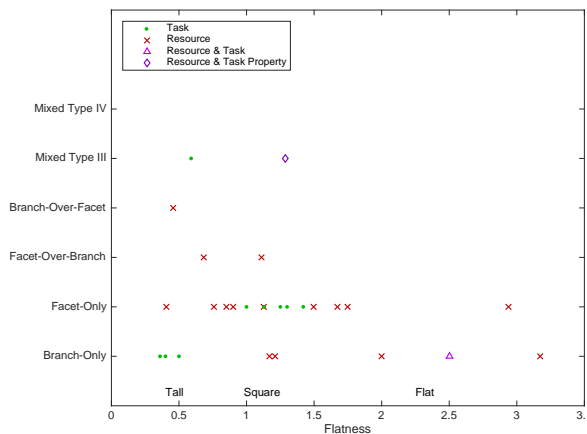}
\caption{Relationship between flatness, branch structure, and content type for each taxonomy.  Each taxonomy is represented as a single point on the graph, with taxonomies containing task information represented with points, those containing resource information with x's, taxonomies that combine resource and task information with triangles, and taxonomies that combine resources and task properties with diamonds. }
\label{fig:formstructcontent}
\end{center}
\end{figure}

Attempts to determine whether there are relationships between particular forms or branch structures and content types fail to show correlations, as seen in Fig.~\ref{fig:formstructcontent}.

It can be seen in this figure that flat taxonomies are restricted to branch-only or facet-only and do not appear in the more complex structures.  This is expected, since our current definition of {\em flat} specifies a shallow structure, which for the taxonomies reviewed here allowed only up to 2 layers, which in turn prevented the more complex structures from arising.  

\begin{figure}[t]
\begin{center}
\includegraphics[width=3in, viewport=0.8in 2.6in 7.5in 8.1in, clip=true]{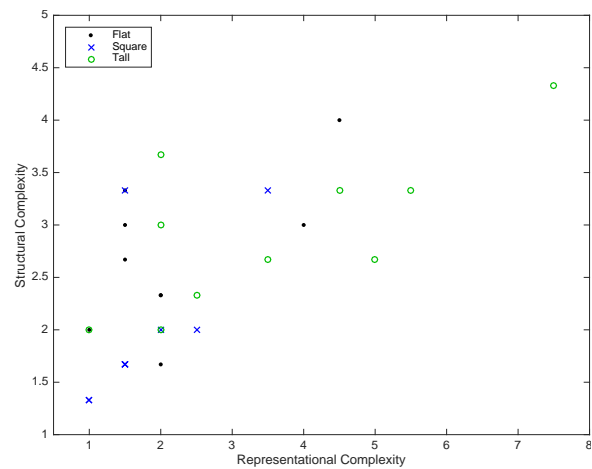}
\caption{Relationship between structural complexity, representational complexity and flatness for each taxonomy.  Each taxonomy is represented as a single point on the graph, with flat taxonomies represented with points, square taxonomies represented with x's, and tall taxonomies represented with circles.}
\label{fig:complexity}
\end{center}
\end{figure}

In Fig.~\ref{fig:complexity} we again see a distinct lack of correlation.  There is no clear relationship between structural and representational complexity and flatness.  There are flat, square, and tall taxonomies at every level of structural and representational complexity.

\begin{figure}[t]
\begin{center}
\includegraphics[width=3in, viewport=0.8in 2.6in 7.5in 8.1in, clip=true]{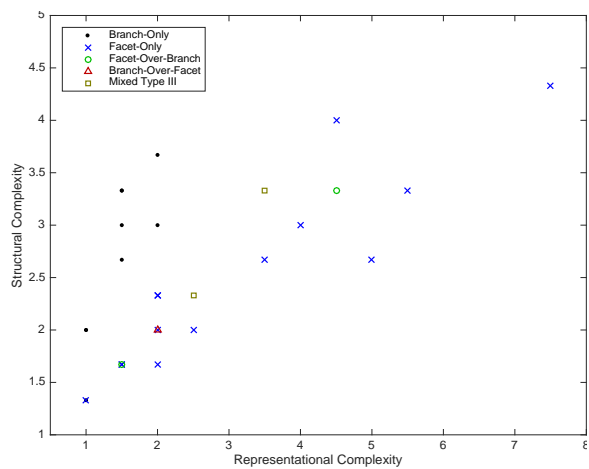}
\caption{Relationship between structural complexity, representational complexity and branch structure for each taxonomy.  Each taxonomy is represented as a single point on the graph, with branch-only taxonomies represented with points, facet-only taxonomies represented with x's, branch-over-facet taxonomies represented with triangles, and mixed type III taxonomies represented with squares.  }
\label{fig:complexitystructure}
\end{center}
\end{figure}

Having determined the relationship between flatness and structure, we turn to the relationships between representational complexity and structure and between structural complexity and structure, as shown in Fig.~\ref{fig:complexitystructure}.

Here, we do see a pattern in the relationship between structural complexity and representational complexity.  There is a clean break between taxonomies with low and high representational complexity, and high representational complexity only occurs in conjunction with high structural complexity.  (High structural complexity does not, however, imply high representational complexity.)  

Mixed type III taxonomies tend to be moderately complex rather than highly complex.  Although the structure itself is complicated, it is not correlated with use cases requiring the ability to represent large numbers of instances or complex relationships involving many factors.  Instead, this structure is mainly used by taxonomies for moderate numbers of both instances and elements.

The only example of a branch-over-facet taxonomy is less complex compared to simpler branch-only and facet-only taxonomies, able to represent relatively few instances and containing a relatively low number of paths.  

There are no branch-only taxonomies with high representational complexity.  Faceted structures with increased representational complexity correlate with increased structural complexity.  If we exclude branch-only taxonomies, we can see a clear correlation between increased representational complexity and increased structural complexity.

\subsection{Discussion}
It would be reasonable to assume that taxonomies with more complicated forms, like mixed types III and IV, would be used to create large taxonomies with increased representational power or to reduce the scale of the taxonomy while maintaining high representational power, but no such relationship is demonstrated.

It would be reasonable to assume that flat taxonomies would be more likely to have been based on simple structures, yet they include taxonomies with relatively high structural and representational complexity and taxonomies with relatively low structural and representational complexity. 

There is only one case where there is a clear relationship between taxonomic properties.  We cannot use the content type, the scale, the form, or the flatness to predict any other property of these taxonomies.  Even the structural complexity is not predictive of any other property.  Similarly, we cannot use any property to predict the content type, the scale, the form, or the flatness.  

Unexpectedly, only the combination of faceted form and representational complexity is sufficient to predict structural complexity.  This review did not identify any other combinations that provided insight into the appropriate form or structural properties of a larger, unified taxonomy.

\section{Conclusions}
Before we can generate a unified taxonomy, there is significant work remaining that must be completed.  We need to define a consistent branch selection notation that will enable us to easily understand which branches of a taxonomy require a select-one (branch-only) approach to categorization and which branches require a faceted approach.  We need to define consistent tip notation that will enable us to easily understand when the purpose of the taxonomy is to allocate something to a category and when termination involves assignment of a value.  

But beyond these notational issues, we have integration issues.  Each of these taxonomies is different --- they address different subjects and different properties of those subjects.  Before we can develop an appropriate unified taxonomy, we need to determine what categorization that specific taxonomy should provide.  Is it sufficient to define a resources taxonomy, or is it important to separate the resources into platform and environmental facets?  Do we need to separate tasks from task properties, or are they properly facets of a combined task categorization taxonomy?  How do we define the overlap between concepts that will allow us to integrate existing taxonomies and identify where they do and do not overlap?  

While we still have many questions left to answer, we have learned several useful things from this review.  

First, some papers didn't provide structures for their taxonomies, just lists.  These did not always translate cleanly into hierarchical or faceted underlying structures, so structures had to be modified or created in order to capture their taxonomic relationships. (This was particularly evident in Section~\ref{sssec:beer}).  Just as it will be important to develop rationales for connecting concepts across taxonomies, as noted in Farinelli et al.'s taxonomy (Section~\ref{sssec:farinelli}) it will be critical to ensure that we have correctly captured the intent of these lists.

Second, while it was possible to assign content types to various aspects of the taxonomies (see Robin and Lacroix's taxonomy in Section~\ref{sssec:robin} and Gerkey and Matari\'{c}'s taxonomy in Section~\ref{sssec:gerkey}), decisions  will need to be made about when and where to separate these taxonomies into their component elements to support the creation of content-based taxonomy facets versus leaving them as designed and supplementing them with references to other facets of the aggregated taxonomy.

Third, it was impossible to simplify some taxonomies beyond a certain point --- there is a minimum level of complexity that our larger taxonomy must be able to represent (see Beer et al.'s taxonomy in Section~\ref{sssec:beer} and Yim's taxonomy in Section~\ref{sssec:yim}).  The fact that it was not possible to simplify Yim's branch-over-facet taxonomy  in particular indicates that while the more complex taxonomic structures may be rare in the simpler taxonomies, they will be necessary as we merge them into the larger taxonomy.  There exists a lower bound on how simple we will be able to make the structure of the aggregated taxonomy.

Fourth, in some cases it was possible to generate a simpler faceted structure to replace a more complex branching structure.  In these cases, additional potential instances were introduced because of invalid combinations of facets (see Bullock and Dollar's taxonomy inSection~\ref{sssec:bullock} and Feix et al.'s taxonomy in Section~\ref{sssec:feix}).  As we construct the larger taxonomy, we will need a mechanism to define invalid combinations of facets.  In theory, a faceted taxonomy includes only facets for which definitions are needed across all instances, but in practice, robotics taxonomies illustrate that in the interests of compactness we are likely to need to figure out how to define and represent invalid instances. 

Fifth, some taxonomies require the ability to be terminated at any level of abstraction within the structure, defining new tips on an instance-by-instance basis from intermediate categories (see Metzler and Shea's taxonomy in Section~\ref{sssec:metzler} and Pulford's taxonomy in Section~\ref{sssec:pulford}).  This can be used to support representation of levels of abstraction, but some notational mechanism will be necessary to differentiate between an instance that is incomplete and is missing information and an instance that is the result of this truncation and has been deliberately defined at a given level of abstraction. 

Sixth, the content types of the taxonomies, which might seem on the surface to be their most important aspect, were often mixed and dependent on use case for categorization (see Korsah et al.'s taxonomy in Section~\ref{sssec:korsah}).  In order to define the larger taxonomy, it will be necessary to identify the various uses to which a given set of concepts may be applied and ensure that each of those uses is explicitly called out.  This is highlighted in the relationship between Yim's taxonomy of locomotion (Section~\ref{sssec:yim}) and the Ab.Acus document's taxonomy of robot structure (Section~\ref{sssec:abacus3}).  It drives the need for any larger taxonomy to have both a platform-specific facet and a behavior-specific facet.

Finally, some taxonomies require the ability to define multiple instances of a single facet as part of a single instance of the larger taxonomy (see Winfield's taxonomy in Section~\ref{sssec:winfield}).  Instead of just one camera sensor on a given robot, most robots will have several.  Instead of just one robot, we will need to specify heterogeneous teams.  Instead of one metric, tasks may have several.   In order to specify the capabilities of that robot or the requirements of that task, we will need to include all of them, not just one.  A larger taxonomy will need to be able to accommodate each of these scenarios.  

The development of this new aggregated taxonomy will require the concurrent development of mechanisms to support all of these structure- and form-based needs.  The fragmented nature of the existing taxonomies captured here illustrates the difficulty of creating the desired taxonomy, but examination of their underlying form and structure identifies properties that the aggregated taxonomy will need and enables us to start developing it.

\begin{acks}
Thanks to the supervisors who supported me in the development of this manuscript.  Many thanks are also due to Victoria Edwards, who contributed to extensive technical discussions regarding the establishment of an aggregated taxonomy.
\end{acks}

%\begin{biog}
%To typeset an
%"Author biography" section.
%\end{biog}

\begin{dci}
The Author declares that there is no conflict of interest.  
\end{dci}

\begin{funding}
This research received no specific grant from any funding agency in the public, commercial, or not-for-profit sectors.
\end{funding}

%\begin{sm}
%To typeset a
%"Supplemental material" section.
%\end{sm}

%\theendnotes

%\bibliographystyle{SageH}
%\bibliography{TaxonomyOfCapabilities}

\end{document}